\documentclass[10pt,twocolumn,letterpaper]{article}
\usepackage[pagenumbers]{iccv} 
\definecolor{iccvblue}{rgb}{0.21,0.49,0.74}
\usepackage[pagebackref,breaklinks,colorlinks,linkcolor=red,citecolor=iccvblue,urlcolor=iccvblue]{hyperref}
\def\paperID{2769} 
\def\confName{ICCV}
\def\confYear{2025}

\usepackage{amsmath,amsfonts,bm}
\usepackage{amssymb}
\usepackage{amsthm}
\usepackage{mathtools}
\theoremstyle{plain}
\newtheorem{theorem}{Theorem}[section]

\newtheorem{lemma}[theorem]{Lemma}

\theoremstyle{definition}
\newtheorem{definition}[theorem]{Definition}
\newtheorem{assumption}[theorem]{Assumption}
\theoremstyle{remark}

\providecommand{\eg}{{\sl e.g.}}

\providecommand{\etal}{{\sl et al.}}

\def\eqref#1{equation~\ref{#1}}

\def\1{\bm{1}}

\DeclareMathAlphabet{\mathsfit}{\encodingdefault}{\sfdefault}{m}{sl}
\SetMathAlphabet{\mathsfit}{bold}{\encodingdefault}{\sfdefault}{bx}{n}

\usepackage{graphicx, lipsum}
\usepackage{url}            
\usepackage{booktabs}       
\usepackage{amsfonts}       
\usepackage{nicefrac}       
\usepackage{microtype}      
\usepackage{xcolor}         
\usepackage{enumitem}
\usepackage{pifont}
\usepackage{graphicx}
\usepackage{cancel}
\usepackage{subcaption}
\usepackage{afterpage}
\usepackage{caption}
\usepackage{overpic}
\usepackage{amsmath,amsfonts,amssymb,amsthm}
\usepackage{mathtools}
\usepackage{algpseudocode}
\usepackage{multicol}
\usepackage{multirow}
\usepackage{indentfirst}
\usepackage{stfloats}
\makeatletter
\newif\ifreviewmode
\@ifpackagewith{iccv}{review}{\reviewmodetrue}{\reviewmodefalse}
\makeatother

\newcommand{\addauthorshipfootnote}{%
    \ifreviewmode
    \else
        \begingroup
        \renewcommand{\thefootnote}{}
        \footnotetext[0]{%
            \raggedright
            \footnotesize
            $^{*}$Equal contribution. \quad $^{\dagger}$Corresponding author.
        }%
        \endgroup
    \fi
}

\usepackage[ruled,vlined]{algorithm2e}

\title{Signal Processing Meets SGD: From Momentum to Filter}

\author{%
Zhipeng Yao \textsuperscript{1} $^{*}$ \quad 
Rui Yu \textsuperscript{2} $^{*}$ \quad 
Guisong Chang \textsuperscript{3} \quad 
Ying Li \textsuperscript{1} \quad 
Yu Zhang \textsuperscript{1,4} $^{\dagger}$ \quad 
Dazhou Li \textsuperscript{1} $^{\dagger}$ \\
\textsuperscript{1}Shenyang University of Chemical Technology \quad 
\textsuperscript{2}University of Louisville \\
\textsuperscript{3}Northeastern University \quad 
\textsuperscript{4}University of Macau \\
{\tt \small yiucp@outlook.com, rui.yu@louisville.edu, gschang@mail.neu.edu.cn} \\
{\tt \small Gooddayli12358@outlook.com, zhangy@syuct.edu.cn, lidazhou@syuct.edu.cn}
}

\begin{document}
\maketitle
\addauthorshipfootnote

\begin{abstract}
In deep learning, stochastic gradient descent (SGD) and its momentum-based variants are widely used for optimization. However, the internal dynamics of these methods remain underexplored. In this paper, we analyze gradient behavior through a signal processing lens, isolating key factors that influence gradient updates and revealing a critical limitation: momentum techniques lack the flexibility to adequately balance bias and variance components in gradients, resulting in gradient estimation inaccuracies. To address this issue, we introduce a novel method SGDF (SGD with Filter) based on Wiener Filter principles, which derives an optimal time-varying gain to refine gradient updates by minimizing the mean square error in gradient estimation. This method yields an optimal first-order gradient estimate, effectively balancing noise reduction and signal preservation. Furthermore, our approach could extend to adaptive optimizers, enhancing their generalization potential. Empirical results show that SGDF achieves superior convergence and generalization compared to traditional momentum methods, and performs competitively with state-of-the-art optimizers. \footnote{The code is available at \textcolor{magenta}{https://github.com/LilYau350/SGDF-Optimizer}}


\end{abstract}

\section{Introduction}
During the training process, the optimizer serves as a critical component of adjusting model parameters to capture underlying data patterns effectively. It refines and adjusts model parameters to ensure that the model can recognize underlying data patterns. Beyond updating weights, the optimizer's role includes strategically navigating complex loss landscapes~\cite{du2018power} to locate regions that offer the best generalization~\cite{keskar2022large}. The chosen optimizer significantly impacts training efficiency, influencing model convergence speed, generalization performance, and resilience to data distribution shifts~\cite{2007Scaling}. A poor optimizer may lead to suboptimal convergence or even failure to converge, while an appropriate one can speed up learning and enhance robustness~\cite{2016An}. Thus, the design and refinement of optimizers remain essential challenges in enhancing the capabilities of machine learning models.

Stochastic Gradient Descent (SGD)~\cite{1951a} and its derivatives, including momentum-based methods~\cite{sutskever2013importance,polyak1964some} and adaptive approaches like Adam~\cite{kingma2014adam} and RMSprop~\cite{hinton2012neural}, are fundamental to deep learning optimization. While these techniques have significantly improved training efficiency~\cite{chandramoorthy2022generalization}, they exhibit inherent limitations in handling the high-dimensional, non-convex landscapes typical of deep learning~\cite{goodfellow2016deep}. Specifically, adaptive methods offer faster convergence but often lead to poor generalization~\cite{keskar2017improving}. In response, numerous Adam variants~\cite{chen2018closing, liu2019variance, luo2019adaptive, zhuang2020adabelief} have been developed to address these issues by refining adaptive learning rate adjustments. Although these modifications provide some improvements, they have yet to fully bridge the generalization gap, underscoring the need for further advancements in optimization techniques.

Actually, the issues that arise from the optimizer during training, particularly in terms of optimization and generalization, are inherently tied to the trade-off between bias and variance~\cite{2014Neural, nguyen2022algorithmic}. High bias leads to underfitting, while high variance results in overfitting. Similarly, the gradients used by the optimizer to update model weights also face this challenge. Intuitively, high bias in the gradients may lead to convergence at a suboptimal plateaus~\citep{yang2023stochastic,2016Understanding}, while high variance can lead to instability in the optimization path, causing oscillations that hinder convergence~\cite{bottou2018optimization, duchi2019variance}. Therefore, a good optimizer should strike a balance between the bias and variance in its gradient estimates.

From a statistical signal processing perspective, we analyze the mechanism behind optimizer updates. Specifically, we decompose the optimizer's gradients used for updating model weight into bias and variance components. Then, We identify a key limitation in momentum-based optimization techniques supplemented with examining the statistical distribution of gradients within the model: they struggle to balance bias and variance components in gradients, often introducing a gradient shift phenomenon, which we term \textit{bias gradient estimate}. This bias estimate, arising from fixed momentum coefficients, accumulates over time, leading to bias. As a result, the model may struggle to adapt to variations in curvature across different layers, resulting in suboptimal or directionally skewed updates~\cite{zhang2017yellowfin, dozat2016incorporating}.

To address this issue, we introduce SGDF, a novel method that uses principles from Wiener Filter to adjust gradient estimation dynamically. SGDF derives an optimal, time-varying gain to minimize mean-squared error in gradient estimation, balancing noise reduction with signal preservation. This filter mechanism provides a more accurate first-order gradient estimate and avoids the limitations of fixed momentum parameters, allowing SGDF to adjust dynamically throughout training. Additionally, SGDF’s flexibility extends to adaptive optimizers, which enhance generalization across a range of tasks. Through extensive empirical validation across diverse model architectures and visual tasks, we demonstrate that SGDF consistently outperforms traditional momentum-based and variance reduction methods, achieving competitive or superior results relative to state-of-the-art optimizers. 
 
The main contributions of this paper can be summarized as follows:
\begin{itemize}[leftmargin=*]
    \item We quantify the bias-variance trade-off in momentum-based gradient estimation (EMA and CM) using a unified SDE framework, revealing their static limitations.
    \item We introduce SGDF, an optimizer that combines historical and current gradient data to estimate the gradient's variance, addressing the trade-off between bias and variance in the momentum method.
    \item We theoretically analyze the convergence property of SGDF in both convex optimization and non-convex stochastic optimization (Sec.~\ref{sec:section3.4}), and empirically verify the effectiveness of SGDF (Sec.~\ref{sec:section4}).
    \item We preliminarily explore the extension of SGDF's first-moment filter estimation to adaptive optimization algorithms (\eg, Adam), which shows a promising enhancement in their generalization capability (Sec.~\ref{sec:section4.4}), surpassing traditional momentum-based methods.
\end{itemize}

\smallskip
\section{Related Works}
\textbf{Variance Reduction to Adaptive Methods.} In the early stages of deep learning development, optimization algorithms focused on reducing the variance of gradient estimation~\citep{balles2018dissecting, Defazio2014saga, johnson2013accelerating, Schmidt2017minimizing} to achieve a linear convergence rate. Subsequently, the emergence of adaptive learning rate methods~\citep{dozat2016incorporating, Duchi2011Adaptive, 2012ADADELTA} marked a significant shift in optimization algorithms. While SGD and its variants have advanced many applications, they come with inherent limitations. They often oscillate or become trapped in sharp minima~\citep{2017The}. Although these methods can lead models to achieve low training loss, such minima frequently fail to generalize effectively to new data~\citep{2015Train, 2022On}. This issue is exacerbated in the high-dimensional, non-convex landscapes characteristic of deep learning settings~\citep{2014Identifying, lucchi2022theoretical}.

\textbf{Sharp and Flat Solutions.} The generalization ability of a deep learning model depends heavily on the nature of the solutions found during the optimization process. Keskar~\etal~\citep{keskar2017large} demonstrated experimentally that flat minima generalize better than sharp minima. SAM~\citep{foret2021sharpness} theoretically showed that the generalization error of smooth minima is lower than that of sharp minima on test data, and further proposed optimizing the zero-order smoothness. GSAM~\citep{zhuang2022surrogate} guides the training of the model by introducing a surrogate gap, which helps to find a smoother solution space with better generalization. GAM~\citep{Zhang2023} improves SAM by simultaneously optimizing the prediction error and the number of paradigms of the maximum gradient in the neighborhood during the training process. Adaptive Inertia~\citep{xie2022adaptive} aims to balance exploration and exploitation in the optimization process by adjusting the inertia of each parameter update. This adaptive inertia mechanism helps the model avoid falling into sharp local minima.

\textbf{Second-Order and Filter Methods.} The recent integration of second-order information into optimization problems has gained popularity~\citep{liu2023sophia, yao2020adahessian}. Methods such as Kalman Filter~\citep{Kalman1960} combined with Gradient Descent incorporate second-order curvature information~\citep{Ollivier2019, Vuckovic2018}. The KOALA algorithm~\citep{davtyan2022koala} posits that the optimizer must adapt to the loss landscape. It adjusts learning rates based on both gradient magnitudes and the curvature of the loss landscape. However, it should be noted that the Kalman Filter framework introduces more complex parameter settings, which can hinder understanding and application.


\begin{figure*}[t]
\centering
\begin{minipage}[b]{0.35\textwidth}
    \centering
    \includegraphics[width=\textwidth]{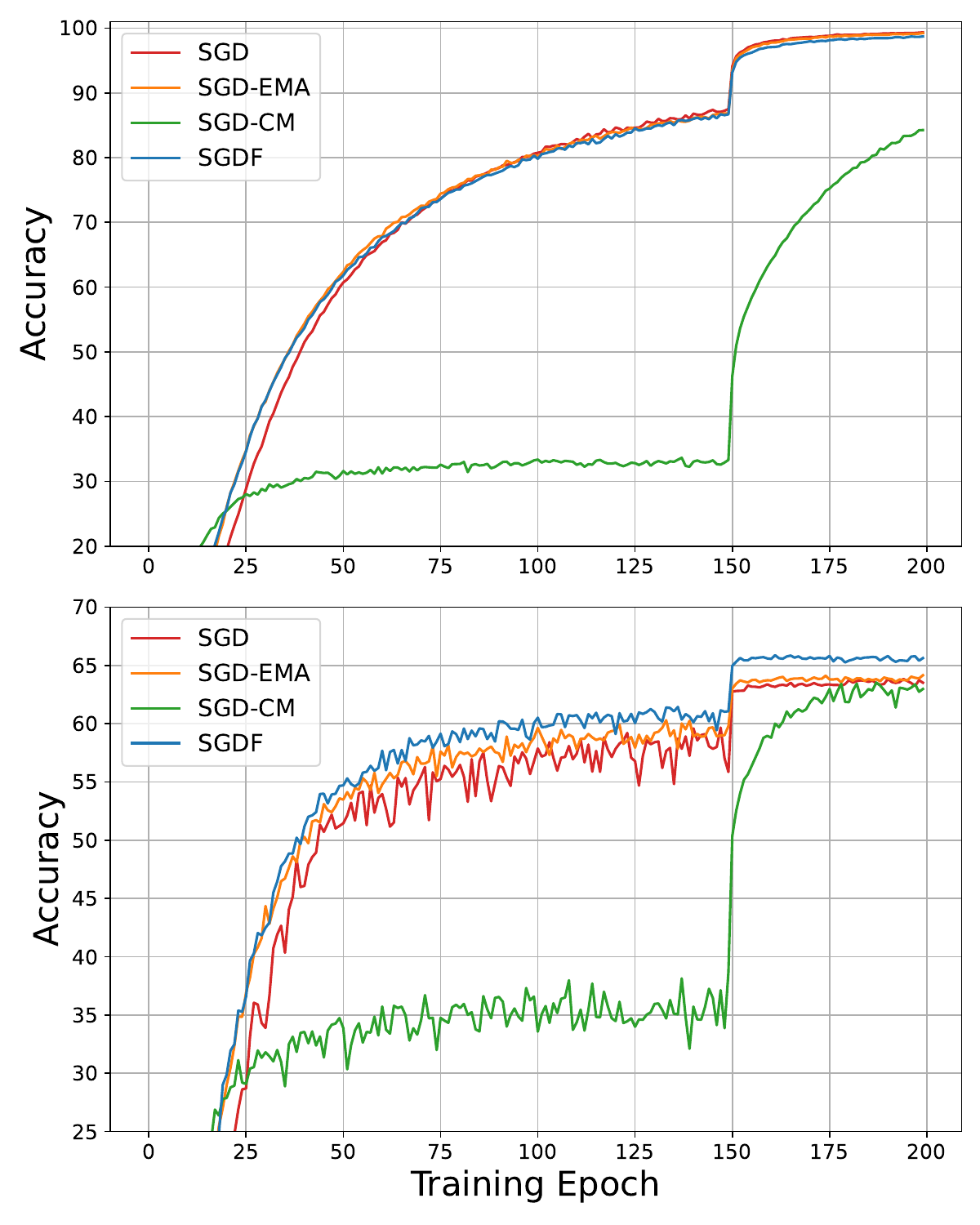}
    \caption{Train the VGG model on the CIFAR-100 dataset using the same initial learning rate of 0.1, and multiply it by a factor of 0.1 at the 150th epoch.}
    \label{fig:accuracy}
\end{minipage}
\hfill
\begin{minipage}[b]{0.60\textwidth}
    \centering
    \begin{subfigure}[b]{0.45\textwidth}
        \centering
        \includegraphics[width=\textwidth]{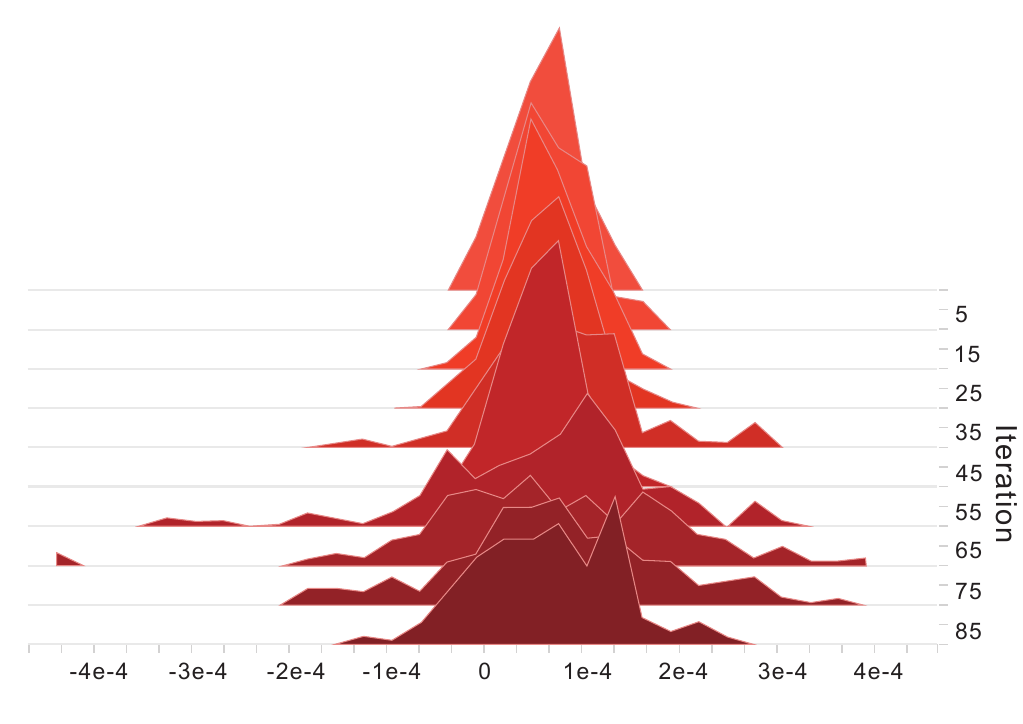}
        \subcaption{SGD}
        \label{subfig:SGD}
    \end{subfigure}%
    \begin{subfigure}[b]{0.45\textwidth}
        \centering
        \includegraphics[width=\textwidth]{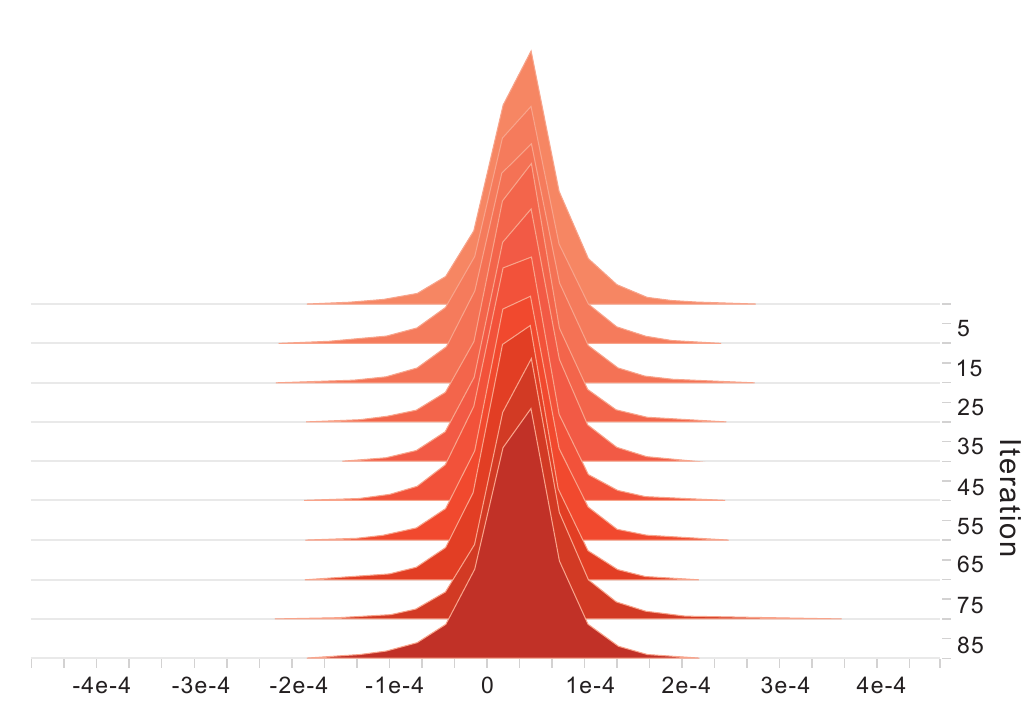}
        \subcaption{SGD-EMA}
        \label{subfig:SGD-EMA}
    \end{subfigure}
    
    \vskip\baselineskip

    \begin{subfigure}[b]{0.45\textwidth}
        \centering
        \includegraphics[width=\textwidth]{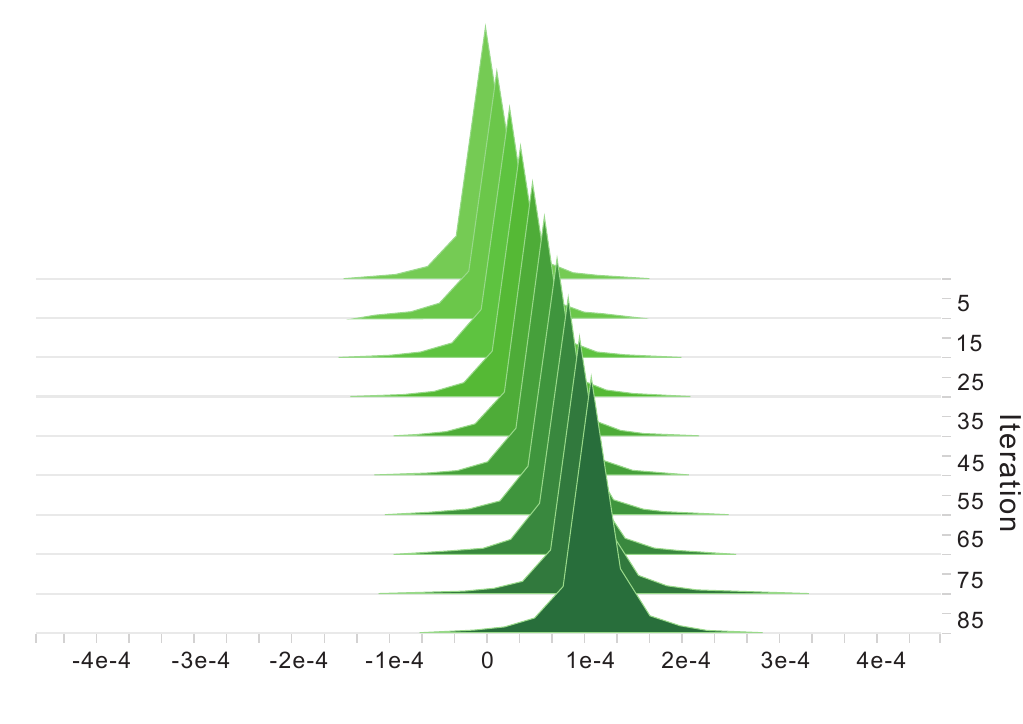}
        \subcaption{SGD-CM}
        \label{subfig:SGD-CM}
    \end{subfigure}
    \begin{subfigure}[b]{0.45\textwidth}
        \centering
        \includegraphics[width=\textwidth]{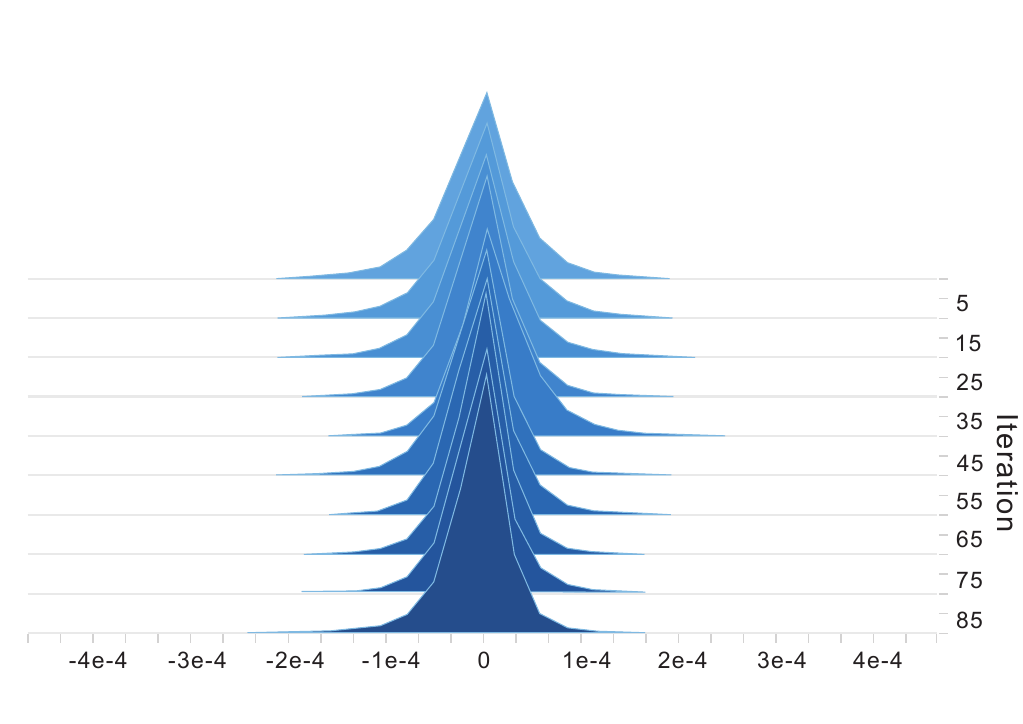}
        \caption{SGDF}
        \label{subfig:SGDF}
    \end{subfigure}%
    \caption{The gradient histogram of the VGG on the CIFAR-100 dataset. The x-axis is the gradient value and the height is the frequency. SGD trains the VGG without BN, the variance of the gradient fluctuates dramatically and the update is unstable.}
    \label{fig:histogram}
\end{minipage}
\end{figure*}

\section{The Gradient Estimation Dilemma}
\subsection{Bias and Variance}
Stochastic gradient-based optimization lies at the core of modern machine learning, yet it grapples with a fundamental challenge: the trade-off between gradient bias and variance. To dissect this dilemma, we begin by unifying two prominent momentum strategies under a single framework. The proof of this section is in Appendix~\ref{sec:appendixa}.

\begin{definition}
\label{def:main_unify_momentum}
The unified momentum update rule is defined as:
\begin{equation}
    m_t = \beta m_{t-1} + \mu g_t, \quad \theta_t = \theta_{t-1} - \alpha m_t,
\end{equation}
     where \( \beta \in [0, 1) \) represents the decay or momentum factor, \( \mu \ge 1-\beta\) is a scaling parameter controlling the gradient contribution. \(g_t = \nabla f_t + \epsilon_t, \epsilon_t \sim \mathcal{N}(0, \sigma^2)\). Specific cases include:
\begin{itemize}
    \item \( \mu = 1 - \beta \): Exponential Moving Average (EMA),
    \item \( \mu = 1 \): Classical Momentum (CM)\cite{polyak1964some,sutskever2013importance}.
\end{itemize}
\end{definition}

This formulation encapsulates EMA and CM, two cornerstones of gradient estimation, differing in how they weight the current gradient against historical trends. EMA through a balanced mean update, while CM aggressively incorporates gradient direction. We dissect the nature of the two methods, quantified by the mean square error.

\begin{lemma}
\label{lem:main_bias_variance}
For any gradient estimator $\hat{g}_t = \mathcal{A}(g_1,...,g_t)$, the estimation mean square error decomposes as:
\begin{equation}
    \mathbb{E}[(\hat{g}_t - \nabla f_t)^2] = \underbrace{(\mathbb{E}[\hat{g}_t] - \nabla f_t)^2}_{\mathrm{Bias}^2} + \underbrace{(\hat{g}_t - \mathbb{E}[\hat{g}_t])^2}_{\mathrm{Variance}}
\end{equation}

\end{lemma}

Lemma~\ref{lem:main_bias_variance} establishes that the error in gradient estimation arises from two sources: bias, reflecting systematic deviation from the true gradient, and variance, capturing sensitivity to stochastic fluctuations. To explore how EMA and CM navigate this trade-off, we extend prior work on stochastic differential equations (SDEs) for vanilla SGD~\cite{stephan2017stochastic, li2019stochastic}, reformulating momentum in continuous time.

\begin{theorem}
\label{thm:main_bias_variance_momentum}
We reformulate the momentum term as an SDE. This reformulation incorporates \(\gamma\) to modulate the effective learning rate \(\alpha\), thereby facilitating a continuous-time representation of the optimization process. Assuming that the gradient \(\nabla f(\theta(t))\) is bounded and Lipschitz continuous, we derive the upper bound of the bias and variance of the momentum estimator \(m(t)\) as follows:

\begin{itemize}
\item \textbf{Bias}:
\begin{equation}
\begin{aligned}
     \left\| \mathrm{Bias}(m(t)) \right \|^2 &\leq \left( \frac{\mu \alpha L G }{(1 - \beta)^2} + \left( \frac{\mu}{1 - \beta} - 1 \right) \cdot G \right)^2,
\end{aligned}
\end{equation}

where \(L\) is the Lipschitz constant, \(G\) bounds \(\|\nabla f(\theta(t))\|\).
\item \textbf{Variance}:
\begin{equation}
	\mathrm{Var}(m(t)) \leq \frac{\mu^2 }{2 (1 - \beta)} \cdot \left( \sigma^2 +G^2\right). 
\end{equation}
where \(\sigma\) is the variance of random gradient sampling, \(G^2\) bounds \(\mathrm{Var}(\nabla f(\theta(t)))\) denote gradient estimates the variance of the sequence.
\end{itemize}
\end{theorem}

Theorem~\ref{thm:main_bias_variance_momentum} quantifies the bias-variance trade-off inherent in the unified momentum framework. For EMA (\( \mu = 1 - \beta \)), the bias simplifies to \(\left( \frac{\alpha L G}{1 - \beta} \right)^2\), scaling inversely with the momentum decay \(1 - \beta\), while the variance reduces to \(\frac{1 - \beta}{2} (\sigma^2 + G^2)\), shrinking as \(\beta\) approaches $1$. The variance-reducing property of EMA is similar to that of a low-pass filter in traditional signal processing to reduce noise~\cite{cutkosky2020momentum}. In statistical signal processing, which aligns closely with real optimization scenarios, gradients are non-stationary signals. Here, the estimator heavily weights early gradients, inflating bias while suppressing noise sensitivity.

Conversely, for CM (\( \mu = 1 \)), the bias bound becomes \(\left( \frac{\alpha L G}{(1 - \beta)^2} + \frac{\beta G}{1 - \beta} \right)^2\), and the variance scales as \(\frac{1}{2 (1 - \beta)} (\sigma^2 + G^2)\), both diverging as \(\beta \to 1\). This underscores CM’s aggressive momentum, which amplifies both systematic lag (bias) and noise susceptibility (variance) under strong momentum. The working mechanism of CM is too complicated, and we will discuss it further in Appendix~\ref{sec:appendix_discuss}.

Consider the extremes: when \(\beta = 0\), both EMA and CM reduce to vanilla SGD, yielding zero bias (assuming \(\mathbb{E}[g_t] = \nabla f_t\)) but retaining full variance \(\frac{1}{2} (\sigma^2 + G^2)\). As \(\beta \to 1\), EMA’s variance collapses, yet its bias grows unbounded, while CM’s estimator fixates on initial gradients, driving both metrics to infinity. These bounds illustrate a critical limitation: static choices of \(\mu\) and \(\beta\) lock the estimator into a fixed trade-off, ill-suited to the dynamic noise and curvature of real objectives. 

\subsection{Visualization of Gradient Distribution}
To better observe the effect of static momentum coefficients on the gradient estimation, while comparing our time-varying SGDF. We use VGG~\cite{2014Very} because it is a very standard network with no modules that interfere with the gradient, allowing for a better representation of the optimizer's update mechanism. We trained it with different SGD-based methods: Vanilla SGD, SGD with EMA, SGD with Wiener Filter, and SGD with CM. Then, we plot convergence curve in Fig~\ref{fig:accuracy} and use kernel density estimates of gradient values distribution over the first 100 iterations in Fig.~\ref{fig:histogram}. 

From Fig.~\ref{fig:accuracy}, applying SGD with EMA and Wiener Filter, convergence is faster than vanilla SGD. EMA has less fluctuation in test curves. WF demonstrates higher test accuracy with the same training set accuracy and reduced generalization gap. On the other hand, CM is slow to converge and results fluctuate because of the larger bias and variance.

Fig.~\ref{fig:histogram}\subref{subfig:SGD} shows high variance and uneven gradient values distribution in Vanilla SGD, resulting in training oscillations that hinder stable convergence. In contract, Fig.~\ref{fig:histogram}\subref{subfig:SGD-EMA} and Fig.~\ref{fig:histogram}\subref{subfig:SGDF} shows concentrated gradient distribution and not distorted. Especially, Fig.~\ref{fig:histogram}\subref{subfig:SGD-CM} shows that SGD-CM smooths values fluctuations but introduces \textit{gradient shift}, causing bias and variance over time. Previous research highlights that momentum struggle to adapt to variations in the curvature of the objective function, potentially causing deviation in updates~\cite{zhang2017yellowfin, dozat2016incorporating}. 

These analysis reveals a critical insight: Momentum methods suffer from the dilemma of bias and variance. Reducing variance amplifies bias, and reducing bias reduces variance. Can we design an adaptive gain that, at low variance, reduces the dependence on momentum to reduce the bias and at high variance, use the momentum update to reduce the variance?

\section{Method}
The analysis in the previous section suggests that a variable gain could be used to strike a balance between bias and variance. Previous work has introduced Kalman Filter~\cite{Kalman1960}, which uses time-varying gains to estimate gradients. However, the Kalman Filter's reliance on prior settings adds hyperparameter complexity. We then considered Wiener Filter~\cite{1950The}, which computes the gain in the frequency domain but requires the sequence to be stationary. The challenge is how to combine the advantages of time-variance and simplicity in a new algorithm. By leveraging the idea of minimizing mean square error~\cite{kay1993fundamentals} in Wiener Filter, we can degenerate Kalman Filter into a time-varying Wiener Filter, essentially a recursive least squares approach. In this section, we will introduce SGDF in detail.

\begin{algorithm}[htbp]
    \caption{SGDF, Wiener Filter Estimate Gradient. All operations are element-wise.}
    \label{alg:WienerOptimization}
    \KwIn{$\{\alpha_t\}^{T}_{t=1}$: step size, $\{\beta_1, \beta_2\}$: attenuation coefficient, $\theta_0$: initial parameter, $f(\theta)$: stochastic objective function}

    \KwOut{$\theta_{T}$: resulting parameters.}  

    Init: $m_0 \leftarrow 0$, $s_0 \leftarrow 0$

    \While{\textnormal{$t=1$ to $T$}}{
        $g_{t} \leftarrow \nabla f_t(\theta_{t-1}) $ 

        $m_{t} \leftarrow \beta_1  m_{t-1} + (1 - \beta_1)  g_t$ 

        $s_{t} \leftarrow \beta_2  s_{t-1} + (1 - \beta_2)  (g_t - m_t)^2$ 

        $\widehat{m}_t \leftarrow \dfrac{m_t}{1 - \beta_1^t}$, $\widehat{s}_t \leftarrow \dfrac{\textcolor{blue}{(1-\beta_1)(1-\beta_1^{2t})}s_t}{\textcolor{blue}{(1+\beta_1)}(1 - \beta_2^t)}$ 

        $K_{t} \leftarrow \dfrac{\widehat{s}_{t} }{\widehat{s}_t + (g_t - \widehat{m}_t)^2}$ 

        
        $\widehat{g}_t \leftarrow \widehat{m}_{t} + \textcolor{blue}{2} K_t  (g_t - \widehat{m}_{t})$ 

        $\theta_{t} \leftarrow \theta_{t-1} - \alpha_t \widehat{g}_t$ 
    }   
    return $\theta_{T}$
\end{algorithm}

\subsection{SGDF General Introduction}

In algorithm~\ref{alg:WienerOptimization}, $s_t$ serves as a key indicator, calculated as the exponential moving average of the squared difference between the current gradient $g_t$ and its momentum $m_t$, acting as a marker for gradient variation with weight-adjusted by $\beta_2$. 
\citep{zhuang2020adabelief} first proposed the calculation of $s_t$, which is utilized for estimating the fluctuation variance of the stochastic gradient. We derived a correction factor \textcolor{blue}{$(1-\beta_1)(1-\beta_1^{2t})/(1+\beta_1)$} under the assumption that $m_t$ and $g_t$ are independently and identically distributed (i.i.d.), to accurately estimate the variance of $m_t$ using $s_t$. Appendix~\ref{sec:appendixe} Fig.~\ref{fig:correction} compares performances with and without the correction factor, showing superior results with correction. For the derivation of the correction factor, please refer to Appendix~\ref{proof_correction}. In the case of maximum divergence, where gradients reach their upper bound, the gain \( K_t \) approximates \( \frac{1}{2} \). Scaling \( K_t \) by \textcolor{blue}{$2$} in SGDF enhances experimental performance.


%

At each time step $t$, $g_t$ represents the stochastic gradient for our objective function, while $m_t$ approximates the historical trend of the gradient through an exponential moving average. The difference $g_t - m_t$ highlights the gradient's deviation from its historical pattern, reflecting the inherent noise or uncertainty in the instantaneous gradient estimate, which can be expressed as $p(g_{t} | \mathcal{D}) \sim \mathcal{N}(g_{t};m_{t}, \sigma_{t}^{2})$~\citep{bernstein2018signsgd,liu2019variance}.

SGDF utilizes the gain $K_t$, where the components of each dimension of the estimated gain range between 0 and 1, to balance the current observed gradient $g_t$ and the past corrected gradient $\widehat{m}_{t}$, thus optimizing the gradient estimate. This balance plays a crucial role in noisy or complex optimization scenarios, helping to mitigate noise and achieve stable gradient direction, faster convergence, and enhanced performance. The computation of $K_t$, based on $s_t$ and $g_t - m_t$, aims to minimize the expected variance of the corrected gradient $\widehat{g}_t$ for optimal linear estimation in noisy conditions. For the method derivation, please refer to Appendix~\ref{proof_method}.

\subsection{Fusion of Gaussian Distributions}
\label{sec:section3.2}
In effect, balancing the bias and variance is re-estimating the mean and variance of the gradient distribution as a new gradient distribution. The properties of SGDF ensure that the estimated gradient \(\widehat{g}_t\) is a linear combination of the current noisy gradient observation \(g_{t}\) and the first-order moment estimate \(\widehat{m}_{t}\). Both components are assumed to follow Gaussian distributions, allowing the Wiener Filter to fuse them optimally, resulting in \(\widehat{g}_t\) as a Gaussian distribution.

Consider the Gaussian distributions for the momentum term \(\widehat{m}_t\) and the current gradient \(g_t\):
\begin{itemize}
	\item The exponential moving average term \(\widehat{m}_t\) is normally distributed with mean \(\mu_{m}\) and variance \(\sigma_{m}^{2}\), denoted as \(\widehat{m}_t \sim \mathcal{N}(\mu_{m}, \sigma_{m}^{2})\).
	\item The current gradient \(g_t\) is normally distributed with mean \(\mu_{g}\) and variance \(\sigma_{g}^{2}\), denoted as \(g_t \sim \mathcal{N}(\mu_{g}, \sigma_{g}^{2})\).
\end{itemize}

The product of probability density functions is given by:
\begin{multline}
    N(\widehat{m}_t; \mu_{m}, \sigma_{m}) \cdot N(g_t; \mu_{g}, \sigma_{g}) = \\
      \frac{1}{2\pi\sigma_{m}\sigma_{g}} \exp\left(-\frac{(\widehat{m}_t-\mu_{m})^2}{2\sigma_{m}^2} 
    - \frac{(g_t-\mu_{g})^2}{2\sigma_{g}^2}\right)
\end{multline}

By matching coefficients in the exponential terms, we obtain the new mean \(\mu_{\widehat{g}_t}\) and variance \(\sigma_{\widehat{g}_t}^2\) for the fused Gaussian distribution:
\begin{equation}
    \mu_{\widehat{g}_t} = \frac{\sigma_{g}^2 \mu_{m} + \sigma_{m}^2 \mu_{g}}{\sigma_{m}^2 + \sigma_{g}^2}, \quad \sigma_{\widehat{g}_t}^2 = \frac{\sigma_{m}^2 \sigma_{g}^2}{\sigma_{m}^2 + \sigma_{g}^2}.
\end{equation}



The fused mean \(\mu_{\widehat{g}_t}\) is a weighted average of \(\mu_{m}\) and \(\mu_{g}\), with weights inversely proportional to their variances, favoring the mean with the smaller variance to reflect greater confidence in stable estimates. Similarly, the fused variance \(\sigma_{\widehat{g}_t}^2\) is smaller than the original variances \(\sigma_{m}^2\) and \(\sigma_{g}^2\), indicating reduced uncertainty in the gradient estimate. This reduction is a result of the Wiener Filter's optimality in minimizing mean-square error. The proof is in Appendix~\ref{proof_section3.2}.

\subsection{Convex and Non-convex Convergence Analysis}
\label{sec:section3.4}
Finally, we provide the convergence property of SGDF as shown in Theorem~\ref{main_convex} and Theorem~\ref{main_non_convex}. The assumptions are common and standard when analyzing the convergence of convex and non-convex functions via SGD-based methods~\cite{chen2018convergence, kingma2014adam,reddi2018convergence}. Proofs for convergence in convex and non-convex cases are provided in Appendix~\ref{convex} and Appendix~\ref{non-convex}, respectively. 


\begin{theorem}{(Convergence in convex optimization)}
	\label{main_convex}
	Assume that the function \( f_t \) has bounded gradients, \( \Vert\nabla f_t(\theta)\Vert_2 \leq G \), \( \Vert\nabla f_t(\theta)\Vert_\infty \leq G_\infty \) for all \( \theta \in \mathbb{R}^d \) and distance between any \( \theta_t \) generated by SGDF is bounded, \( \Vert\theta_n - \theta_m\Vert_2 \leq D \), \( \Vert\theta_m - \theta_n\Vert_\infty \leq D_\infty \) for any \( m, n \in \{1, ..., T\} \), and \( \beta_1, \beta_2 \in [0, 1) \). Let \( \alpha_t = \alpha/\sqrt{t} \). SGDF achieves the following guarantee, for all \( T \geq 1 \):

\begin{multline}
    R(T) \leq \frac{D^2}{\alpha} \sum_{i=1}^{d}  \sqrt{T} + 
    \frac{2 D_\infty G_{\infty}}{1-\beta_{1}} \sum_{i=1}^{d}\left\Vert g_{1: T, i} \right\Vert_{2} \\
    +\frac{2\alpha G_{\infty}^{2}(1 + (1 - \beta_{1})^2)}{\sqrt{T}(1-\beta_{1})^{2}} 
    \sum_{i=1}^{d} \left\Vert g_{1: T, i} \right\Vert_{2}^{2}
\end{multline}

 where $R(T) = \sum_{t=1}^{T} f_t(\theta_t) - f_t(\theta^* )$ denotes the cumulative performance gap between the generated solution and the optimal solution.
\end{theorem}

For the convex case, Theorem~\ref{main_convex} implies that the regret of SGDF is upper bounded by $O(\sqrt{T}) $. In the Adam-type optimizers, it's crucial for the convex analysis to decay $\beta_{1, t}$ towards zero~\citep{kingma2014adam,zhuang2020adabelief}. We have relaxed the analysis assumption by introducing a time-varying gain $K_t$, which can adapt with variance.

    
    


\begin{theorem}{(Convergence for non-convex stochastic optimization)}
\label{main_non_convex}
Consider a non-convex optimization problem. Suppose Assumptions~\ref{nonassume} in Appendix~\ref{non-convex} is satisfied, and let \( \alpha_t = \alpha/\sqrt{t} \). For all \( T \geq 1 \), SGDF achieves the following guarantee:
\begin{equation}
    \mathbb{E}(T) \leq \frac{C_{7}\alpha^2 (\log T + 1) + C_{8}}{2\alpha\sqrt{T}}
\end{equation}

where $\mathbb{E}(T) =\min_{t=1,2,\ldots,T}\mathbb{E}_{t-1}\left[\left\Vert \nabla f\left(\theta_{t}\right)\right\Vert _{2}^{2}\right]$ denotes the minimum of the squared-paradigm expectation of the gradient, $\alpha$ is the learning rate at the $1$-th step, $C_{7}$ are constants independent of $d$ and $T$, $C_{8}$ is a constant independent of $T$, and the expectation is taken w.r.t all randomness corresponding to ${g_{t}}$.
\end{theorem}

Theorem~\ref{main_non_convex} indicates that the convergence rate for SGDF in the non-convex case is $O(\log T / \sqrt{T})$, which is comparable to Adam-type optimizers~\citep{chen2018convergence, reddi2018convergence}. In our derivation, the terms related to the estimated gain $K_t$ were scaled to their maximum upper bounds, simplifying the upper bound results. Importantly, we did not rely on the $\mu$-strongly convex assumption~\citep{balles2018dissecting} but used the most general L-smoothness assumption to obtain this convergence rate. 


\begin{figure*}[t]
    \vspace{-3mm}
    \centering
    \setlength{\tabcolsep}{0pt}
    \begin{tabular}{ccc}
        \subfloat[VGG11 on CIFAR-10]{\includegraphics[width=0.33\linewidth]{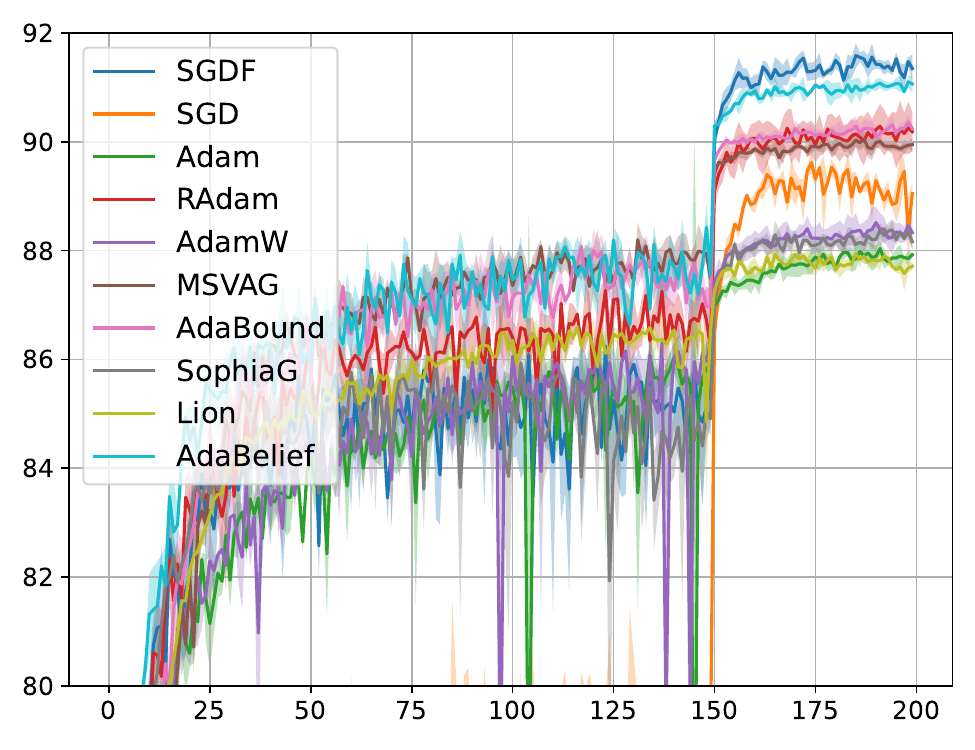}\label{subfig:main_vgg_test_cifar10}} &
        \subfloat[ResNet34 on CIFAR-10]{\includegraphics[width=0.33\linewidth]{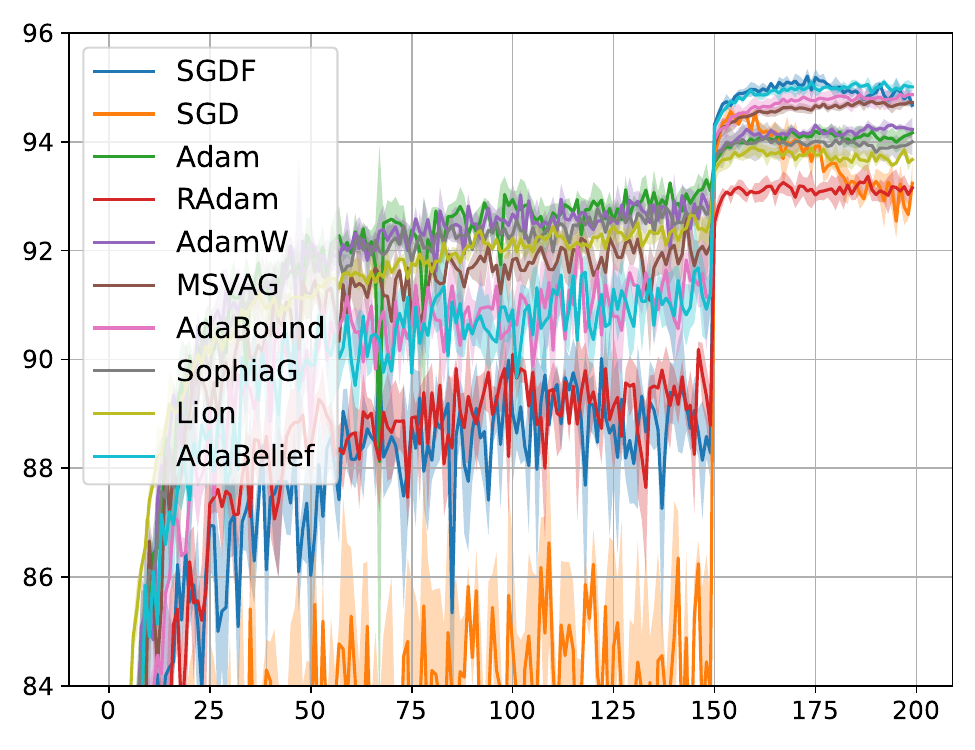}\label{subfig:main_resnet_test_cifar10}} &
        \subfloat[DenseNet121 on CIFAR-10]{\includegraphics[width=0.33\linewidth]{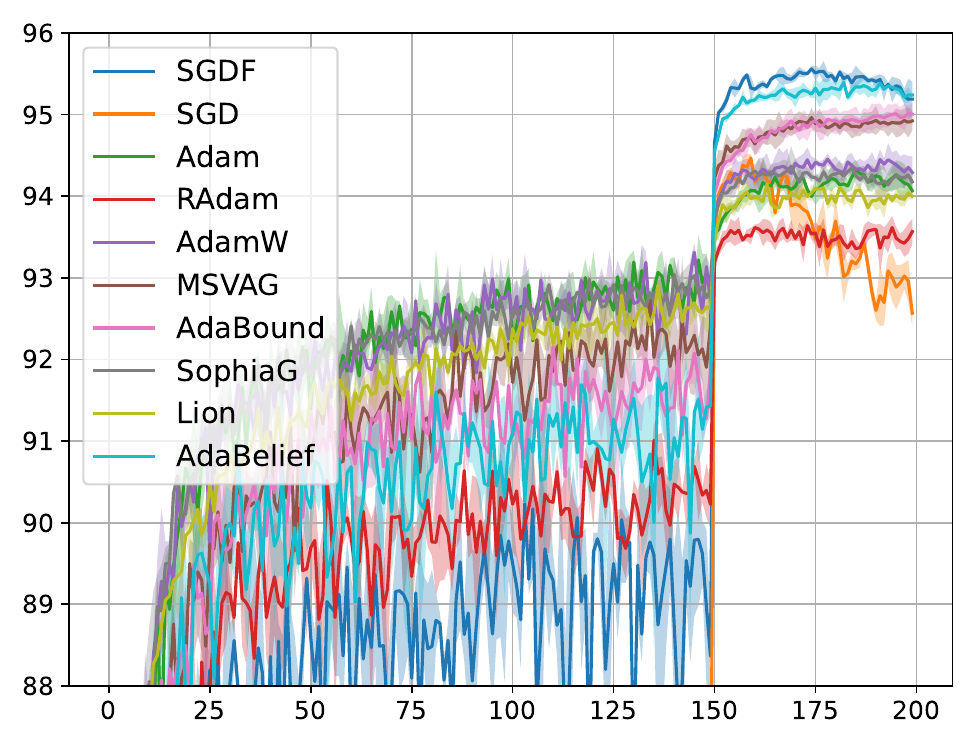}\label{subfig:main_densenet_test_cifar10}} \\
        
        \subfloat[VGG11 on CIFAR-100]{\includegraphics[width=0.33\linewidth]{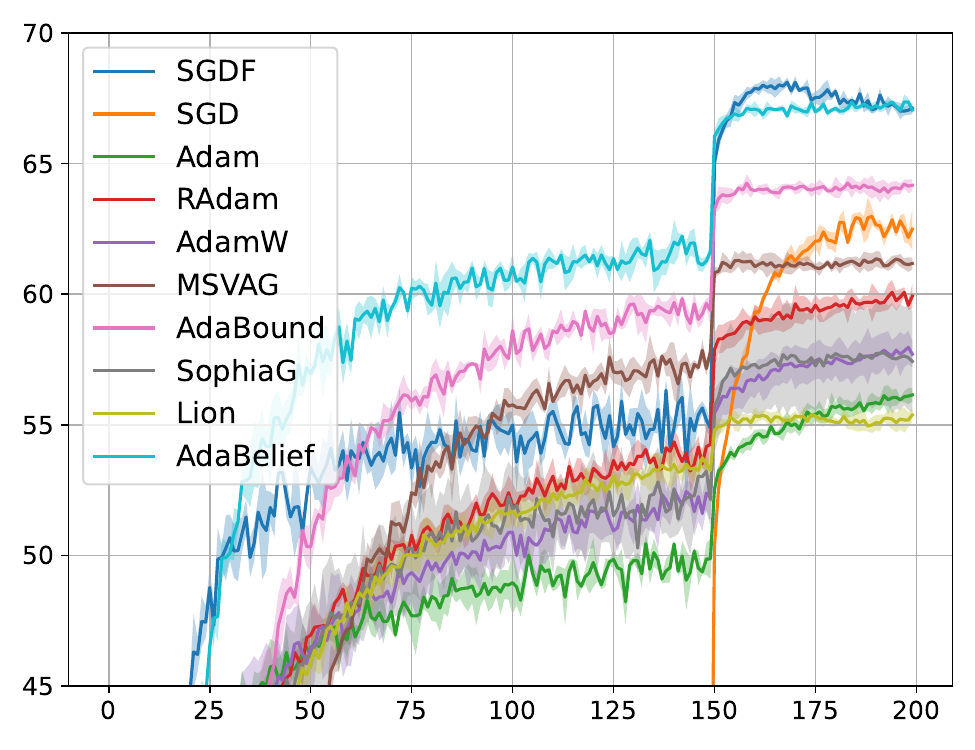}\label{subfig:main_vgg_test_cifar100}} &
        \subfloat[ResNet34 on CIFAR-100]{\includegraphics[width=0.33\linewidth]{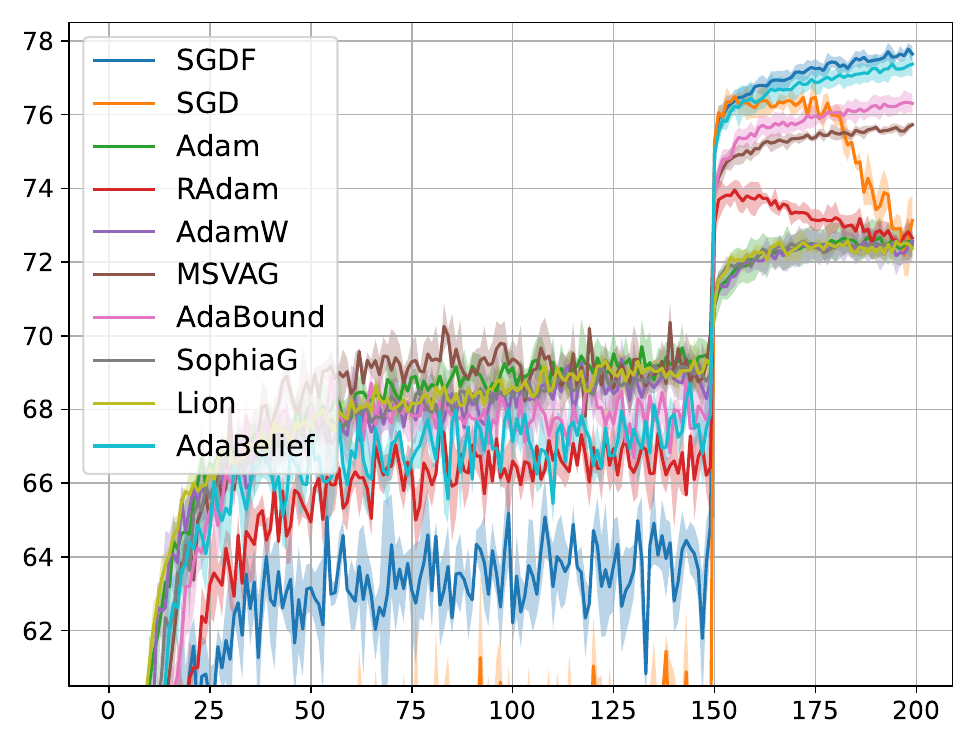}\label{subfig:main_resnet_test_cifar100}} &
        \subfloat[DenseNet121 on CIFAR-100]{\includegraphics[width=0.33\linewidth]{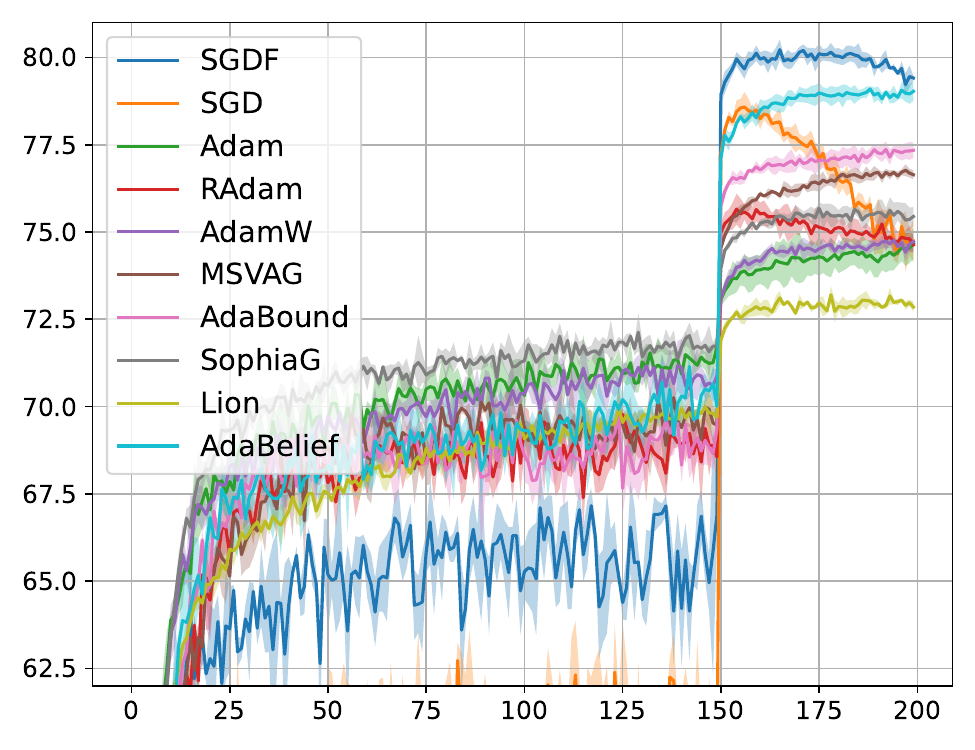}\label{subfig:main_densenet_test_cifar100}}
    \end{tabular}
    \caption{Test accuracy ([$\mu \pm \sigma$]) on CIFAR.}
    \label{fig:main_cifar_experiments}
    \vspace{-3mm}
\end{figure*}


\section{Experiments}
\label{sec:section4}
\subsection{Empirical Evaluation}
\label{sec:section4.1}
In this study, we focus on the following tasks: \textbf{Image Classification.} We employed the VGG~\cite{2014Very}, ResNet~\cite{he2016deep}, and DenseNet~\cite{huang2017densely} models for image classification tasks on the CIFAR dataset~\cite{krizhevsky2009learning}. The major difference between these three network architectures is the residual connectivity, which we will discuss in Sec.~\ref{sec:section4.4}. We evaluated and compared the performance of SGDF with other optimizers such as SGD, Adam, RAdam~\cite{liu2019variance}, AdamW~\cite{loshchilov2017decoupled}, MSVAG~\cite{balles2018dissecting}, Adabound~\cite{luo2019adaptive}, Sophia~\cite{liu2023sophia}, Lion~\cite{chen2023symbolic}, and AdaBelif~\cite{zhuang2020adabelief}, all of which were implemented based on the official PyTorch. Additionally, we further tested the performance of SGDF on the ImageNet dataset~\cite{Deng2009} using the ResNet model. \textbf{Object Detection.} Object detection was performed on the PASCAL VOC dataset~\cite{pascal-voc-2010} using Faster-RCNN~\cite{faster_rcnn} integrated with FPN. For hyper-parameter tuning related to image classification and object detection, refer to~\citep{zhuang2020adabelief}. \textbf{Fine-tuning in ViT.} We test the performance of transformer architecture networks by fine-tuning ViT~\cite{dosovitskiy2020image} on six benchmark dataset. \textbf{More experimental results are summarized in Appendix~\ref{sec:appendixe}.}

\textbf{Hyperparameter tuning.} Following Zhuang~\etal~\cite{zhuang2020adabelief}, we delved deep into the optimal hyperparameter settings for our experiments. In the image classification task, we employed these settings:
\begin{itemize}[leftmargin=*]
\item \textit{SGDF:} We adhered to Adam's original parameter values except learning rate: $\alpha=0.5$, $\beta_1=0.9$, $\beta_2=0.999$, $\epsilon=10^{-8}$. The learning rate was searched same as SGD research set.
\item \textit{SGD:} We set the momentum 0.9, which is the default for networks like ResNet and DenseNet. The learning rate was searched in the set \{10.0, 1.0, 0.5, 0.1, 0.01, 0.001\}.
\item \textit{Adam, RAdam, MSVAG, AdaBound, AdaBelief:} Traversing the hyperparameter landscape, we scoured \(\beta_{1} \) values in \(\{0.5,0.6,0.7,0.8,0.9\}\), probed \(\alpha \) as in SGD, while tethering other parameters to their literary defaults.
\item \textit{AdamW, SophiaG, Lion:} Mirroring Adam's parameter search schema, we fixed weight decay at \(5 \times 10^{-4}\); yet for AdamW, whose optimal decay often exceeds norms~\citep{loshchilov2017decoupled}, we ranged weight decay over \(\left\{10^{-4}, 5 \times 10^{-4}, 10^{-3}, 10^{-2}, 10^{-1}\right\}\).
\item \textit{SophiaG, Lion:} We searched for the learning rate among $\{10^{-3}, 10^{-4}, 10^{-5}\}$ and adjusted Lion's learning rate~\cite{liu2023sophia}. Following ~\cite{liu2023sophia, chen2023symbolic}, we set $\beta_{1}$=0.965, 0.9 and $\beta_2$=0.99 as the default parameters.
\end{itemize}

\textbf{CIFAR-10/100 Experiments.} We trained on the CIFAR-10 and CIFAR-100 datasets using the VGG, ResNet, and DenseNet models and assessed the performance of the SGDF optimizer. In these experiments, we employed basic data augmentation techniques such as random horizontal flip and random cropping. The results represent the mean and standard deviation of 3 runs by fixing random seeds \{0, 1, 2\}, visualized as curve graphs in Fig.~\ref{fig:main_cifar_experiments}. To facilitate result reproduction, we provide the parameter table for this subpart in Appendix~\ref{sec:appendixe} Tab.~\ref{tab:cifarhyperparameters}.

As Fig.~\ref{fig:main_cifar_experiments} shows, that it is evident that the SGDF optimizer exhibited convergence speeds comparable to adaptive optimization algorithms. Additionally, SGDF's final test set accuracy was either better than others. We summarized the mean best test accuracies and their standard deviations for each algorithm in Appendix~\ref{sec:appendixe} Tab.~\ref{tab:cifar_values}.

\textbf{ImageNet Experiments.} We applied basic data augmentation strategies such as random cropping and random horizontal flipping~\cite{zhuang2020adabelief} and the random seed is set to 2025 same as the current year. Additionally, to mitigate the effect of learning rate scheduling, we employed cosine learning rate scheduling as suggested by~\cite{chen2023symbolic,Zhang2023}. We trained ResNet18 for 100 epochs aligned with~\cite{chen2018closing,zhuang2020adabelief} to compare with other popular optimizers by using the best-reported results from~\cite{chen2018closing,liu2019variance,zhuang2020adabelief}. We also trained more model architectures for 90 epochs~\cite{Zhang2023} to compare with SGD. For SGD, we used the results reported by \textit{PyTorch formal pre-trained models}\footnote{\textcolor{magenta}{https://pytorch.org/vision/main/models.html\#table-of-all-available-classification-weights}}. Detailed training and test curves are depicted in Appendix~\ref{sec:appendixe} Fig.~\ref{fig:imagenet_curve}. To facilitate result reproduction, we provide the parameter table for this subpart in Appendix~\ref{sec:appendixe} Tab.~\ref{tab:imagenethyperparameters}.  

The results are summarized in Tab.~\ref{tab:resnet18_imagenet} and~\ref{tab:more_imagenet}. Experiments on the ImageNet dataset demonstrate that SGDF has improved convergence speed and achieves superior accuracy compared to SGD on the test set.

\begin{table*}[t]
       \vspace{-2mm}
	\centering
	\caption{Top-1, 5 accuracy of ResNet18 on ImageNet. $^{*}$ \ $^{\dagger}$ \ $^{\ddagger}$ is reported in~\cite{zhuang2020adabelief, chen2018closing, liu2019variance}.}
    \resizebox{1.0\linewidth}{!}{
	\begin{tabular}{c|cccccccccc}
		\toprule 
		Method & SGDF & SGD & AdaBelief & PAdam &AdaBound & Yogi & MSVAG & Adam & RAdam & AdamW \\
		\midrule 
		Top-1 & \textbf{70.55} & 70.23$^{\dagger}$ & 70.08$^{*}$& 70.07$^{\dagger}$ & 68.13$^{\dagger}$ & 68.23$^{\dagger}$ & 65.99$^{*}$ & 63.79$^{\dagger}$ (66.54$^{\ddagger}$) & 67.62$^{\ddagger}$ & 67.93$^{\dagger}$ \\
		Top-5 & \textbf{89.76} & 89.40$^{\dagger}$& -& 89.47$^{\dagger}$ &88.55$^{\dagger}$& 88.59$^{\dagger}$ & - & 85.61$^{\dagger}$& - & 88.47$^{\dagger}$ \\
		\bottomrule 
	\end{tabular}
    }
	\label{tab:resnet18_imagenet}
       \vspace{-1mm}
\end{table*}

\begin{table*}[htbp]
        \vspace{-1mm}
	\centering
	\caption{More model results to compare with SGD in ImageNet.} 
    \resizebox{0.75\linewidth}{!}{
	\begin{tabular}{c|ccccccc}
		\toprule 
		Model & VGG11\_BN & VGG13\_BN &  ResNet34 & ResNet50 & DenseNet121 & DenseNet 161 \\
		\midrule 
		SGD &  70.37 & 71.58 & 73.31 & 76.13 & 74.43 & 77.13\\		
		\midrule
            SGDF & \textbf{71.47} & \textbf{72.54} & \textbf{74.03} & \textbf{76.65} & \textbf{75.78} & \textbf{78.44}\\
		\bottomrule 
	\end{tabular}
    }
	\label{tab:more_imagenet}
	\vspace{-3mm}
\end{table*}

\textbf{Object Detection.} We conducted object detection experiments on the PASCAL VOC dataset~\citep{pascal-voc-2010}. The model used in these experiments was pre-trained on the COCO dataset~\citep{coco-dataset}, obtained from the official website. We trained this model on the VOC2007 and VOC2012 trainval dataset (17K) and evaluated it on the VOC2007 test dataset (5K). The utilized model was Faster-RCNN~\cite{faster_rcnn} with FPN, and the backbone was ResNet50~\citep{he2016deep}. Results are summarized in Tab.~\ref{tab:object_detection}. To facilitate result reproduction, we provide the parameter table for this subpart in Appendix~\ref{sec:appendixe} Tab.~\ref{tab:hyperparameters_object_detection}. As expected, SGDF outperforms other methods. These results also illustrate the efficiency of our method in object detection tasks.

\begin{table}[htbp]
\centering
\caption{The mAP on PASCAL VOC using Faster-RCNN+FPN. $^{*}$ \ $^{\dagger}$ is reported in~\cite{zhuang2020adabelief, yuan2020eadam}.}
\resizebox{1.0\linewidth}{!}{
\begin{tabular}{c|ccccccc}
    \toprule 
    Method & SGDF & AdaBelief & EAdam & SGD & Adam & AdamW & RAdam \\
    \midrule 
    mAP & \textbf{83.81} & 81.02$^{*}$ & 80.62$^{\dagger}$ & 80.43 & 78.67 & 78.48 & 75.21 \\
    \bottomrule 
\end{tabular}
\label{tab:object_detection}
}
     \vspace{-2mm}
\end{table}

\textbf{Fine-tuning in ViT.} To evaluate SGDF's performance, we used Vision Transformers (ViT)~\cite{dosovitskiy2020image} on six benchmark datasets: CIFAR-10, CIFAR-100, Oxford-IIIT-Pets~\cite{parkhi2012cats}, Oxford Flowers-102~\cite{nilsback2008automated}, Food101~\cite{bossard2014food}, and ImageNet-1K. Two ViT variants, ViT-B/32 and ViT-L/32, pre-trained on ImageNet-21K, were selected. For fine-tuning, we replaced the original MLP classification head with a new fully connected layer, tailored to the dataset categories. All Transformer backbone weights were retained, preserving the rich representations learned from ImageNet-21K. We increased the image resolution from $224 \times 224$ to $384 \times 384$ to improve accuracy while adjusting the position encoding through 2D interpolation to match the new resolution. For optimization, SGDF was compared to SGD with momentum as a baseline, using cosine learning rate decay and no weight decay. A batch size of 512 and global gradient clipping (norm of 1) were used to prevent gradient explosion. All experiments were trained uniformly for 10 epochs and the random seed is set to 2025. Results are summarized in Table~\ref{tab:vitresult}. We summarized the hyperparameter in Appendix~\ref{sec:appendixe} Tab.~\ref{tab:vithyperparameters}.

\begin{table*}[htp]
\vspace{-4mm}
	\centering
	\caption{Fine-tuning in ViT. Train for 10 epochs and report Top-1 accuracy.}
	\resizebox{0.9\linewidth}{!}{
	\begin{tabular}{@{}lccccccc@{}}
		\toprule
		Model & Method & CIFAR-10 & CIFAR-100 & Oxford-IIIT-Pets & Oxford Flowers-102 & Food101 & ImageNet \\
		\midrule
		\multirow{2}{*}{ViT-B/32}  & SGD  & 98.60 & 89.72  & 90.26 & 96.71 & 87.79 & 81.30 \\
          & SGDF & \textbf{98.64} & \textbf{90.77}   & \textbf{92.34} & \textbf{96.92} & \textbf{88.68} & \textbf{81.40} \\
		\midrule
		\multirow{2}{*}{ViT-L/32} & SGD & 98.74  & 91.51 & 86.09 & 96.68 & 89.23 & 81.21 \\
        & SGDF & \textbf{98.87} & \textbf{92.14}   & \textbf{91.98}  & \textbf{97.02} & \textbf{90.05} & \textbf{81.31}\\
		\bottomrule
	\end{tabular}
	}
	\label{tab:vitresult}
\end{table*}

\subsection{Top Eigenvalues of Hessian and Hessian Trace}
\begin{figure}[htbp] 
    \centering
    \begin{subfigure}[t]{0.48\linewidth}
        \centering
        \begin{overpic}[width=\linewidth]{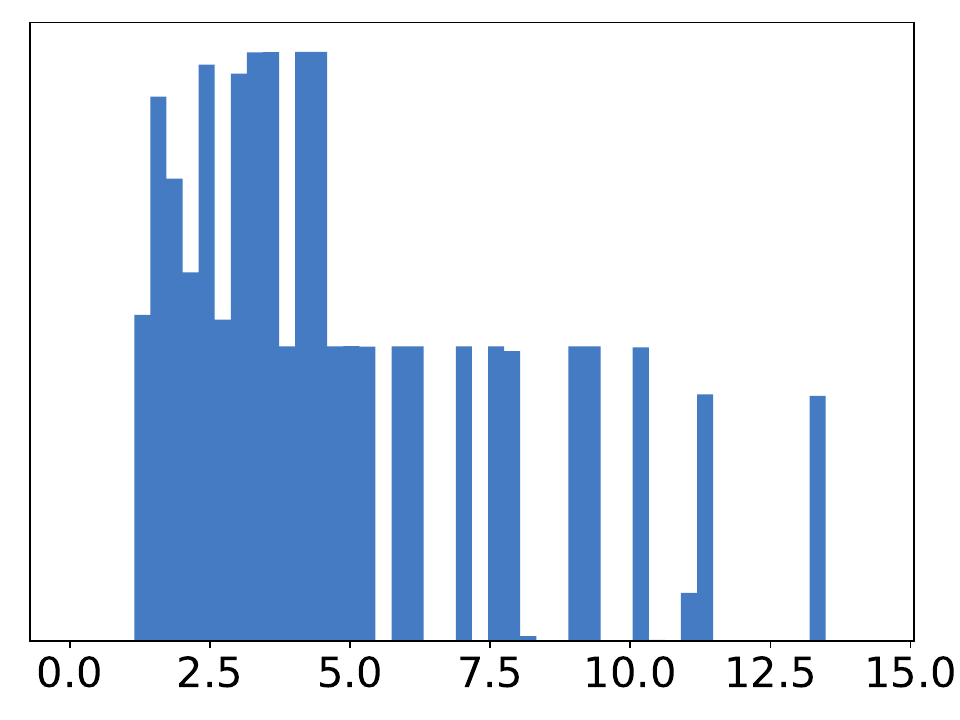}
            \put(60,65){\scriptsize Trace: 192.47 }
            \put(64,55){\scriptsize $\lambda_{\text{max}}$: 13.32}
        \end{overpic}
        \caption{SGDF}
        \label{subfig:main_hessian_sgdf}
    \end{subfigure}%
    \begin{subfigure}[t]{0.48\linewidth}
        \centering
        \begin{overpic}[width=\linewidth]{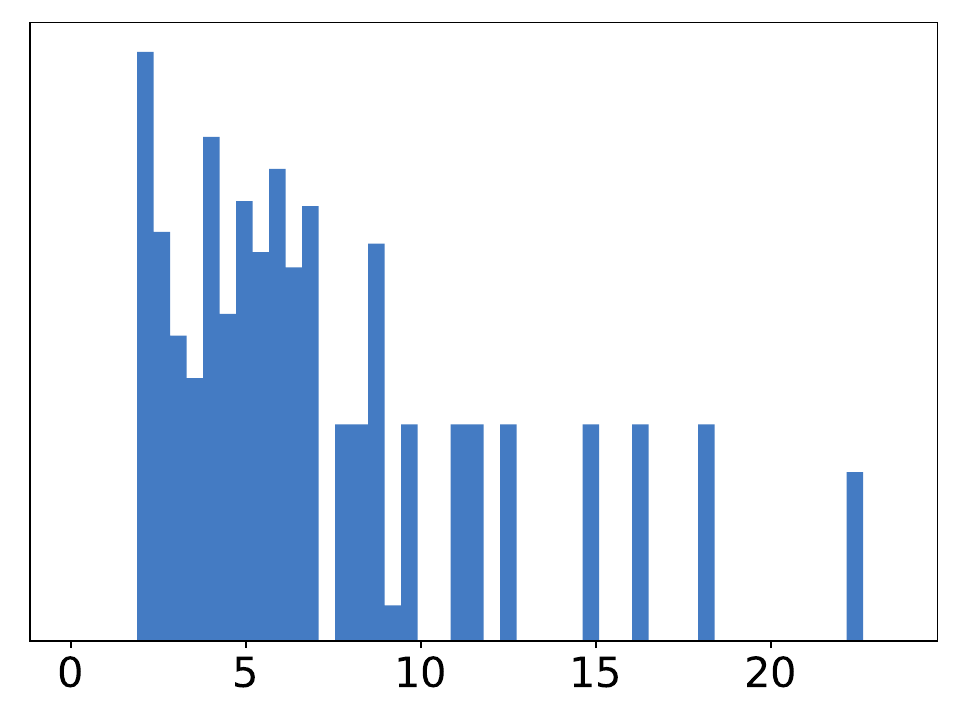}
            \put(62,65){\scriptsize Trace: 419.30}
            \put(67,55){\scriptsize $\lambda_{\text{max}}$: 22.51}
        \end{overpic}
        \caption{SGD}
        \label{subfig:main_hessian_sgdm}
    \end{subfigure}%

    \vspace{1em} 
    \begin{subfigure}[t]{0.48\linewidth}
        \centering
        \begin{overpic}[width=\linewidth]{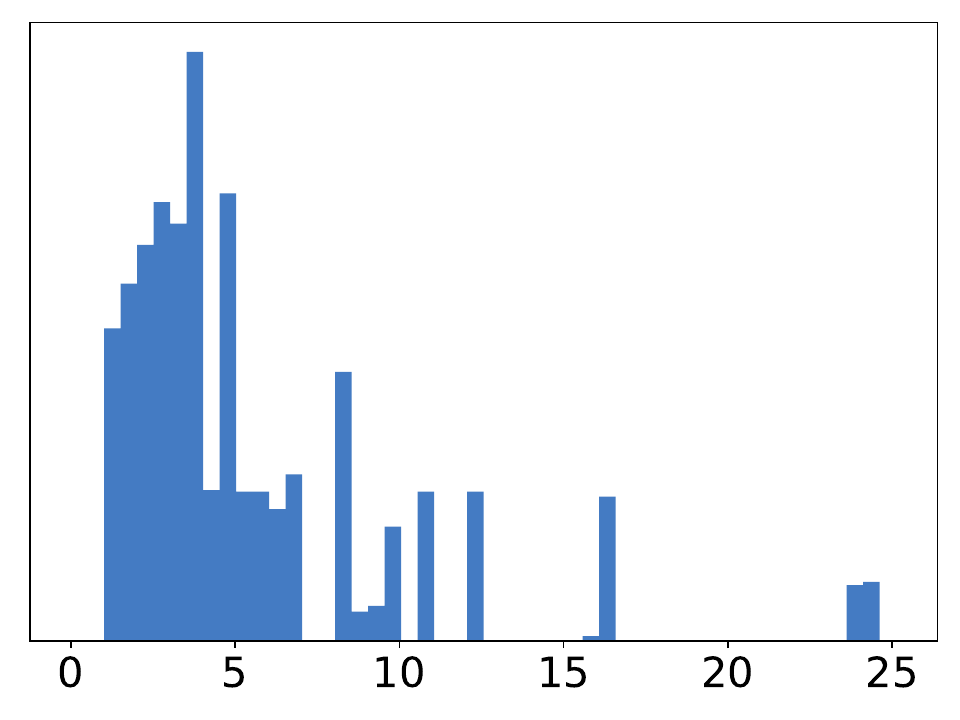}
            \put(62,65){\scriptsize Trace: 284.38}
            \put(67,55){\scriptsize $\lambda_{\text{max}}$: 24.11}
        \end{overpic}
        \caption{SGD-EMA}
        \label{subfig:main_hessian_sgd}
    \end{subfigure}%
    \begin{subfigure}[t]{0.48\linewidth}
        \centering
        \begin{overpic}[width=\linewidth]{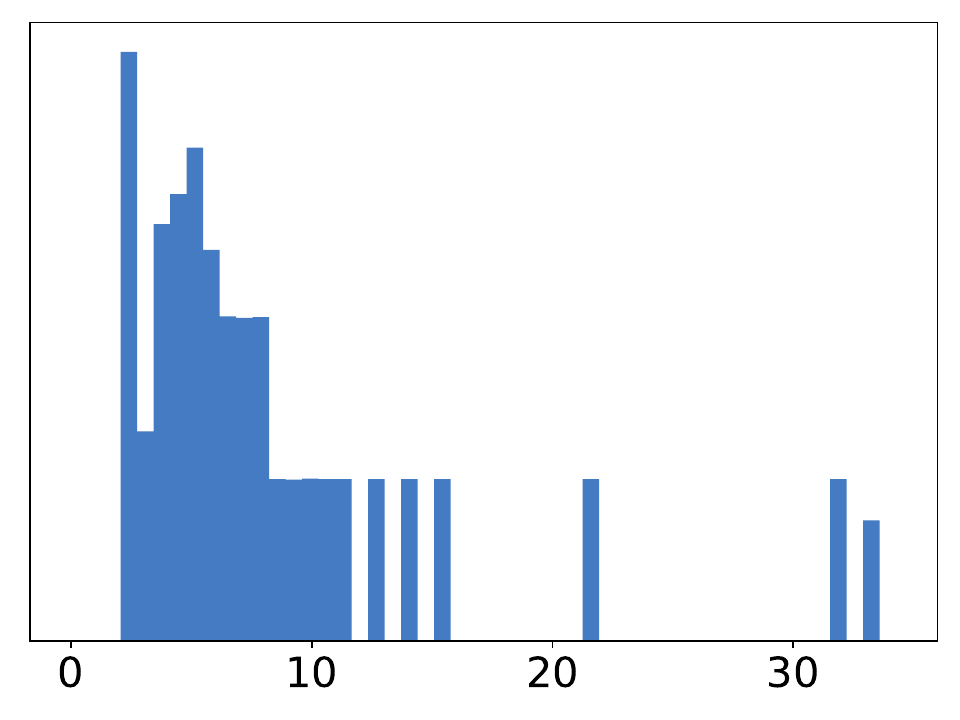}
            \put(62,65){\scriptsize Trace: 491.63}
            \put(67,55){\scriptsize $\lambda_{\text{max}}$: 32.61}
        \end{overpic}
        \caption{SGD-CM}
        \label{subfig:main_hessian_adam}
    \end{subfigure}
    \caption{Histogram of Top 50 Hessian Eigenvalues. Lower values indicate better performance on the test dataset.}
    \label{fig:main_hessian_spectrum}
\end{figure}

The success of optimization algorithms in deep learning depends on both minimizing training loss and the quality of the solutions they find. So we numerically verified the hessian matrix properties between the different methods. We computed the Hessian spectrum of ResNet-18 trained on the CIFAR-100 dataset for 200 epochs. These experiments ensure that all methods achieve similar results on the training set. We employed power iteration~\citep{2018Hessian} to compute the top eigenvalues of Hessian and Hutchinson’s method~\citep{2020PyHessian} to compute the Hessian trace. Histograms illustrating the distribution of the top 50 Hessian eigenvalues for each optimization method are presented in Fig.~\ref{fig:main_hessian_spectrum}. SGDF brings lower eigenvalue and trace of the hessian matrix, which explains the fact that SGDF demonstrates better performance than SGD as the categorization category increases.

\subsection{Wiener Filter combines Adam}
\label{sec:section4.4}
We've conducted comparative experiments on the CIFAR-100 dataset, evaluating both the vanilla Adam algorithm and Adam with Wiener Filter, which substitutes the first-moment gradient estimates in the Adam optimizer with Wiener Filter estimates. The results are presented in Tab.~\ref{tab:accuracy_comparison}, and the detailed test curves are depicted in Appendix~\ref{sec:appendixe} Fig.~\ref{fig:compare}. This suggests that our first-moment filter estimation method has the potential to be applied to other optimization methods.

 \begin{table}[htbp]
	\centering
	\caption{Accuracy comparison between Adam and Wiener-Adam.}
    \resizebox{0.8\linewidth}{!}{
	\begin{tabular}{c|cccc}
		\toprule
		Model & VGG11 & ResNet34 & DenseNet121  \\
		\midrule
		Filter-Adam & \textbf{62.64} & \textbf{73.98} & \textbf{74.89} \\
		\midrule
		Vanilla-Adam & 56.73 & 72.34 & \textbf{74.89}\\
		\bottomrule
	\end{tabular}
    }
	\label{tab:accuracy_comparison}
\end{table}
 
For VGG without BN, the Wiener Filter significantly improves performance by providing more accurate gradient estimates, reducing noise-induced errors, and ultimately enhancing accuracy. In contrast, for ResNet and DenseNet, which already incorporate BN and leverage residual and dense connections to stabilize gradient flow, the benefits of the Wiener Filter are less pronounced. These architectures inherently promote stable gradient updates through their structural design, reducing the additional advantages offered by Wiener Filter. This explains why the performance improvements vary across different architectures, as seen in Tab.~\ref{tab:accuracy_comparison}. While Adam with Filter provides a notable boost in simpler architectures like VGG, its impact is diminished in more complex networks where existing mechanisms already aid gradient stability.
\section{Limitations and Future Work}
Our analysis is based on the commonly used L-smoothness assumption in non-convex cases, but not advanced enough. While some works~\cite{wang2022provable, wang2023convergence} analyze algorithms using gradient affine variance~\cite{zhang2019gradient}, which better bridges the gap between theory and practice, this is not the focus of this work. Furthermore, we initially attempted to extend the first-order filter estimation of SGDF to Adam inspired by bias and variance. This extension can improve performance in certain scenarios, but due to the more complex gradient estimation in Adam, further analysis is required. Additionally, SGDF increases the computational resource consumption compared to vanilla momentum or EMA. Future work should explore using low-cost statistical information (\eg the stability of stochastic gradients over sliding window) to design adaptive coefficients or hard-cut strategies, enabling seamless replacement of the most common momentum techniques in deep learning.

\section{Conclusion}
In this work, we introduced SGDF, an optimization approach inspired by statistical signal processing principles, to enhance gradient estimation in deep learning. Our approach provides a refined first-moment estimate, dynamically balancing trend and noise in gradient updates. This improvement addresses a core limitation in traditional momentum-based methods, which struggle to adaptively handle bias and variance within gradient distributions, often leading to biased or suboptimal updates. By minimizing mean-squared error in gradient estimation, SGDF estimates a more balanced gradient, promoting both convergence speed and generalization. Through extensive experiments employing various deep learning architectures on benchmark datasets, we showcase SGDF's superior performance compared to other state-of-the-art optimizers between convergence speed and generalization.
\section*{Acknowledgement}
Zhipeng Yao thanks to \href{https://araachie.github.io/}{Aram Davtyan} and \href{https://cvg.unibe.ch/people/favaro}{Prof. Dr. Paolo Favaro} in the Computer Vision Group at the University of Bern for discussing and improving the paper. Thanks to computational support from the Ascend AI Eco-Innovation Centre at the Shenyang AI Computing Hub.
{
    \small
    \bibliographystyle{ieeenat_fullname}
    \bibliography{main}
}
\newpage
\appendix
\onecolumn

\section{Bias-Variance Decomposition (Section 2 in main paper)}
\label{sec:appendixa}
\begin{definition}
	\label{def:unify_momentum}
	The unified momentum update rule is defined as:
	\begin{equation}
		m_t = \beta m_{t-1} + \mu g_t, \quad \theta_t = \theta_{t-1} - \alpha m_t,
	\end{equation}
	where \( \beta \in [0, 1) \) represents the decay or momentum factor, \( \mu \ge 1-\beta\) is a scaling parameter controlling the gradient contribution.  \(g_t = \nabla f_t + \epsilon_t, \epsilon_t \sim \mathcal{N}(0, \sigma^2)\).
	Specific cases include:
	\begin{itemize}
		\item \( \mu = 1 - \beta \): Exponential Moving Average (EMA),
		\item \( \mu = 1 \): Classical Momentum (CM).
	\end{itemize}
\end{definition}

\begin{assumption}
    \label{bias-variance}
    We assume that the gradient of the target function $f$ satisfies the following conditions:
    \begin{enumerate}
        \item \textbf{Lipschitz Continuity} There exists a constant $L > 0$ such that, for any $\theta$ and $\phi$, the inequality $||\nabla f(\theta) - \nabla f(\phi)|| \leq L ||\theta - \phi||$ holds.
        \item \textbf{Boundedness} There exists a constant $G > 0$ such that, for any $t$, the bound $\|\nabla f(\theta(t))\| \leq G$ is satisfied.
    \end{enumerate}
\end{assumption}

\begin{lemma}
\label{lem:bias_variance}
Let the true gradient $\nabla f_t$ be a deterministic quantity, and the stochastic gradient estimate $g_t$ follows:
\begin{equation}
    g_t = \nabla f_t + \epsilon_t,\quad \epsilon_t \sim \mathcal{N}(0,\sigma^2)
\end{equation}

For any gradient estimator $\hat{g}_t = \mathcal{A}(g_1,...,g_t)$, the mean squared error (MSE) decomposes as:
\begin{equation}
    \mathbb{E}[(\hat{g}_t - \nabla f_t)^2] = \underbrace{(\mathbb{E}[\hat{g}_t] - \nabla f_t)^2}_{\mathrm{Bias}^2} + \underbrace{\mathrm{Var}(\hat{g}_t)}_{\mathrm{Variance}}
\end{equation}

\end{lemma}

\begin{proof}
\begin{equation}
\begin{aligned}
    \mathbb{E}\left[(\hat{g}_t - \nabla f_t)^2\right] &= \mathbb{E}\left[(\hat{g}_t - \mathbb{E}[\hat{g}_t] + \mathbb{E}[\hat{g}_t] - \nabla f_t)^2\right] \\
    &= \mathbb{E}\left[(\hat{g}_t - \mathbb{E}[\hat{g}_t])^2\right] + (\mathbb{E}[\hat{g}_t] - \nabla f_t)^2 + 2\mathbb{E}\left[(\hat{g}_t - \mathbb{E}[\hat{g}_t])(\mathbb{E}[\hat{g}_t] - \nabla f_t)\right]  \\
    &= \mathrm{Var}(\hat{g}_t) + \mathrm{Bias}^2(\hat{g}_t) + 2(\mathbb{E}[\hat{g}_t] - \nabla f_t)\underbrace{\mathbb{E}[\hat{g}_t - \mathbb{E}[\hat{g}_t]]}_{=0} \\
    &= \mathrm{Var}(\hat{g}_t) + \mathrm{Bias}^2(\hat{g}_t)
\end{aligned}
\end{equation}

Thus, the fundamental decomposition holds:
\begin{equation}
    \mathbb{E}\left[(\hat{g}_t - \nabla f_t)^2\right] = \mathrm{Var}(\hat{g}_t) + \mathrm{Bias}^2(\hat{g}_t)
\end{equation}

\end{proof}

\begin{lemma}
\label{lem:unify_momentum_sde}
Refer to the SDE of vanilla SGD~\cite{stephan2017stochastic,li2019stochastic}, the Definition~\ref{def:unify_momentum} with learning rate \( \alpha_t = \gamma \alpha \) can be represented in continuous time as the stochastic differential equation (SDE):
\begin{equation}
    \begin{cases}
        d m(t) = [-(1 - \beta) m(t) + \mu \nabla f(\theta(t))] dt + \mu \sigma dW(t), \\
        d \theta(t) = -\gamma \alpha m(t) dt,
    \end{cases}
\end{equation}
where \( m(t) \) is the momentum, \( \theta(t) \) is the parameter, \( \beta \in [0, 1) \) is the momentum coefficient, \( \mu \in [0, 1) \) is the gradient scaling factor, \( \gamma \alpha \) is the effective learning rate with \( \gamma > 0 \) and \( \alpha > 0 \), \( \nabla f(\theta(t)) \) is the gradient of the objective function, \( \sigma \) is the noise standard deviation, and \( W(t) \) is a standard Wiener process. This approximation holds when the learning rate \( \gamma \alpha \) is sufficiently small.
\end{lemma}

\begin{proof}
Start with the discrete momentum update rule from Definition~\ref{def:unify_momentum}:
\begin{equation}
	m_{n+1} = \beta m_n + \mu g(t_n), \quad \theta_{n+1} = \theta_n - \alpha_t m_{n+1},
\end{equation}
where \( m_n = m(t_n) \) is the momentum, \( \theta_n = \theta(t_n) \) is the parameter, \( g(t_n) = \nabla f(t_n) + \epsilon_n \) with \( \epsilon_n \sim \mathcal{N}(0, \sigma^2) \), and \( \alpha_t = \gamma \alpha \) is the learning rate with \( \gamma > 0 \) and \( \alpha > 0 \) as constants.

Rewrite the momentum update in terms of the increment:
\begin{equation}
	\begin{aligned}
		m_{n+1} - m_n &= \beta m_n + \mu g(t_n) - m_n \\
		&= -(1 - \beta) m_n + \mu g(t_n)\\
		&=-(1 - \beta) m_n + \mu \nabla f(t_n) + \mu \epsilon_n. 
	\end{aligned}
\end{equation}

For the parameter:
\begin{equation}
	\theta_{n+1} - \theta_n = -\alpha_t m_{n+1} = -\gamma \alpha m_{n+1}.
\end{equation}

To model this as a continuous-time SDE, assume the learning rate \( \alpha_t = \gamma \alpha \) is sufficiently small, controlling the step size of the discrete updates. Define \( t_n = n \gamma \alpha \) as a time-like index scaled by the learning rate, and consider the increments \( m_{n+1} - m_n \) and \( \theta_{n+1} - \theta_n \) as approximations to differentials over a small time interval governed by \( \gamma \alpha \).

For the momentum, interpret the increment as a rate of change scaled by \( \gamma \alpha \):
\begin{equation}
	m_{n+1} - m_n \approx [-(1 - \beta) m_n + \mu \nabla f(\theta_n)] \gamma \alpha + \mu \epsilon_n.
\end{equation}

As \( \gamma \alpha \to 0 \), this approximates the deterministic drift:
\begin{equation}
	d m(t) = [-(1 - \beta) m(t) + \mu \nabla f(\theta(t))] dt.
\end{equation}

For the stochastic part, \( \epsilon_n \sim \mathcal{N}(0, \sigma^2) \) is discrete noise. In continuous time, it corresponds to white noise modeled by \( dW(t) \), with variance scaling as \( \sigma^2 \) per unit time. Adjust the noise amplitude with the learning rate scale:
\begin{equation}
	\mu \epsilon_n \approx \mu \sigma \sqrt{\gamma \alpha} Z_n, \quad Z_n \sim \mathcal{N}(0, 1),
\end{equation}

but in the SDE limit, the noise term becomes \( \mu \sigma dW(t) \), as the variance \( \mu^2 \sigma^2 dt \) matches the continuous process when \( \gamma \alpha \) defines the time step. Thus:
\begin{equation}
	d m(t) = [-(1 - \beta) m(t) + \mu \nabla f(\theta(t))] dt + \mu \sigma dW(t).
\end{equation}

For the parameter update:
\begin{equation}
	\theta_{n+1} - \theta_n = -\gamma \alpha m_{n+1} \approx -\gamma \alpha m(t) \cdot (\text{time step}),
\end{equation}

where the time step is implicitly \( dt \) in the continuous limit, yielding:
\begin{equation}
	d \theta(t) = -\gamma \alpha m(t) dt.
\end{equation}

Combining both, when \( \gamma \alpha \) is small, the discrete updates approximate:
\begin{equation}
	\begin{cases}
		d m(t) = [-(1 - \beta) m(t) + \mu \nabla f(\theta(t))] dt + \mu \sigma dW(t), \\
		d \theta(t) = -\gamma \alpha m(t) dt.
	\end{cases}
\end{equation}

This SDE captures the dynamics of the momentum and parameter updates, with \( \gamma \alpha \) as the effective learning rate driving the continuous approximation.
	
\end{proof}

\begin{lemma}
\label{lem:unify_momentum_solution}
Under Lemma~\ref{lem:unify_momentum_sde}, the solution to the stochastic differential equation (SDE), with initial conditions \( m(0) = 0 \) and \( \theta(0) = \theta_0 \), is given by:
\begin{equation}
	\begin{cases}
		m(t) = \mu \int_0^t e^{-(1 - \beta) (t - s)} \nabla f(\theta(s)) ds + \mu \sigma \int_0^t e^{-(1 - \beta) (t - s)} dW(s), \\
		\theta(t) = \theta_0 - \gamma \alpha \int_0^t m(s) ds,
	\end{cases}
\end{equation}

where \( W(t) \) is a standard Wiener process, and the integrals represent the stochastic evolution driven by the gradient \( \nabla f(\theta(t)) \) and noise.
\end{lemma}

\begin{proof}
We solve the coupled stochastic differential equation (SDE) system step-by-step:
\begin{equation}
	\begin{cases}
		d m(t) = [-(1 - \beta) m(t) + \mu \nabla f(\theta(t))] dt + \mu \sigma dW(t), \\
		d \theta(t) = -\gamma \alpha m(t) dt,
	\end{cases}
\end{equation}

with initial conditions \( m(0) = 0 \) and \( \theta(0) = \theta_0 \).

The equation for \( \theta(t) \) is deterministic and driven by \( m(t) \). Integrate:
\begin{equation}
	\begin{aligned}
		d \theta(t) &= -\gamma \alpha m(t) dt, \\
		\theta(t) - \theta(0) &= -\gamma \alpha \int_0^t m(s) ds.
	\end{aligned}
\end{equation}

Since \( \theta(0) = \theta_0 \), we obtain:
\begin{equation}
	\theta(t) = \theta_0 - \gamma \alpha \int_0^t m(s) ds.
\end{equation}

This expresses \( \theta(t) \) as a functional of \( m(t) \), which we now determine.

Consider the linear SDE for \( m(t) \) with a time-dependent forcing term:
\begin{equation}
	d m(t) = [-(1 - \beta) m(t) + \mu \nabla f(\theta(t))] dt + \mu \sigma dW(t).
\end{equation}

Rewrite it in standard form:
\begin{equation}
	d m(t) + (1 - \beta) m(t) dt = \mu \nabla f(\theta(t)) dt + \mu \sigma dW(t).
\end{equation}

To solve this, apply the integrating factor \( e^{\int_0^t (1 - \beta) ds} = e^{(1 - \beta) t} \). Multiply through by \( e^{(1 - \beta) t} \):
\begin{equation}
	\begin{aligned}
		e^{(1 - \beta) t} d m(t) + (1 - \beta) e^{(1 - \beta) t} m(t) dt &= \mu e^{(1 - \beta) t} \nabla f(\theta(t)) dt + \mu \sigma e^{(1 - \beta) t} dW(t).
	\end{aligned}
\end{equation}

Recognize the left-hand side as the differential of a product:
\begin{equation}
	\begin{aligned}
		d [e^{(1 - \beta) t} m(t)] &= e^{(1 - \beta) t} d m(t) + (1 - \beta) e^{(1 - \beta) t} m(t) dt, \\
		&= \mu e^{(1 - \beta) t} \nabla f(\theta(t)) dt + \mu \sigma e^{(1 - \beta) t} dW(t).
	\end{aligned}
\end{equation}

Integrate both sides from 0 to \( t \), with \( m(0) = 0 \), this simplifies to:
\begin{equation}
	\begin{aligned}
		e^{(1 - \beta) t} m(t) - e^{(1 - \beta) \cdot 0} m(0) &= \mu \int_0^t e^{(1 - \beta) s} \nabla f(\theta(s)) ds + \mu \sigma \int_0^t e^{(1 - \beta) s} dW(s)\\
		e^{(1 - \beta) t} m(t) &= \mu \int_0^t e^{(1 - \beta) s} \nabla f(\theta(s)) ds + \mu \sigma \int_0^t e^{(1 - \beta) s} dW(s), \\
		m(t) &= \mu \int_0^t e^{-(1 - \beta) (t - s)} \nabla f(\theta(s)) ds + \mu \sigma \int_0^t e^{-(1 - \beta) (t - s)} dW(s).
	\end{aligned}
\end{equation}

where the exponent is adjusted using \( e^{(1 - \beta) s} / e^{(1 - \beta) t} = e^{-(1 - \beta) (t - s)} \).

The expression for \( m(t) \) depends on \( \theta(s) \) via \( \nabla f(\theta(s)) \), where:
\begin{equation}
	\theta(s) = \theta_0 - \gamma \alpha \int_0^s m(u) du.
\end{equation}

Thus, the complete solution is:
\begin{equation}
	\begin{cases}
		m(t) = \mu \int_0^t e^{-(1 - \beta) (t - s)} \nabla f(\theta(s)) ds + \mu \sigma \int_0^t e^{-(1 - \beta) (t - s)} dW(s) \\
		\theta(t) = \theta_0 - \gamma \alpha \int_0^t m(s) ds.
	\end{cases}
\end{equation}

This integral form encapsulates the coupled dynamics, with \( \nabla f(\theta(t)) \) linking the equations and the stochastic term \( \int e^{-(1 - \beta) (t - s)} dW(s) \) as an Itô integral.

\end{proof}

\begin{theorem}
\label{thm:bias_variance_momentum}
Consider the unified momentum estimator \( m(t) \) defined by the stochastic differential equation (SDE) from Lemma~\ref{lem:unify_momentum_sde}, with solution given in Lemma~\ref{lem:unify_momentum_solution}. Assuming that the gradient \(\nabla f(\theta(t))\) is bounded and Lipschitz continuous, the bias and variance of \( m(t) \) as an estimator of \( \nabla f(\theta(t)) \) are approximately:
\begin{itemize}
\item \textbf{Bias}:
\begin{equation}
\begin{aligned}
     \left\| \mathrm{Bias}(m(t)) \right \|^2 &\leq \left( \frac{\mu \alpha L G }{(1 - \beta)^2} + \left( \frac{\mu}{1 - \beta} - 1 \right) \cdot G \right)^2,
\end{aligned}
\end{equation}

where \(L\) is the Lipschitz constant, \(G\) bounds \(\|\nabla f(\theta(t))\|\).
\item \textbf{Variance}:
\begin{equation}
	\mathrm{Var}(m(t)) \leq \frac{\mu^2 }{2 (1 - \beta)} \cdot \left( \sigma^2 +G^2\right). 
\end{equation}
where \(\sigma\) is the variance of random gradient sampling, \(G^2\) bounds \(\mathrm{Var}(\nabla f(\theta(t)))\).
\end{itemize}
\end{theorem}

\begin{proof}
We compute the bias and variance of \( m(t) \) relative to \( \nabla f(\theta(t)) \).

\textbf{1. Bias Calculation}  

Consider the unified momentum update rule:
\begin{equation}
    m_t = \beta m_{t-1} + \mu g_t, \quad \theta_t = \theta_{t-1} - \alpha m_t,
\end{equation}

where \(\beta \in [0, 1)\) represents the decay or momentum factor, \(\mu \in [1 - \beta, 1)\) is a scaling parameter controlling the gradient contribution, \(\alpha > 0\) is the learning rate, and \(g_t = \nabla f_t + \epsilon_t\) with \(\epsilon_t \sim \mathcal{N}(0, \sigma^2)\).

In continuous time, the expectation of \(m(t)\) is:
\begin{equation}
    \mathbb{E}[m(t)] = \mu \int_0^t e^{-(1 - \beta) (t - s)} \mathbb{E}[\nabla f(\theta(s))] \, ds,
\end{equation}

since the stochastic term has zero mean:
\begin{equation}
    \mathbb{E}\left[ \mu \sigma \int_0^t e^{-(1 - \beta) (t - s)} \, dW(s) \right] = 0.
\end{equation}

The squared bias is defined as:
\begin{equation}
\begin{aligned}
    \left( \mathrm{Bias}(m(t)) \right)^2 &= \left( \mathbb{E}[m(t)] - \nabla f(\theta(t)) \right)^2 \\
    &= \left( \mu \int_0^t e^{-(1 - \beta) (t - s)} \mathbb{E}[\nabla f(\theta(s))] \, ds - \nabla f(\theta(t)) \right)^2.
\end{aligned}
\end{equation}

We assume \(\nabla f\) is Lipschitz continuous with constant \(L > 0\):
\begin{equation}
    \|\nabla f(\theta) - \nabla f(\phi)\| \leq L \|\theta - \phi\|, \quad \forall \theta, \phi.
\end{equation}

Given \(\mu \geq 1 - \beta\), so \(\frac{\mu}{1 - \beta} \geq 1\).

From the continuous-time dynamics:
\begin{equation}
    \frac{d\theta}{dt} = -\alpha m(t),
\end{equation}

for \(s < t\), integrate from \(s\) to \(t\):
\begin{equation}
    \begin{aligned}
    \theta(t) &= \theta(s) - \alpha \int_s^t m(u) \, du, \\
    \theta(s) - \theta(t) &= \alpha \int_s^t m(u) \, du.
\end{aligned}
\end{equation}

Take the expected norm:
\begin{equation}
    \begin{aligned}
    \mathbb{E} \left[ \|\theta(s) - \theta(t)\| \right] &\leq \alpha \mathbb{E} \left[ \left\| \int_s^t m(u) \, du \right\| \right] \\
    &\leq \alpha \int_s^t \mathbb{E} \left[ \|m(u)\| \right] \, du\\
    &\leq \alpha G (t - s).
\end{aligned}
\end{equation}

According to Assumption~\ref{bias-variance}, \(\mathbb{E} \left[ \|m(u)\| \right] \leq \mathbb{E}[\nabla f(\theta(t))] \leq G\). \quad (Jensen's inequality)

Rewrite the bias by splitting the integral:
\begin{equation}
    \begin{aligned}
    \mathrm{Bias}(m(t)) &= \mu \int_0^t e^{-(1 - \beta) (t - s)} \mathbb{E}[\nabla f(\theta(s))] \, ds - \nabla f(\theta(t)) \\
    &= \mu \int_0^t e^{-(1 - \beta) (t - s)} \left( \mathbb{E}[\nabla f(\theta(s))] - \nabla f(\theta(t)) \right) \, ds \\
    & + \mu \int_0^t e^{-(1 - \beta) (t - s)} \nabla f(\theta(t)) \, ds - \nabla f(\theta(t)).
\end{aligned}
\end{equation}

Compute the second integral:
\begin{equation}
    \mu \int_0^t e^{-(1 - \beta) (t - s)} \, ds = \mu \cdot \frac{1 - e^{-(1 - \beta) t}}{1 - \beta},
\end{equation}

so:
\begin{equation}
\begin{aligned}
    \mathrm{Bias}(m(t)) &= \mu \int_0^t e^{-(1 - \beta) (t - s)} \left( \mathbb{E}[\nabla f(\theta(s))] - \nabla f(\theta(t)) \right) \, ds \\
    &+ \nabla f(\theta(t)) \left( \mu \cdot \frac{1 - e^{-(1 - \beta) t}}{1 - \beta} - 1 \right).
\end{aligned}
\end{equation}

Apply the triangle inequality:
\begin{equation}
    \begin{aligned}
    \|\mathrm{Bias}(m(t))\| &\leq \left\| \mu \int_0^t e^{-(1 - \beta) (t - s)} \left( \mathbb{E}[\nabla f(\theta(s))] - \nabla f(\theta(t)) \right) \, ds \right\| \\
    &\quad + \left| \mu \cdot \frac{1 - e^{-(1 - \beta) t}}{1 - \beta} - 1 \right| \cdot \|\nabla f(\theta(t))\|.
\end{aligned}
\end{equation}

Define:
\begin{equation}
    I_1 = \mu \int_0^t e^{-(1 - \beta) (t - s)} \left( \mathbb{E}[\nabla f(\theta(s))] - \nabla f(\theta(t)) \right) \, ds.
\end{equation}

Bound \(I_1\) using Lipschitz continuity:
\begin{equation}
    \begin{aligned}
    \|I_1\| &\leq \mu \int_0^t e^{-(1 - \beta) (t - s)} \left\| \mathbb{E}[\nabla f(\theta(s))] - \nabla f(\theta(t)) \right\| \, ds \\
    &\leq \mu L \int_0^t e^{-(1 - \beta) (t - s)} \mathbb{E} \left[ \|\theta(s) - \theta(t)\| \right] \, ds \\
    &\leq \mu L \alpha G \int_0^t e^{-(1 - \beta) (t - s)} (t - s) \, ds.
\end{aligned}
\end{equation}

Evaluate the integral:
\begin{equation}
\begin{aligned}
    \int_0^t e^{-(1 - \beta) (t - s)} (t - s) \, ds &= \int_0^t e^{-(1 - \beta) \tau} \tau \, d\tau \\
    &= \frac{1}{(1 - \beta)^2} - \left( \frac{t}{1 - \beta} + \frac{1}{(1 - \beta)^2} \right) e^{-(1 - \beta) t} \\
    &\leq \frac{1}{(1 - \beta)^2},
\end{aligned}
\end{equation}

thus:
\begin{equation}
    \|I_1\| \leq \frac{\mu \alpha L G }{(1 - \beta)^2}.
\end{equation}

According to Assumption~\ref{bias-variance}, \(\|\nabla f(\theta(t))\| \leq G\). Then:
\begin{equation}
    \|\mathrm{Bias}(m(t))\| \leq \frac{\mu \alpha L G }{(1 - \beta)^2} + \left| \mu \cdot \frac{1 - e^{-(1 - \beta) t}}{1 - \beta} - 1 \right| \cdot G.
\end{equation}

Since \(\mu \geq 1 - \beta\), we have \(\frac{\mu}{1 - \beta} \geq 1\). For the second term:
\begin{equation}
    \mu \cdot \frac{1 - e^{-(1 - \beta) t}}{1 - \beta} - 1 = \frac{\mu - (1 - \beta) + (1 - \beta) e^{-(1 - \beta) t}}{1 - \beta},
\end{equation}

where \(\mu - (1 - \beta) \geq 0\) and \((1 - \beta) e^{-(1 - \beta) t} \geq 0\), so:
\begin{equation}
    \left| \mu \cdot \frac{1 - e^{-(1 - \beta) t}}{1 - \beta} - 1 \right| = \mu \cdot \frac{1 - e^{-(1 - \beta) t}}{1 - \beta} - 1.
\end{equation}

As \(t \to \infty\), \(e^{-(1 - \beta) t} \to 0\):
\begin{equation}
    \left| \mu \cdot \frac{1 - e^{-(1 - \beta) t}}{1 - \beta} - 1 \right| \to \frac{\mu}{1 - \beta} - 1.
\end{equation}

Square the bound:
\begin{equation}
\begin{aligned}
    \left\| \mathrm{Bias}(m(t)) \right \|^2 &\leq \left( \frac{\mu \alpha L G }{(1 - \beta)^2} + \left( \mu \cdot \frac{1 - e^{-(1 - \beta) t}}{1 - \beta} - 1 \right) \cdot G \right)^2\\
     &\leq \left( \frac{\mu \alpha L G }{(1 - \beta)^2} + \left( \frac{\mu}{1 - \beta} - 1 \right) \cdot G \right)^2.
\end{aligned}
\end{equation}

where \(L\) is the Lipschitz constant, \(G\) bounds \(\|\nabla f(\theta(t))\|\).

\textbf{2. Variance Calculation}  

The fluctuation \( m(t) - \mathbb{E}[m(t)] \) is:
\begin{equation}
    \begin{aligned}
    m(t) - \mathbb{E}[m(t)] &= \mu \int_0^t e^{-(1 - \beta) (t - s)} \left[ \nabla f(\theta(s)) - \mathbb{E}[\nabla f(\theta(s))] \right] ds + \mu \sigma \int_0^t e^{-(1 - \beta) (t - s)} \, dW(s).
    \end{aligned}
\end{equation}

Thus, the variance becomes:
Thus, the variance becomes:
\begin{equation}
\begin{aligned}
    \mathrm{Var}(m(t)) &= \mathbb{E}\left[ \left\| m(t) - \mathbb{E}[m(t)] \right\|^2 \right] \\
    &= \mathbb{E}\left[ \left\| \mu \int_0^t e^{-(1 - \beta) (t - s)} \left[ \nabla f(\theta(s)) - \mathbb{E}[\nabla f(\theta(s))] \right] ds + \mu \sigma \int_0^t e^{-(1 - \beta) (t - s)} \, dW(s) \right\|^2 \right] \\
    &= \mathbb{E}\left[ \left\| \mu \int_0^t e^{-(1 - \beta) (t - s)} \left[ \nabla f(\theta(s)) - \mathbb{E}[\nabla f(\theta(s))] \right] ds \right\|^2 \right] + \mathbb{E}\left[ \left\| \mu \sigma \int_0^t e^{-(1 - \beta) (t - s)} \, dW(s) \right\|^2 \right] \\
    &= \underbrace{\mu^2 \int_0^t e^{-2(1 - \beta) (t - s)} \mathrm{Var}(\nabla f(\theta(s))) \, ds}_{\text{Gradient Variance}} + \underbrace{\mathbb{E}\left[ \left\| \mu \sigma \int_0^t e^{-(1 - \beta) (t - s)} \, dW(s) \right\|^2 \right]}_{\text{Noise Variance}}
\end{aligned}
\end{equation}

The noise variance term is derived as:
\begin{equation}
	\begin{aligned}
		\text{Noise Variance} &= \mathbb{E}\left[ \left( \mu \sigma \int_0^t e^{-(1 - \beta) (t - s)} \, dW(s) \right)^2 \right] \\
		&= \mu^2 \sigma^2 \mathbb{E}\left[ \left( \int_0^t e^{-(1 - \beta) (t - s)} \, dW(s) \right)^2 \right] \\
		&= \mu^2 \sigma^2 \int_0^t e^{-2 (1 - \beta) (t - s)} \, ds \quad (\text{Itô isometry}) \\
		&= \mu^2 \sigma^2 \int_0^t e^{-2 (1 - \beta) u} \, du \quad (\text{let } u = t - s) \\
		&= \mu^2 \sigma^2 \cdot \frac{1 - e^{-2 (1 - \beta) t}}{2 (1 - \beta)}.
	\end{aligned}
\end{equation}

The gradient variance term is derived as:
\begin{equation}
	\begin{aligned}
		\text{Gradient Variance} &= \mu^2 \int_0^t e^{-2(1 - \beta) (t - s)} \mathrm{Var}(\nabla f(\theta(s))) \, ds \\
		&\leq \mu^2 \int_0^t e^{-2(1 - \beta) (t - s)} G^2 \, ds \quad (\text{since } \mathrm{Var}(\nabla f(\theta(s))) \leq G^2 \text{ by Assumption~\ref{bias-variance}}) \\
		&= \mu^2 G^2 \int_0^t e^{-2(1 - \beta) (t - s)} \, ds \\
		&= \mu^2 G^2 \cdot \frac{1 - e^{-2(1 - \beta) t}}{2(1 - \beta)} \quad (\text{let } u = t - s).
	\end{aligned}
\end{equation}

The variance is:
\begin{equation}
\begin{aligned}
    \mathrm{Var}(m(t)&= \mu^2 \sigma^2 \cdot \frac{1 - e^{-2 (1 - \beta) t}}{2 (1 - \beta)} + \mu^2 \left( \frac{1 - e^{-2(1 - \beta) t}}{2(1 - \beta)} \right) G^2,\\
    &\leq \frac{\mu^2 }{2 (1 - \beta)} \cdot \left( \sigma^2 +G^2\right). 
\end{aligned}
\end{equation}

\end{proof}

\newpage

\section{Method Derivation (Section 3 in main paper)}
\subsection{Wiener Filter Derivation for Gradient Estimation (Main paper Section 3.1)}
\label{proof_method}

In the stochastic gradient descent (SGD) process, given the sequence of gradients $\{g_t\}$, our objective is to estimate $\widehat{g}_t$, which incorporates information from both historical gradients and the current gradient. The Wiener filter provides a mechanism to minimize the mean squared error in this estimation. We start by constructing $\widehat{g}_t$ as a simple average and then refine it using the properties of the Wiener filter.

\begin{equation}
\begin{aligned}
	\widehat{g}_t &= \frac{1}{T+1} \sum_{i=1}^{T} g_i + \frac{1}{T+1} g_t \\
	&= \frac{1}{T+1} \cdot \frac{T}{T} \sum_{i=1}^{T} g_i + \frac{1}{T+1} g_t \\
	&= \frac{T}{T+1} \bar{g}_t + \frac{1}{T+1} g_t,
\end{aligned}
\end{equation}

where $\bar{g}_t = \frac{1}{T} \sum_{i=1}^{T} g_i$ represents the arithmetic mean of the previous gradients $\{g_i\}$.

We replace the arithmetic mean of gradients $\bar{g}_t$ with the momentum term $\widehat{m}_t$ to capture historical gradient information more effectively. Thus, we rewrite $\widehat{g}_t$ as follows:

\begin{equation}
\begin{aligned}
	\widehat{g}_t &\approx \frac{T}{T+1} \widehat{m}_t + \frac{1}{T+1} g_t \\
	&= \left(1 - \frac{1}{T+1}\right) \widehat{m}_t + \frac{1}{T+1} g_t \\
	&= \widehat{m}_t - K_t \widehat{m}_t + K_t g_t \\
	&= \widehat{m}_t + K_t (g_t - \widehat{m}_t),
\end{aligned}
\end{equation}

where $K_t = \frac{1}{T+1}$ serves as the initial Wiener gain, controlling the balance between historical information (via $\widehat{m}_t$) and new information (via $g_t$).

To achieve an optimal balance, we define $\widehat{g}_t$ as a weighted combination of the momentum term $\widehat{m}_t$ and the current gradient $g_t$, aiming to minimize the variance of $\widehat{g}_t$. The variance of $\widehat{g}_t$ can be expressed as:

\begin{equation}
\begin{aligned}
	\mathrm{Var}(\widehat{g}_t) &= \mathrm{Var}((1 - K_t)\widehat{m}_t + K_t g_t) \\
	&= (1 - K_t)^2 \mathrm{Var}(\widehat{m}_t) + K_t^2 \mathrm{Var}(g_t).
\end{aligned}
\end{equation}

To find the optimal value of $K_t$, we take the derivative of $\mathrm{Var}(\widehat{g}_t)$ with respect to $K_t$ and set it to zero:

\begin{equation}
\begin{aligned}
	\frac{\mathrm{d} \mathrm{Var}(\widehat{g}_t)}{\mathrm{d} K_t} &= 2(1 - K_t) \mathrm{Var}(\widehat{m}_{t}) - 2 K_t \mathrm{Var}(g_{t}) = 0, \\
	(1 - K_t) \mathrm{Var}(\widehat{m}_{t}) &= K_t \mathrm{Var}(g_{t}),
\end{aligned}
\end{equation}

solving for $K_t$ gives:

\begin{equation}
	K_t = \frac{\mathrm{Var}(\widehat{m}_{t})}{\mathrm{Var}(\widehat{m}_{t}) + \mathrm{Var}(g_{t})}.
\end{equation}

The final expression for $K_t$ indicates that the optimal interpolation coefficient is the ratio of the variance of the momentum term to the sum of the variances of the momentum term and the current gradient. This balance embodies the Wiener filter's purpose: to optimally combine past information with new observations, thus minimizing estimation error caused by noise in the gradient data.

\subsection{Variance Correction (Correction factor in main paper Section 3.1)}
\label{proof_correction}

The momentum term \( m_{t} \) in stochastic gradient descent is defined as:
\begin{equation}
	m_{t} = (1 - \beta_{1}) \sum_{i=1}^{t} \beta_{1}^{t-i} g_{i},
\end{equation}

which means that \( m_t \) is a weighted sum of past gradients, where the weights decrease exponentially over time according to the factor \( \beta_1 \).

To accurately estimate the variance of \( m_t \) using the variance of \( g_t \), we derive a correction factor under the assumption that \( g_t \) terms are independent and identically distributed (i.i.d.) with constant variance \( \sigma_{g}^{2} \).

Each weighted gradient term \( \beta_{1}^{t-i} g_i \) has a variance of \( \beta_{1}^{2(t-i)} \sigma_{g}^{2} \), because the variance scaling factor becomes \( \beta_{1}^{2(t-i)} \) in the variance computation due to the quadratic nature of the variance operator.

Given that \( m_t \) is a sum of these weighted terms and assuming independence among \( g_i \), the variance of \( m_t \) is the sum of the variances of all weighted gradients:
\begin{equation}
	\sigma_{m_{t}}^{2} = (1 - \beta_{1})^{2} \sigma_{g}^{2} \sum_{i=1}^{t} \beta_{1}^{2(t-i)}.
\end{equation}

The factor \( (1 - \beta_{1})^2 \) appears from the multiplication factor \( (1 - \beta_{1}) \) in the definition of \( m_t \), which also applies to the variance calculation.

The summation \( \sum_{i=1}^{t} \beta_{1}^{2(t-i)} \) forms a geometric series:
\begin{equation}
	\sum_{i=1}^{t} \beta_{1}^{2(t-i)} = \frac{1 - \beta_{1}^{2t}}{1 - \beta_{1}^{2}}.
\end{equation}

As \( t \rightarrow \infty \) and given that \( \beta_1 < 1 \), we find that \( \beta_1^{2t} \rightarrow 0 \), so the series converges to:
\begin{equation}
	\sum_{i=1}^{t} \beta_1^{2(t-i)} \approx \frac{1}{1 - \beta_1^2}.
\end{equation}

Substituting back, we obtain the long-term variance of \( m_t \) as:
\begin{equation}
	\sigma^2_{m_t} = \frac{\left(1 - \beta_1\right)^2}{1 - \beta_1^2} \qquad \sigma^2_g = \frac{1 - \beta_1}{1 + \beta_1} \sigma^2_g.
\end{equation}

Thus, the correction factor we derived is:
\begin{equation}
    \left(\frac{1 - \beta_1}{1 + \beta_1}\right) \cdot (1 - \beta_1^{2t}).
\end{equation}

This correction factor \(\left(\frac{1 - \beta_1}{1 + \beta_1}\right) \cdot (1 - \beta_1^{2t})\) allows us to adjust the variance of the EMA gradient to accurately estimate the variance of the momentum gradient \( m_t \) using the original variance \( \sigma^2_g \). This adjustment reflects the effect of exponentially decaying weights in \( m_t \), yielding a more stable gradient estimate with reduced noise over time.

\subsection{Fusion of Gaussian Distributions (Main paper Section 3.2)}
\label{proof_section3.2}

In this section, we address the fusion of two Gaussian distributions to produce a more reliable gradient estimate in the stochastic gradient descent (SGD) process. This fusion combines information from both the historical momentum term $\widehat{m}_t$ and the current gradient $g_t$, resulting in an estimate with reduced uncertainty. Here, "fusion" refers to finding an optimal combined distribution that minimizes mean-square error by utilizing both sources of information.

Consider the following two Gaussian distributions:
\begin{itemize}
	\item The momentum term $\widehat{m}_t$ follows a normal distribution with mean $\mu_{m}$ and variance $\sigma_{m}^2$, denoted as $\widehat{m}_t \sim \mathcal{N}(\mu_{m}, \sigma_{m}^2)$.
	\item The current gradient $g_t$ follows a normal distribution with mean $\mu_{g}$ and variance $\sigma_{g}^2$, denoted as $g_t \sim \mathcal{N}(\mu_{g}, \sigma_{g}^2)$.
\end{itemize}

Our objective is to derive a new Gaussian distribution $\mathcal{N}(\mu_{\widehat{g}_t}, \sigma_{\widehat{g}_t}^2)$ that combines these two distributions, yielding a more accurate estimate for $\widehat{g}_t$.

The probability density function (PDF) of the product of these two Gaussian distributions is given by:

\begin{equation}
	N(\widehat{m}_t; \mu_{m}, \sigma_{m}) \cdot N(g_t; \mu_{g}, \sigma_{g}) = \frac{1}{2\pi\sigma_{m}\sigma_{g}} \exp\left(-\frac{(\widehat{m}_t-\mu_{m})^2}{2\sigma_{m}^2} -\frac{(g_t-\mu_{g})^2}{2\sigma_{g}^2}\right).
\end{equation}

We seek a new Gaussian distribution with mean $\mu_{\widehat{g}_t}$ and variance $\sigma_{\widehat{g}_t}^2$ that best approximates this product:

\begin{equation}
	N(\widehat{g}_t; \mu_{\widehat{g}_t}, \sigma_{\widehat{g}_t}^2) = \frac{1}{\sqrt{2\pi}\sigma_{\widehat{g}_t}} \exp\left(-\frac{(\widehat{g}_t - \mu_{\widehat{g}_t})^2}{2\sigma_{\widehat{g}_t}^2}\right).
\end{equation}

We start by simplifying the exponent terms. Defining the combined expression as $t$ for clarity, we have:

\begin{equation}
	\begin{aligned}
		t &= -\frac{\left(\widehat{g}_t - \mu_{m}\right)^2}{2 \sigma_{m}^2} - \frac{\left(\widehat{g}_t - \mu_{g}\right)^2}{2 \sigma_{g}^2} \\
		&= -\frac{\sigma_{g}^2\left(\widehat{g}_t - \mu_{m}\right)^2 + \sigma_{m}^2\left(\widehat{g}_t - \mu_{g}\right)^2}{2 \sigma_{m}^2 \sigma_{g}^2} \\
		&= -\frac{\left(\widehat{g}_t - \frac{\sigma_{g}^2 \mu_{m} + \sigma_{m}^2 \mu_{g}}{\sigma_{m}^2 + \sigma_{g}^2}\right)^2}{\frac{2 \sigma_{m}^2 \sigma_{g}^2}{\sigma_{m}^2 + \sigma_{g}^2}} + \frac{\left(\mu_{m} - \mu_{g}\right)^2}{2\left(\sigma_{m}^2 + \sigma_{g}^2\right)}.
	\end{aligned}
\end{equation}

Matching coefficients in the exponent terms, we obtain the fused mean $\mu_{\widehat{g}_t}$ and variance $\sigma_{\widehat{g}_t}^2$ as follows:

\begin{equation}
	\mu_{\widehat{g}_t} = \frac{\sigma_{g}^2 \mu_{m} + \sigma_{m}^2 \mu_{g}}{\sigma_{m}^2 + \sigma_{g}^2}, \quad \sigma_{\widehat{g}_t}^2 = \frac{\sigma_{m}^2 \sigma_{g}^2}{\sigma_{m}^2 + \sigma_{g}^2}.
\end{equation}

The fused mean $\mu_{\widehat{g}_t}$ represents a weighted average of the two means, $\mu_{m}$ and $\mu_{g}$, with weights inversely proportional to their variances. This places $\mu_{\widehat{g}_t}$ between $\mu_{m}$ and $\mu_{g}$, closer to the mean with the smaller variance, reflecting greater confidence in estimates with less uncertainty.

The fused variance $\sigma_{\widehat{g}_t}^2$ is smaller than either of the original variances $\sigma_{m}^2$ and $\sigma_{g}^2$, indicating reduced uncertainty due to the combined information. This reduction highlights the benefit of fusing historical momentum and current gradient estimates: by incorporating information from both sources, the resulting gradient estimate $\widehat{g}_t$ is more stable and less affected by noise, enhancing the overall reliability of the gradient descent process.

\newpage

\section{Convergence analysis in convex online learning case (Theorem 3.2 in main paper).}
\label{convex}
\begin{assumption}
	\label{conas}	
	Variables are bounded: \(\exists D \text{ such that } \forall t, \Vert\theta_t\Vert_2 \leq D\). Gradients are bounded: \(\exists G \text{ such that } \forall t, \Vert g_t\Vert_2 \leq G\).
\end{assumption}

\begin{definition}
	Let \(f_t(\theta_t)\) be the loss at time \(t\) and \(f_t(\theta^*)\) be the loss of the best possible strategy at the same time. The cumulative regret \(R(T)\) at time \(T\) is defined as:
	\begin{align}
		R(T) = \sum_{t=1}^{T} f_t(\theta_t) - f_t(\theta^* )
	\end{align}
\end{definition}

\begin{definition}
	\label{defA.9}
	If a function \(f\): \(R^d \rightarrow R\) is convex if for all \(x, y \in R^d\)  for all \(\lambda \in [0, 1]\),
	\begin{align}
		\lambda f(x) + (1-\lambda) f(y) \geq f(\lambda x + (1-\lambda) y)
	\end{align}
\end{definition}

Also, notice that a convex function can be lower bounded by a hyperplane at its tangent.

\begin{lemma}
\label{lem.B.4}
If a function  \(f: R^{d} \rightarrow R\)  is convex, then for all  \(x, y \in R^{d}\) ,
\begin{align}
    f(y) \geq f(x)+\nabla f(x)^{T}(y-x)
\end{align}
\end{lemma}

The above lemma can be used to upper bound the regret, and our proof for the main theorem is constructed by substituting the hyperplane with SGDF update rules.

The following two lemmas are used to support our main theorem. We also use some definitions to simplify our notation, where  \(g_{t} \triangleq \nabla f_{t}\left(\theta_{t}\right)\)  and  \(g_{t, i}\)  as the  \(i^{\text {th }}\)  element. We denote  \(g_{1: t, i} \in \mathbb{R}^{t}\)  as a vector that contains the  \(i^{\text {th }}\)  dimension of the gradients over all iterations till  \(t, g_{1: t, i}=\left[g_{1, i}, g_{2, i}, \cdots, g_{t, i}\right]\) 

\begin{lemma}
\label{lem.B.5}
Let \( g_{t} = \nabla f_{t}(\theta_{t}) \) and \( g_{1: t} \) be defined as above and bounded, 
\begin{align} 
    \left\Vert g_{t} \right\Vert_{2} \leq G, \left\Vert g_{t} \right\Vert_{\infty} \leq G_{\infty}. 
\end{align}

Then,
\begin{align} 
    \sum_{t=1}^{T} g_{t, i} \leq 2 G_{\infty} \left\Vert g_{1: T, i} \right\Vert_{2}.
\end{align}
\end{lemma}

\begin{proof}
We will prove the inequality using induction over \( T \).
For the base case \( T = 1 \):
\begin{align} 
	g_{1, i} \leq 2 G_{\infty} \left\Vert g_{1, i} \right\Vert_{2}.
\end{align}	 

Assuming the inductive hypothesis holds for \( T-1 \), for the inductive step:
\begin{equation}
	\begin{aligned}
		\sum_{t=1}^{T} g_{t, i} 
		&= \sum_{t=1}^{T-1} g_{t, i} + g_{T, i} \\
		&\leq 2 G_{\infty} \left\Vert g_{1: T-1, i} \right\Vert_{2} + g_{T, i} \\
		&= 2 G_{\infty} \sqrt{\left\Vert g_{1: T, i} \right\Vert_{2}^{2} - g_{T}^{2}} + g_{T, i}^{2}.
	\end{aligned}
\end{equation}

Given,
\begin{align}
	\left\Vert g_{1: T, i} \right\Vert_{2}^{2} - g_{T, i}^{2} + \frac{g_{T, i}^{4}}{4\left\Vert g_{1: T, i} \right\Vert_{2}^{2}} \geq \left\Vert g_{1: T, i} \right\Vert_{2}^{2} - g_{T, i}^{2}, 
\end{align}	

taking the square root of both sides, we get:
\begin{equation}
	\begin{aligned}
		\sqrt{\left\Vert g_{1: T, i} \right\Vert_{2}^{2} - g_{T, i}^{2}} 
		&\leq \left\Vert g_{1: T, i} \right\Vert_{2} - \frac{g_{T, i}^{2}}{2\left\Vert g_{1: T, i} \right\Vert_{2}} \\
		&\leq \left\Vert g_{1: T, i} \right\Vert_{2} - \frac{g_{T, i}^{2}}{2 \sqrt{G_{\infty}^{2}}}.
	\end{aligned}
\end{equation}

Substituting into the previous inequality:
\begin{align}
	G_{\infty} \sqrt{\left\Vert g_{1: T, i} \right\Vert_{2}^{2} - g_{T, i}^{2}} + \sqrt{g_{T, i}^{2}} \leq 2 G_{\infty} \left\Vert g_{1: T, i} \right\Vert_{2}
\end{align}	
    
\end{proof}

\begin{lemma}
Let bounded  $g_{t},\left\Vert g_{t}\right\Vert_{2} \leq G$ ,  $\left\Vert g_{t}\right\Vert_{\infty} \leq G_{\infty}$, the following inequality holds
\begin{align}
    \sum_{t=1}^{T} \widehat{m}_{t, i}^{2} \leq \frac{4 G_{\infty}^{2}}{(1-\beta_{1})^{2}} \left\Vert g_{1: T, i}\right\Vert_{2}^{2}
\end{align}
\label{lem.B.6}
\end{lemma}

\begin{proof}

Under the inequality: $\frac{1}{\left(1-\beta_{1}^{t}\right)^{2}} \leq \frac{1}{\left(1-\beta_{1}\right)^{2}}$ . We can expand the last term in the summation using the updated rules in Algorithm 1,

\begin{equation}
	\begin{aligned}
		\sum_{t=1}^{T} \widehat{m}_{t, i}^{2} & =\sum_{t=1}^{T-1} \widehat{m}_{t, i}^{2} + \frac{\left(\sum_{k=1}^{T}\left(1-\beta_{1}\right) \beta_{1}^{T-k} g_{k, i}\right)^{2}}{\left(1-\beta_{1}^{T}\right)^{2}}  \\
		& \leq \sum_{t=1}^{T-1} \widehat{m}_{t, i}^{2}+\frac{\sum_{k=1}^{T} T\left(\left(1-\beta_{1}\right) \beta_{1}^{T-k} g_{k, i}\right)^{2}}{\left(1-\beta_{1}^{T}\right)^{2}}  \\
		& \leq \sum_{t=1}^{T-1}\widehat{m}_{t, i}^{2}+\frac{\left(1-\beta_{1}\right)^{2}}{\left(1-\beta_{1}^{T}\right)^{2}} \sum_{k=1}^{T} T\left(\beta_{1}^{2}\right)^{T-k}\left\Vert g_{k, i}\right\Vert_{2}^{2} \\
		& \leq \sum_{t=1}^{T-1} \widehat{m}_{t, i}^{2}+ T \sum_{k=1}^{T} \left(\beta_{1}^{2}\right)^{T-k} \left\Vert g_{k, i}\right\Vert_{2}^{2}
	\end{aligned}
\end{equation}

Similarly, we can upper-bound the rest of the terms in the summation.
\begin{equation}
	\begin{aligned}
		\sum_{t=1}^{T} \widehat{m}_{t, i}^{2} & \leq \sum_{t=1}^{T} \left\Vert g_{t, i}\right\Vert_{2}^{2} \sum_{j=0}^{T-t} t \beta_{1}^{j} \\
		& \leq \sum_{t=1}^{T} \left\Vert g_{t, i}\right\Vert_{2}^{2} \sum_{j=0}^{T} t \beta_{1}^{j}
	\end{aligned}
\end{equation}

For  $\beta_{1}<1$ , using the upper bound on the arithmetic-geometric series,  $\sum_{t} t \beta_{1}^{t}<\frac{1}{(1-\beta_{1})^{2}}$  :
\begin{align}
	\sum_{t=1}^{T} \left\|g_{t, i}\right\|_{2}^{2} \sum_{j=0}^{T} t \beta_{1}^{j} \leq \frac{1}{(1-\beta_{1})^{2}} \sum_{t=1}^{T} \left\Vert g_{t, i}\right\Vert_{2}^{2}
\end{align}

Apply Lemma~\ref{lem.B.5},
\begin{align}
	\sum_{t=1}^{T} \widehat{m}_{t, i}^{2} \leq \frac{4 G_{\infty}^{2}}{(1-\beta_{1})^{2}}\left\Vert g_{1: T, i}\right\Vert_{2}^{2}
\end{align}

\end{proof}

\begin{theorem}
	\label{the.A.13}
	Assume that the function \( f_t \) has bounded gradients, \( \Vert\nabla f_t(\theta)\Vert_2 \leq G \), \( \Vert\nabla f_t(\theta)\Vert_\infty \leq G_\infty \) for all \( \theta \in \mathbb{R}^d \) and the distance between any \( \theta_t \) generated by SGDF is bounded, \( \Vert\theta_n - \theta_m\Vert_2 \leq D \), \( \Vert\theta_m - \theta_n\Vert_\infty \leq D_\infty \) for any \( m, n \in \{1, ..., T\} \), and \( \beta_1, \beta_2 \in [0, 1) \). Let \( \alpha_t = \alpha/\sqrt{t} \). For all \( T \geq 1 \), SGDF achieves the following guarantee:
	\begin{equation}
		\begin{aligned}
			R(T) \leq &\frac{D^2}{\alpha} \sum_{i=1}^{d}  \sqrt{T} + 
			\frac{2 D_\infty G_{\infty}}{1-\beta_{1}} \sum_{i=1}^{d}\left\Vert g_{1: T, i} \right\Vert_{2} +\frac{2\alpha G_{\infty}^{2}(1 + (1 - \beta_{1})^2)}{\sqrt{T}(1-\beta_{1})^{2}} \sum_{i=1}^{d} \left\Vert g_{1: T, i} \right\Vert_{2}^{2}
		\end{aligned}
	\end{equation}
\end{theorem}

\begin{proof}

We aim to prove the convergence of the algorithm by showing that \(R(T)\) is bounded, or equivalently, that \(\dfrac{R(T)}{T}\) converges to zero as \(T\) goes to infinity.

To express the cumulative regret in terms of each dimension, let \( f_t(\theta_t) \) and \( f_t(\theta^*) \) represent the loss and the best strategy's loss for the \(d\)th dimension, respectively. Define \( R_{T,d} \) as:
\begin{align}
	R_{T,i} = \sum_{t=1}^{T} f_t(\theta_t) - f_t(\theta^*) 
\end{align}

Then, the overall regret \( R(T) \) can be expressed in terms of all dimensions \( D \) as:
\begin{align}
	R(T) = \sum_{d=1}^{D} R_{T,i}
\end{align}

Establishing the Connection: From the Iteration of \(\theta_t\) to \(\left< g_t,\theta_t - \theta^* \right>\)

Using Lemma~\ref{lem.B.4}, we have,
\begin{align}
	f_t(\theta_{t})-f_t(\theta^*)\leq g^T_t(\theta_{t}-\theta^*)=\sum_{i=1}^{d}g_{t,i}(\theta_{t,i}-\theta_{,i}^*)
\end{align}

From the update rules presented in algorithm 1,
\begin{equation}
	\begin{aligned}
		\theta_{t+1} &=\theta_{t} - \alpha_t \widehat{g}_t\\
		&= \theta_{t} - \alpha_t \big(\widehat{m}_{t}+K_{t,d}(g_{t}-\widehat{m}_{t})\big)
	\end{aligned}
\end{equation}

We focus on the \(i^{\text{th}}\) dimension of the parameter vector \(\theta_{t} \in R^d\). Subtract the scalar \(\theta^{*}_{,i}\) and square both sides of the above update rule, we have,
\begin{equation}
	\begin{aligned}
		(\theta_{t+1,d} - \theta_{,i}^*)^2 &= (\theta_{t,i} -  \theta_{,i}^*)^2 -2\alpha_t (\widehat{m}_{t,i}+K_{t,d}(g_{t,i}-\widehat{m}_{t,i}))(\theta_{t,i} - \theta_{,i}^*) + \alpha_t^2 \widehat{g}^2_t
	\end{aligned}
\end{equation}

Separating items\(g_{t,i}(\theta_{t,i} - \theta_{,i}^*)\):
\begin{equation}
	\begin{aligned}
		g_{t,d}(\theta_{t,i} - \theta_{,i}^*) &= \underset{(1)}{\underbrace{\frac{\left( \theta_{t,i} - \theta_{,i}^* \right)^2 - \left( \theta_{t+1,i} - \theta_{,i}^* \right)^2}{2\alpha_t K_{t,i}} }} \underset{(2)}{\underbrace{-\frac{1 - K_{t,i}}{K_{t,i}}\widehat{m}_{t,i}\left( \theta_{t,i} - \theta_{,i}^* \right) }} +\underset{(3)}{\underbrace{\frac{\alpha_t}{2K_{t,i}}(\widehat{g}_{t,i})^2 }}
	\end{aligned}
\end{equation}

We then deal with (1), (2) and (3) separately.

For the first term (1), we have:
\begin{equation}
	\begin{aligned}
		&\sum_{t=1}^{T} \frac{\left( \theta_{t,i} - \theta_{,i}^* \right)^2 - \left( \theta_{t+1,i} - \theta_{,i}^* \right)^2}{2\alpha_t K_{t,i}}\\
		&\leq \sum_{t=1}^{T} \frac{\left( \theta_{t,i} - \theta_{,i}^* \right)^2 - \left( \theta_{t+1,i} - \theta_{,i}^* \right)^2}{2\alpha_t K_{t,i}}\\
		&= \frac{\left( \theta_{1,i} - \theta_{,i}^* \right)^2}{2\alpha_1 K_{1,i}} - \frac{\left( \theta_{T+1,i} - \theta_{,i}^* \right)^2}{2\alpha_T K_{T,i}} + \sum_{t=2}^{T} ( \theta_{t,i} - \theta_{,i}^*)^2 \left[ \frac{1}{2\alpha_t K_{t,i}} - \frac{1}{2\alpha_{t-1} K_{t-1,i}} \right]
	\end{aligned}
\end{equation}

Given that \(-\dfrac{\left(\theta_{T+1,i}-\theta_{,i}^{*}\right)^{2}}{2 \alpha_{T}\left(K_{1}\right)} \leq 0\) and \(\dfrac{\left(\theta_{1,i}-\theta_{,i}^{*}\right)^{2}}{2 \alpha_{1}\left(K_{T}\right)} \leq \dfrac{D_{i}^{2}}{2 \alpha_{1}\left(K_{T}\right)}\), we can bound it as:
\begin{equation}
	\begin{aligned}
		&\sum_{t=1}^{T} \frac{\left( \theta_{t,i} - \theta_{,i}^* \right)^2 - \left( \theta_{t+1,i} - \theta_{,i}^* \right)^2}{2\alpha_t K_{t,i}}\\
		&\leq \sum_{i=1}^{d} \frac{( \theta_{t,i} - \theta_{,i}^*)^2}{2 \alpha_{t}K_{t,i}} \\
	\end{aligned}
\end{equation}

For the second term (2), we have:
\begin{equation}
	\begin{aligned}
		&\sum_{t=1}^{T} -\frac{1 - K_{t,i}}{K_{t,i}}\widehat{m}_{t,i}\left( \theta_{t,i} - \theta_{,i}^* \right) \\
		&= \sum_{t=1}^{T} -\frac{1 - K_{t,i}}{K_{t,i}(1-\beta_1^t)} \left( \sum_{i=1}^{T} (1-\beta_{1,i}) \prod_{j=i+1}^{T} \beta_{1,j} \right)g_{t,i} \left( \theta_{t,i} - \theta_{,i}^* \right) \\
		&\leq \sum_{t=1}^{T} -\frac{1 - K_{t,i}}{K_{t,d}(1-\beta_1^t)} \left( 1- \prod_{i=1}^{T} \beta_{1,i} \right)g_{t,i} ( \theta_{t,i} - \theta_{,i}^* ) \\
		&\leq \sum_{t=1}^{T} \frac{1 - K_{t,i}}{K_{t,d}(1-\beta_1^t)} g_{t,i} ( \theta_{t,i} - \theta_{,i}^* )
	\end{aligned}
\end{equation}

For the third term (3), we have:
\begin{equation}
	\begin{aligned}
		\sum_{t=1}^{T} \frac{\alpha_t}{2K_{t,i}}(\widehat{g}_{t,i})^2 
		&\leq \sum_{t=1}^{T} \frac{\alpha_t}{2K_{t,i}} \left( \widehat{m}_{t,i} + K_t  (g_{t,i} - \widehat{m}_{t,i}) \right)^2 \\
		&\leq \sum_{t=1}^{T} \frac{\alpha_t}{2K_{t,i}} \left( (1-K_{t,i}) \widehat{m}_{t,i} + K_{t,d} g_{t,i} \right)^2 \\
		&\leq \sum_{t=1}^{T} \frac{\alpha_t}{2K_{t,i}} \left( 2(1-K_{t,i})^{2} \widehat{m}_{t,i}^{2} + 2K_{t,i}^{2} g_{t,i}^{2} \right)\\
		&\leq \sum_{t=1}^{T} \frac{\alpha_t}{K_{t,i}} \left( (1-K_{t,i})^{2} \widehat{m}_{t,i}^{2} + K_{t,i}^{2} g_{t,i}^{2} \right)
	\end{aligned}
\end{equation}

Collate all the items that we have:
\begin{equation}
	\begin{aligned}
		R(T) &\leq \sum_{i=1}^{d} \sum_{t=1}^{T} \frac{( \theta_{t,i} - \theta_{,i}^*)^2}{2 \alpha_{t}K_{t,i}}  + \sum_{i=1}^{d} \sum_{t=1}^{T} \frac{1 - K_{t,i}}{K_{t,i}(1-\beta_1^t)} g_{t,i} ( \theta_{t,i} - \theta_{,i}^* ) + \sum_{i=1}^{d} \sum_{t=1}^{T} \frac{\alpha_t}{K_{t,i}} \left( (1-K_{t,i})^{2} \widehat{m}_{t,i}^{2} + K_{t,i}^{2} g_{t,i}^{2} \right)
	\end{aligned}
\end{equation}

Using Lemma~\ref{lem.B.5} and Lemma~\ref{lem.B.6} From $\sum_{t=1}^{T} \widehat{s}_t > \sum_{t=1}^{T} (g_t - \widehat{m}_t)^2$, we have $\frac{1}{T} \sum_{t=1}^{T} K_t > \frac{1}{2}$. Therefore, from the assumption, \(\Vert \theta_t - \theta^* \Vert_2^2 \leq D,\Vert \theta_m - \theta_n \Vert_\infty \leq D_\infty\), we have the following regret bound:
\begin{equation}
	\begin{aligned}
		R(T) &\leq \frac{D^2}{\alpha} \sum_{i=1}^{d}  \sqrt{T} + \frac{2 D_\infty G_{\infty}}{1-\beta_{1}} \sum_{i=1}^{d}\left\Vert g_{1: T, i} \right\Vert_{2}   + \frac{2\alpha G_{\infty}^{2}(1 + (1 - \beta_{1})^2)}{\sqrt{T}(1-\beta_{1})^{2}} \sum_{i=1}^{d} \left\Vert g_{1: T, i} \right\Vert_{2}^{2}
	\end{aligned}
\end{equation}
    
\end{proof}

\newpage
\section{Convergence analysis for non-convex stochastic optimization (Theorem 3.3 in main paper).}
\label{non-convex}
We have relaxed the assumption on the objective function, allowing it to be non-convex, and adjusted the criterion for convergence from the statistic $R(T)$ to $\mathbb{E}(T)$. Let's briefly review the assumptions and the criterion for convergence after relaxing the assumption:

\begin{assumption}
\label{nonassume}
	\
\begin{itemize}
    \item A1 Bounded variables (same as convex). $\left\Vert \theta-\theta^{*}\right\Vert _{2}\leq D,\,\,\forall\theta,\theta^{*}$ or for any dimension $i$ of the variable, $\left\Vert \theta_{i}-\theta_{i}^{*}\right\Vert _{2}\leq D_{i},\,\,\forall\theta_{i},\theta_{i}^{*}$
    
    \item A2 The noisy gradient is unbiased. For $\forall t$, the random variable $\zeta_{t} $ is defined as $\zeta_{t}=g_{t}-\nabla f\left(\theta_{t}\right)$, $\zeta_{t}$ satisfy $\mathbb{E}\left[\zeta_{t}\right]=0$, $\mathbb{E}\left[\left\Vert \zeta_{t}\right\Vert _{2}^{2}\right]\leq\sigma^{2} $, and when $t_{1}\neq t_{2}$, $\zeta_{t_{1}}$ and $\zeta_{t_{2}}$ are statistically independent, i.e., $\zeta_{t_{1}} \perp \zeta_{t_{2}}$.
    
    \item A3 Bounded gradient and noisy gradient. At step $t$, the algorithm can access a bounded noisy gradient, and the true gradient is also bounded. $i.e.\ \  \vert  \vert \nabla f(\theta_t ) \vert  \vert \leq G,\  \vert  \vert g_t \vert  \vert \leq G,\ \ \forall t > 1$.
    
	\item A4 The property of function. The objective function $f\left(\theta\right)$ is a global loss function, defined as $f\left(\theta\right)=\lim_{T\longrightarrow\infty}\frac{1}{T}\sum_{t=1}^{T}f_{t}\left(\theta\right)$. Although $f\left(\theta\right)$ is no longer a convex function, it must still be a $L$-smooth function, i.e., it satisfies (1) $f$ is differentiable, $\nabla f $ exists everywhere in the domain; (2) there exists $L>0$ such that for any $\theta_{1}$ and $\theta_{2}$ in the domain, (first definition)
	\begin{equation}
	f\left(\theta_{2}\right)\leq f\left(\theta_{1}\right)+\left\langle \nabla f\left(\theta_{1}\right),\theta_{2}-\theta_{1}\right\rangle +\frac{L}{2}\left\Vert \theta_{2}-\theta_{1}\right\Vert _{2}^{2}
	\end{equation}
	or (second definition)
	\begin{equation}
	\left\Vert \nabla f\left(\theta_{1}\right)-\nabla f\left(\theta_{2}\right)\right\Vert _{2}\leq L\left\Vert \theta_{1}-\theta_{2}\right\Vert _{2}
	\end{equation}
	This condition is also known as \textit{L - Lipschitz}.
\end{itemize}
\end{assumption}

\begin{definition}
	The criterion for convergence is the statistic $\mathbb{E}\left(T\right)$:
	\begin{equation}
		\mathbb{E}\left(T\right)=\min_{t=1,2,\ldots,T}\mathbb{E}_{t-1}\left[\left\Vert \nabla f\left(\theta_{t}\right)\right\Vert _{2}^{2}\right]
	\end{equation}
	
	When $T\rightarrow\infty$, if the amortized value of $\mathbb{E}\left(T\right)$, $\mathbb{E}\left(T\right)/T\rightarrow0$, we consider such an algorithm to be convergent, and generally, the slower $\mathbb{E}\left(T\right)$ grows with $T$, the faster the algorithm converges.
\end{definition}

\begin{definition}
	Define $\xi_t$ as
	\begin{equation}
		\xi_t =
		\begin{cases} 
			\theta_t & t = 1\\ 
			\theta_t + \frac{\beta_{1}}{1 - \beta_{1}} \left(\theta_t - \theta_{t-1}\right) & t \geq 2 
		\end{cases}
	\end{equation}
\end{definition}

\begin{lemma}
\label{non-assume}
Let \( f \) be an \( L \)-smooth function. Then, for any points \( \xi_t \) and \( \theta_t \), the following inequality holds:
\begin{equation}
    f\left(\xi_{t+1}\right) - f\left(\xi_t\right) \leq \frac{L}{2} \left\Vert \xi_t - \theta_t \right\Vert_2^2 + L \left\Vert \xi_{t+1} - \xi_t \right\Vert_2^2 + \left\langle \nabla f\left(\theta_t\right), \xi_{t+1} - \xi_t \right\rangle
\end{equation}
\end{lemma}

\begin{proof}

Since $f$ is an L-smooth function, 
\begin{equation}
	\left\Vert \nabla f\left(\xi_t\right)-\nabla f\left(\theta_t\right)\right\Vert _{2}^{2}\leq L^{2}\left\Vert \xi_t-\theta_t\right\Vert _{2}^{2}
\end{equation}

Thus,
\begin{equation}
	\begin{aligned} 
		f\left(\xi_{t+1}\right) &- f\left(\xi_t\right) \\ 
		\leq & \left\langle \nabla f\left(\xi_t\right), \xi_{t+1} - \xi_t\right\rangle + \frac{L}{2} \left\Vert \xi_{t+1} - \xi_t\right\Vert_{2}^{2} \\ 
		= & \left\langle \frac{1}{\sqrt{L}}\left(\nabla f\left(\xi_t\right) - \nabla f\left(\theta_t\right)\right), \sqrt{L}\left(\xi_{t+1} - \xi_t\right)\right\rangle + \left\langle \nabla f\left(\theta_t\right), \xi_{t+1} - \xi_t\right\rangle + \frac{L}{2} \left\Vert \xi_{t+1} - \xi_t\right\Vert_{2}^{2} \\ 
		\leq & \frac{1}{2}\left(\frac{1}{L}\left\Vert \nabla f\left(\xi_t\right) - \nabla f\left(\theta_t\right)\right\Vert_{2}^{2} + L\left\Vert \xi_{t+1} - \xi_t\right\Vert_{2}^{2}\right) + \left\langle \nabla f\left(\theta_t\right), \xi_{t+1} - \xi_t\right\rangle + \frac{L}{2} \left\Vert \xi_{t+1} - \xi_t\right\Vert_{2}^{2} \\ 
		\leq & \frac{1}{2L}\left\Vert \nabla f\left(\xi_t\right) - \nabla f\left(\theta_t\right)\right\Vert_{2}^{2} + L\left\Vert \xi_{t+1} - \xi_t\right\Vert_{2}^{2} + \left\langle \nabla f\left(\theta_t\right), \xi_{t+1} - \xi_t\right\rangle \\ 
		\leq & \frac{1}{2L}L^{2}\left\Vert \xi_t - \theta_t\right\Vert_{2}^{2} + L\left\Vert \xi_{t+1} - \xi_t\right\Vert_{2}^{2} + \left\langle \nabla f\left(\theta_t\right), \xi_{t+1} - \xi_t\right\rangle \\ 
		= & \frac{L}{2}\underset{\left(1\right)}{\underbrace{\left\Vert \xi_t-\theta_t\right\Vert _{2}^{2}}}+L\underset{\left(2\right)}{\underbrace{\left\Vert \xi_{t+1}-\xi_t\right\Vert _{2}^{2}}}+\underset{\left(3\right)}{\underbrace{\left\langle \nabla f\left(\theta_t\right),\xi_{t+1}-\xi_t\right\rangle }}
	\end{aligned}
\end{equation}

\end{proof}

\begin{theorem}
	Consider a non-convex optimization problem. Suppose assumptions~\ref{nonassume} are satisfied, and let \( \alpha_t = \alpha/\sqrt{t} \). For all \( T \geq 1 \), SGDF achieves the following guarantee:
	\begin{equation}
		\mathbb{E}(T) \leq \frac{C_{7}\alpha^2 (\log T + 1) + C_{8}}{2\alpha\sqrt{T}}
	\end{equation}
	where $\mathbb{E}(T) =\min_{t=1,2,\ldots,T}\mathbb{E}_{t-1}\left[\left\Vert \nabla f\left(\theta_{t}\right)\right\Vert _{2}^{2}\right]$ denotes the minimum of the squared-paradigm expectation of the gradient, $\alpha$ is the learning rate at the $1$-th step, $C_{7}$ are constants independent of $d$ and $T$, $C_{8}$ is a constant independent of $T$, and the expectation is taken w.r.t all randomness corresponding to ${g_{t}}$.
\end{theorem}

\begin{proof}

According to Lemma~\ref{non-assume}, we deal with the three terms (1), (2), and (3) separately.

\textbf{For term (1)}

When $t=1$, $\left\Vert \xi_{t}-\theta_{t}\right\Vert _{2}^{2}=0$

When $t\geq2$,
\begin{equation}
\begin{aligned} 
	& \left\Vert \xi_{t}-\theta_{t}\right\Vert _{2}^{2}=\left\Vert \frac{\beta_{1}}{1-\beta_{1}}\left(\theta_{t}-\theta_{t-1}\right)\right\Vert _{2}^{2}\\ 
	& =\frac{\beta_{1}^{2}}{\left(1-\beta_{1}\right)^{2}}\alpha_{t-1}^{2}\left\Vert \widehat{g}_{t-1,i}\right\Vert _{2}^{2}\\ 
	&=\frac{\beta_{1}^{2}}{\left(1-\beta_{1}\right)^{2}}\alpha_{t-1}^{2}\sum_{i=1}^{d}\left(1-K_{t}\right)\left(\widehat{m}_{t-1,i}\right)^{2}+K_{t} g_{t}^2\\ 
	&\overset{\left(a\right)}{\leq}\frac{\beta_{1}^{2}}{\left(1-\beta_{1}\right)^{2}}\alpha_{t-1}^{2}\sum_{i=1}^{d}G_{i}^{2}
\end{aligned}
\end{equation}

Where (a) holds because for any $t$:
\begin{itemize}
	\item \( \left|\widehat{m}_{t,i}\right|\leq\frac{1}{1-\beta_{1}^{t}}\sum_{s=1}^{t}\left(1-\beta_{1}\right)\beta_{1}^{t-s}\left|g_{s,i}\right|\leq\frac{1}{1-\beta_{1}^{t}}\sum_{s=1}^{t}\left(1-\beta_{1}\right)\beta_{1}^{t-s}G_{i}=G_{i} 
	\).
	\item $\| g_{t} \|_2 \leq G,\, \forall t$, or for any dimension of the variable $i$: $\| g_{t,i} \|_2 \leq G_{i},\, \forall t$
\end{itemize}

\textbf{For term (2)}

When $t=1$,
\begin{equation}
\begin{aligned}
	\xi_{t+1}-\xi_t=&\theta_{t+1}+\frac{\beta_{1}}{1-\beta_{1}}\left(\theta_{t+1}-\theta_t\right)-\theta_t\\ 
	= & \frac{1}{1-\beta_{1}}\left(\theta_{t+1}-\theta_t\right)\\ 
	= & -\frac{\alpha_t}{1-\beta_{1}}\left(\widehat{g}_t\right)\\ 
	= & -\frac{\alpha_t}{1-\beta_{1}} \left(\frac{1-K_t}{1-\beta_{1}^{t}}m_t+K_{t} g_{t}\right)\\ 
	= &-\frac{\alpha_t}{1-\beta_{1}}\frac{1-K_t}{1-\beta_{1}^{t}}\left(\beta_{1}\cancelto{0}{m_{t-1}}+\left(1-\beta_{1}\right)g_{t}\right)-\frac{\alpha_t}{1-\beta_{1}}K_{t} g_{t}\\ 
	= & -\frac{\alpha_t\left(1-K_t\right)}{1-\beta_{1}^{t}}g_{t} -\frac{\alpha_tK_{t}}{1-\beta_{1}} g_{t}\\
	= & -\frac{\alpha_t}{1-\beta_{1}}g_{t} 
\end{aligned}
\end{equation}

Thus,
\begin{equation}
\begin{aligned}
	\left\Vert \xi_{t+1}-\xi_{t}\right\Vert _{2}^{2}= & \left\Vert -\frac{\alpha_{t}(1-K_t)}{1-\beta_{1}}g_{t} -\frac{\alpha_{t}K_{t}}{1-\beta_{1}} g_{t}\right\Vert _{2}^{2}\\ 
	= & \left( -\frac{\alpha_{t}}{1-\beta_{1}} \right)^2 \|g_{t}\|_2^2 \\
	= & \frac{\alpha_{t}^2}{(1-\beta_{1})^2} \|g_{t}\|_2^2\\
	= & \frac{\alpha_{t}^{2}}{\left(1-\beta_{1}\right)^{2}}\sum_{i=1}^{d}g_{t,i}^{2}\\ 
	\leq & \frac{\alpha_{t}^{2}}{\left(1-\beta_{1}\right)^{2}}\sum_{i=1}^{d}G_{i}^{2}
\end{aligned}
\end{equation}

When $t\geq2$,
\begin{equation}
\begin{aligned}
	\xi_{t+1}-\xi_t=&\theta_{t+1}+\frac{\beta_{1}}{1-\beta_{1}}\left(\theta_{t+1}-\theta_t\right) -\theta_t-\frac{\beta_{1}}{1-\beta_{1}}\left(\theta_t-\theta_{t-1}\right)\\ 
	=&\frac{1}{1-\beta_{1}}\left(\theta_{t+1}-\theta_t\right)-\frac{\beta_{1}}{1-\beta_{1}}\left(\theta_t-\theta_{t-1}\right) 
\end{aligned}
\end{equation}

Due to
\begin{equation}
\begin{aligned}
	\theta_{t+1}-\theta_t= & -\alpha_t\widehat{g}_t\\ 
	=& -\frac{\alpha_t(1-K_{t})}{1-\beta_{1}^{t}}m_t - \alpha_tK_{t}g_{t} \\ 
	=&-\frac{\alpha_t(1-K_{t})}{1-\beta_{1}^{t}}\left(\beta_{1}m_{t-1}+\left(1-\beta_{1}\right)g_{t}\right)- \alpha_tK_{t}g_{t}
\end{aligned}
\end{equation}

So,
\begin{equation}
\begin{aligned} 
	& \xi_{t+1}-\xi_t\\ = & \frac{1}{1-\beta_{1}}\left(-\frac{\alpha_t(1-K_{t})}{1-\beta_{1}^{t}}\left(\beta_{1}m_{t-1}+\left(1-\beta_{1}\right)g_{t}\right)- \alpha_tK_{t}g_{t}\right) -\frac{\beta_{1}}{1-\beta_{1}}\left(-\frac{\alpha_{t-1}(1-K_{t-1})}{1-\beta_{1}^{t-1}}m_{t-1}- \alpha_{t-1}K_{t-1}g_{t-1}\right)\\ 
	=&-\frac{\beta_{1}}{1-\beta_{1}}m_{t-1}\odot\left(\frac{\alpha_t(1-K_{t})}{1-\beta_{1}^{t}}-\frac{\alpha_{t-1}(1-K_{t-1})}{1-\beta_{1}^{t-1}}\right)-\frac{\alpha_t(1-K_{t})}{1-\beta_{1}^{t}}g_{t} -\frac{\alpha_tK_{t}}{1-\beta_{1}}g_{t} + \frac{\beta_{1}}{1-\beta_{1}}\alpha_{t-1}K_{t-1}g_{t-1}\\
	=&-\frac{\beta_{1}}{1-\beta_{1}}m_{t-1}\odot\left(\frac{\alpha_t(1-K_{t})}{1-\beta_{1}^{t}}-\frac{\alpha_{t-1}(1-K_{t-1})}{1-\beta_{1}^{t-1}}\right)-\left(\frac{\alpha_t(1-K_{t})}{1-\beta_{1}^{t}} + \frac{\alpha_tK_{t}}{1-\beta_{1}}\right)g_{t}  + \frac{\beta_{1}\alpha_{t-1}K_{t-1}}{1-\beta_{1}}g_{t-1}
\end{aligned}
\end{equation}

We have:
\begin{equation}
\begin{aligned} 
	\left\Vert \xi_{t+1}-\xi_t\right\Vert _{2}^{2}&\leq2\left\Vert -\frac{\beta_{1}}{1-\beta_{1}}m_{t-1}\odot\Bigg(\frac{\alpha_t(1-K_{t})}{1-\beta_{1}^{t}}\right. \left.-\frac{\alpha_{t-1}(1-K_{t-1})}{1-\beta_{1}^{t-1}}\Bigg)\right\Vert _{2}^{2}\\
	&+2\left\Vert -\left(\frac{\alpha_t(1-K_{t})}{1-\beta_{1}^{t}} + \frac{\alpha_tK_{t}}{1-\beta_{1}}\right)g_{t}\right\Vert _{2}^{2}+2\left\Vert \frac{\beta_{1}\alpha_{t-1}K_{t-1}}{1-\beta_{1}}g_{t-1} \right\Vert_{2}^{2}\\ 
	& \leq2\frac{\beta_{1}^{2}}{\left(1-\beta_{1}\right)^{2}}\left\Vert m_{t-1}\right\Vert _{\infty}^{2}\left\Vert \frac{\alpha_t(1-K_{t})}{1-\beta_{1}^{t}}-\frac{\alpha_{t-1}(1-K_{t-1})}{1-\beta_{1}^{t-1}}\right\Vert _{\infty}\cdot\left\Vert\frac{\alpha_t(1-K_{t})}{1-\beta_{1}^{t}}-\frac{\alpha_{t-1}(1-K_{t-1})}{1-\beta_{1}^{t-1}}\right\Vert _{1}\\
	&+2\left\Vert -\left(\frac{\alpha_t(1-K_{t})}{1-\beta_{1}^{t}} + \frac{\alpha_tK_{t}}{1-\beta_{1}}\right)g_{t}\right\Vert _{2}^{2}+2\left\Vert \frac{\beta_{1}\alpha_{t-1}K_{t-1}}{1-\beta_{1}}g_{t-1} \right\Vert_{2}^{2}\\ 
\end{aligned}
\end{equation}

Because
\begin{itemize}
	\item $\left|m_{t-1,i}\right|=\left(1-\beta_{1}^{t}\right)\left|\widehat{m}_{t,i}\right|\leq\left|\widehat{m}_{t,i}\right|\leq G_{i}$, $\left\Vert m_{t-1}\right\Vert {\infty}^{2}\leq\left(\max{i}G_{i}\right)^{2}$
	
	\item $\left\Vert g_{t}\right\Vert_{2}^{2}=\sum_{i=1}^{d}g_{t,i}^{2}\leq\sum_{i=1}^{d}G_{i}^{2}$
	
	\item $K_t \in {0,1}^d$, we have $\left\Vert K_t\right\Vert_{\infty} \leq \sum_{i=1}^{d} \textbf{1}_i, \left\Vert 1 - K_t\right\Vert{\infty}\leq \sum_{i=1}^{d} \textbf{1}_i \leq d$
\end{itemize}

\begin{equation}
\begin{aligned} 
	&\alpha_{t}/\left(1-\beta_{1}^{t}\right)\geq0,\thinspace\thinspace\alpha_{t-1}/\left(1-\beta_{1}^{t-1}\right)/\geq0\\ 
	&\alpha_{t}\leq\alpha_{t-1},\thinspace\thinspace\frac{1}{1-\beta_{1}^{t}}\leq\frac{1}{1-\beta_{1}^{t-1}}\thinspace\\ 
	\Longrightarrow &\frac{\alpha_{t}}{1-\beta_{1}^{t}}\leq\frac{\alpha_{t-1}}{1-\beta_{1}^{t-1}}\\ 
	\Longrightarrow&\left|\frac{\alpha_{t}}{1-\beta_{1}^{t}}-\frac{\alpha_{t-1}}{1-\beta_{1}^{t-1}}\right|\\ 
	&=\alpha_{t-1}/\left(1-\beta_{1}^{t-1}\right)-\alpha_{t}/\left(1-\beta_{1}^{t}\right)\\ 
	& \leq\alpha_{t-1}/\left(1-\beta_{1}^{t-1}\right)\leq\alpha_{1}/\left(1-\beta_{1}\right)\\ \Longrightarrow & \left\Vert \frac{\alpha_{t}\left(1-K_{t}\right)}{1-\beta_{1}^{t}}-\frac{\alpha_{t-1}\left( 1-K_{t-1}\right)}{1-\beta_{1}^{t-1}}\right\Vert _{\infty} \leq\frac{\alpha_{1}}{\left(1-\beta_{1}\right)} 
\end{aligned}
\end{equation}

\begin{equation}
    \left\Vert \frac{\alpha_{t}\left(1-K_{t}\right)}{1-\beta_{1}^{t}}-\frac{\alpha_{t-1}\left(1-K_{t-1}\right)}{1-\beta_{1}^{t-1}}\right\Vert _{1}\leq \sum_{i=1}^{d}\left(\alpha_{t-1} / \left(1-\beta_{1}^{t-1}\right)-\alpha_{t} / \left(1-\beta_{1}^{t}\right)\right)\textbf{1}_i \leq d\left(\alpha_{t-1}/\left(1-\beta_{1}^{t-1}\right)-\alpha_{t}/\left(1-\beta_{1}^{t}\right)\right)
\end{equation}

Therefore
\begin{equation}
\begin{aligned} 
	 \left\Vert \xi_{t+1}-\xi_{t}\right\Vert _{2}^{2}
	\leq&2\frac{\beta_{1}^{2}}{\left(1-\beta_{1}\right)^{2}}\left(\max_{i}G_{i}\right)^{2}\frac{d\alpha_{1}}{\left(1-\beta_{1}\right)}\cdot\left(\frac{\alpha_{t-1}}{\left(1-\beta_{1}^{t-1}\right)}-\frac{\alpha_{t}}{\left(1-\beta_{1}^{t}\right)}\right)+4\frac{\alpha_{t}^{2}}{\left(1-\beta_{1}\right)^{2}}\sum_{i=1}^{d}G_{i}^{2}
\end{aligned}
\end{equation}

\textbf{For term (3)}

When $t=1$, referring to the case of $t=1$ in the previous subsection,
\begin{equation}
\begin{aligned} 
	 \left\langle \nabla f\left(\theta_t\right),\xi_{t+1}-\xi_t\right\rangle 
	= & \left\langle \nabla f\left(\theta_t\right),-\frac{\alpha_t}{1-\beta_{1}}g_t\right\rangle \\ 
	= & \left\langle \nabla f\left(\theta_t\right),-\frac{\alpha_t}{1-\beta_{1}}\nabla f\left(\theta_t\right)\right\rangle  +\left\langle \nabla f\left(\theta_t\right),-\frac{\alpha_t}{1-\beta_{1}}\zeta_t\right\rangle 
\end{aligned}
\end{equation}

The last equality is due to the definition of $g_{t}$: $g_{t}=\nabla f\left(\theta_{t}\right)+\zeta_{t}$. Let's consider them separately:
\begin{equation}
\begin{aligned} 
	 \left\langle \nabla f\left(\theta_t\right),-\frac{\alpha_t}{1-\beta_{1}}\nabla f\left(\theta_t\right)\right\rangle 
	= & -\frac{\alpha_t}{1-\beta_{1}}\left[\nabla f\left(\theta_t\right)\right]\left[\nabla f\left(\theta_t\right)\right]\\ 
	\leq & -\frac{\alpha_t}{1-\beta_{1}}\left\Vert\nabla f\left(\theta_t\right)\right\Vert _{2}^{2}
\end{aligned}
\end{equation}

\begin{equation}
\begin{aligned} 
	 \left\langle \nabla f\left(\theta_t\right),-\frac{\alpha_t}{1-\beta_{1}}\zeta_t\right\rangle 
	\leq & \frac{\alpha_t}{1-\beta_{1}}\left\Vert \nabla f\left(\theta_t\right)\right\Vert _{2}\left\Vert \zeta_t\right\Vert _{2} \\ 
	= & \frac{\alpha_t}{1-\beta_{1}}\left\Vert \nabla f\left(\theta_t\right)\right\Vert _{2}\left\Vert g_t-\nabla f\left(\theta_t\right)\right\Vert _{2}\\ 
	\leq & \frac{\alpha_t}{1-\beta_{1}}\cdot2 \sum_{i=1}^{d} G_{i}^{2}
\end{aligned}
\end{equation}

Thus
\begin{equation}
\begin{aligned} 
	& \left\langle \nabla f\left(\theta_t\right),\xi_{t+1}-\xi_t\right\rangle \\ 
	\leq & -\frac{\alpha_t}{\left(1-\beta_{1}\right)}\left\Vert \nabla f\left(\theta_t\right)\right\Vert _{2}^{2} +\frac{2\alpha_t}{1-\beta_{1}}\cdot \sum_{i=1}^{d} G_{i}^{2}
\end{aligned}
\end{equation}

When $t\geq2$,
\begin{equation}
\begin{aligned} 
	 \left\langle \nabla f\left(\theta_t\right),\xi_{t+1}-\xi_t\right\rangle 
	= & \left\langle \nabla f\left(\theta_t\right),-\frac{\beta_{1}}{1-\beta_{1}}m_{t-1}\odot\right.\left.\left(\frac{\alpha_t(1-K_{t})}{1-\beta_{1}^{t}}-\frac{\alpha_{t-1}(1-K_{t-1})}{1-\beta_{1}^{t-1}}\right)\right\rangle \\ 
	& +\left\langle \nabla f\left(\theta_t\right),-\left(\frac{\alpha_t(1-K_{t})}{1-\beta_{1}^{t}} + \frac{\alpha_tK_{t}}{1-\beta_{1}}\right)\nabla f\left(\theta_t\right)\right\rangle +\left\langle \nabla f\left(\theta_t\right),-\left(\frac{\alpha_t(1-K_{t})}{1-\beta_{1}^{t}} + \frac{\alpha_tK_{t}}{1-\beta_{1}}\right)\zeta_t\right\rangle \\
	& +\left\langle \nabla f\left(\theta_{t-1}\right),\frac{\beta_{1}\alpha_{t-1}K_{t-1}}{1-\beta_{1}}\nabla f\left(\theta_{t-1}\right)\right\rangle +\left\langle \nabla f\left(\theta_{t-1}\right),\frac{\beta_{1}\alpha_{t-1}K_{t-1}}{1-\beta_{1}}\zeta_{t-1}\right\rangle
\end{aligned}
\end{equation}

Start by looking at the first item after the equal sign:
\begin{equation}
\begin{aligned} 
	& \left\langle \nabla f\left(\theta_t\right),-\frac{\beta_{1}}{1-\beta_{1}}m_{t-1}\odot\right.\left.\left(\frac{\alpha_t(1-K_{t})}{1-\beta_{1}^{t}}-\frac{\alpha_{t-1}(1-K_{t-1})}{1-\beta_{1}^{t-1}}\right)\right\rangle \\ 
	\leq & \frac{\beta_{1}}{1-\beta_{1}}\left\Vert \nabla f\left(\theta_t\right)\right\Vert _{\infty}\left\Vert m_{t-1}\right\Vert _{\infty}\cdot \left\Vert \frac{\alpha_t(1-K_{t})}{1-\beta_{1}^{t}}-\frac{\alpha_{t-1}(1-K_{t-1})}{1-\beta_{1}^{t-1}}\right\Vert _{1}\\ 
	\leq & \frac{\beta_{1}}{1-\beta_{1}}\left(\max_{i}G_{i}\right)\left(\max_{i}G_{i}\right)\cdot \sum_{i=1}^{d}\left(\frac{\alpha_{t-1}}{\left(1-\beta_{1}^{t-1}\right)}-\frac{\alpha_{t}}{\left(1-\beta_{1}^{t}\right)}\right)\textbf{1}_i \\
	\leq & \frac{\beta_{1}}{1-\beta_{1}}\left(\max_{i}G_{i}\right)\left(\max_{i}G_{i}\right)\cdot d\left(\frac{\alpha_{t-1}}{\left(1-\beta_{1}^{t-1}\right)}-\frac{\alpha_{t}}{\left(1-\beta_{1}^{t}\right)}\right)
\end{aligned}
\end{equation}

The second and third terms after the equal sign:
\begin{equation}
\begin{aligned}
	&\left\langle \nabla f\left(\theta_t\right),-\left(\frac{\alpha_t(1-K_{t})}{1-\beta_{1}^{t}} + \frac{\alpha_tK_{t}}{1-\beta_{1}}\right)\nabla f\left(\theta_t\right)\right\rangle +\left\langle \nabla f\left(\theta_t\right),-\left(\frac{\alpha_t(1-K_{t})}{1-\beta_{1}^{t}} + \frac{\alpha_tK_{t}}{1-\beta_{1}}\right)\zeta_t\right\rangle\\
	\leq &-\frac{\alpha_t}{1-\beta_{1}^{t}} \left\Vert \nabla f\left(\theta_t\right)\right\Vert _{2}^{2} +\left\langle \nabla f\left(\theta_t\right), -\frac{\alpha_t}{1-\beta_{1}^{t}}\zeta_t\right\rangle
\end{aligned}
\end{equation}

The fourth and fifth terms after the equal sign:
\begin{equation}
\begin{aligned} 
	& \left\langle \nabla f\left(\theta_{t-1}\right),\frac{\beta_{1}\alpha_{t-1}K_{t-1}}{1-\beta_{1}}\nabla f\left(\theta_{t-1}\right)\right\rangle +\left\langle \nabla f\left(\theta_{t-1}\right),\frac{\beta_{1}\alpha_{t-1}K_{t-1}}{1-\beta_{1}}\zeta_{t-1}\right\rangle \\ 
	\leq &\frac{\beta_{1}\alpha_{t-1}}{1-\beta_{1}} \left\Vert \nabla f\left(\theta_t\right)\right\Vert _{\infty} \left\Vert \nabla f\left(\theta_t\right)\right\Vert _{\infty}\left\Vert \textbf{1}_i\right\Vert _{1} + \frac{\beta_{1}\alpha_{t-1}}{1-\beta_{1}} \left\Vert \nabla f\left(\theta_t\right)\right\Vert _{\infty} \left\Vert \zeta_t\right\Vert _{\infty}\left\Vert \textbf{1}_i\right\Vert _{1}\\
	\leq & \frac{\beta_{1}\alpha_{t-1}}{1-\beta_{1}} \left(\max_{i}G_{i}\right)\left(\max_{i}G_{i}\right)\sum_{i=1}^{d}\textbf{1}_i + \frac{\beta_{1}\alpha_{t-1}}{1-\beta_{1}} \left(\max_{i}G_{i}\right)\left(2\max_{i}G_{i}\right)\sum_{i=1}^{d}\textbf{1}_i\\
	\leq & \frac{\beta_{1}\alpha_{t-1}}{1-\beta_{1}} \left(\max_{i}G_{i}\right)\left(\max_{i}G_{i}\right)d + \frac{\beta_{1}\alpha_{t-1}}{1-\beta_{1}} \left(\max_{i}G_{i}\right)\left(2\max_{i}G_{i}\right)d
\end{aligned}
\end{equation}

Final:
\begin{equation}
\begin{aligned} 
	& \left\langle \nabla f\left(\theta_{t}\right),\xi_{t+1}-\xi_{t}\right\rangle \\ 
	\leq & \frac{\beta_{1}}{1-\beta_{1}}\left(\max_{i}G_{i}\right)\left(\max_{i}G_{i}\right)\cdot d\left(\frac{\alpha_{t-1}}{\left(1-\beta_{1}^{t-1}\right)}-\frac{\alpha_{t}}{\left(1-\beta_{1}^{t}\right)}\right) -\frac{\alpha_{t}}{\left(1-\beta_{1}^{t}\right)}\left\Vert \nabla f\left(\theta_{t}\right)\right\Vert _{2}^{2}\\ 
	&+\frac{\beta_{1}\alpha_{t-1}}{1-\beta_{1}} \left(\max_{i}G_{i}\right)\left(\max_{i}G_{i}\right)d + \frac{\beta_{1}\alpha_{t-1}}{1-\beta_{1}} \left(\max_{i}G_{i}\right)\left(2\max_{i}G_{i}\right)d +\left\langle \nabla f\left(\theta_{t}\right),-\frac{\alpha_{t}}{1-\beta_{1}^{t}}\zeta_{t}\right\rangle 
\end{aligned}
\end{equation}

\textbf{Summarizing the results}

Let's start summarizing: when $t=1$,
\begin{equation}
\begin{aligned} 
	f\left(\xi_{t+1}\right)-f\left(\xi_t\right)
	\leq & \frac{L}{2}\cdot 0 + L \cdot \frac{\alpha_t^{2}}{\left(1-\beta_{1}\right)^{2}}\sum_{i=1}^{d}G_{i}^{2} -\frac{\alpha_t}{\left(1-\beta_{1}\right)}\left\Vert \nabla f\left(\theta_t\right)\right\Vert _{2}^{2} +\frac{2\alpha_t}{1-\beta_{1}}\cdot \sum_{i=1}^{d} G_{i}^{2}
\end{aligned}
\end{equation}

Taking the expectation over the random distribution of $\zeta_{1},\zeta_{2},\ldots,\zeta_{t}$ on both sides of the inequality:
\begin{equation}
\begin{aligned} 
	 \mathbb{E}_{t}\left[f\left(\xi_{t+1}\right)-f\left(\xi_t\right)\right]
	\leq &  L \cdot \frac{\alpha_t^{2}}{\left(1-\beta_{1}\right)^{2}}\sum_{i=1}^{d}G_{i}^{2} -\frac{\alpha_t}{\left(1-\beta_{1}\right)}\mathbb{E}_{t}\left\Vert \nabla f\left(\theta_t\right)\right\Vert _{2}^{2} +\frac{2\alpha_t}{1-\beta_{1}}\cdot \sum_{i=1}^{d} G_{i}^{2}
\end{aligned}
\end{equation}

When $t\geq2$,
\begin{equation}
\begin{aligned} 
	& f\left(\xi_{t+1}\right)-f\left(\xi_t\right)\\ 
	\leq & \frac{L}{2}\frac{\beta_{1}^{2}}{\left(1-\beta_{1}\right)^{2}}\alpha_{t-1}^{2}\sum_{i=1}^{d}G_{i}^{2}+L\cdot2\frac{\beta_{1}^{2}}{\left(1-\beta_{1}\right)^{2}}\left(\max_{i}G_{i}\right)^{2}\frac{d\alpha_{1}}{\left(1-\beta_{1}\right)}\cdot\left(\frac{\alpha_{t-1}}{\left(1-\beta_{1}^{t-1}\right)}-\frac{\alpha_{t}}{\left(1-\beta_{1}^{t}\right)}\right)\\ &+L\cdot4\frac{\alpha_{t}^{2}}{\left(1-\beta_{1}\right)^{2}}\sum_{i=1}^{d}G_{i}^{2} +\frac{\beta_{1}}{1-\beta_{1}}\left(\max_{i}G_{i}\right)\left(\max_{i}G_{i}\right)\cdot d\left(\frac{\alpha_{t-1}}{\left(1-\beta_{1}^{t-1}\right)}-\frac{\alpha_{t}}{\left(1-\beta_{1}^{t}\right)}\right)\\ 
	&-\frac{\alpha_{t}}{\left(1-\beta_{1}^{t}\right)}\left\Vert \nabla f\left(\theta_t\right)\right\Vert _{2}^{2}+\frac{\beta_{1}\alpha_{t-1}}{1-\beta_{1}} \left(\max_{i}G_{i}\right)\left(\max_{i}G_{i}\right)d + \frac{\beta_{1}\alpha_{t-1}}{1-\beta_{1}} \left(\max_{i}G_{i}\right)\left(2\max_{i}G_{i}\right)d \\ 
	&+\left\langle \nabla f\left(\theta_t\right),-\frac{\alpha_{t}}{1-\beta_{1}^{t}}\zeta_t\right\rangle 
\end{aligned}
\end{equation}

Taking the expectation over the random distribution of $\zeta_{1},\zeta_{2},\ldots,\zeta_{t}$ on both sides of the inequality:
\begin{equation}
\begin{aligned} 
	& \mathbb{E}_{t}\left[f\left(\xi_{t+1}\right)-f\left(\xi_t\right)\right]\\ 
	\leq & \frac{L}{2}\frac{\beta_{1}^{2}}{\left(1-\beta_{1}\right)^{2}}\alpha_{t-1}^{2}\sum_{i=1}^{d}G_{i}^{2}+L\cdot2\frac{\beta_{1}^{2}}{\left(1-\beta_{1}\right)^{2}}\left(\max_{i}G_{i}\right)^{2}\frac{d\alpha_{1}}{\left(1-\beta_{1}\right)}\cdot\left(\frac{\alpha_{t-1}}{\left(1-\beta_{1}^{t-1}\right)}-\frac{\alpha_{t}}{\left(1-\beta_{1}^{t}\right)}\right)\\ &+L\cdot4\frac{\alpha_{t}^{2}}{\left(1-\beta_{1}\right)^{2}}\sum_{i=1}^{d}G_{i}^{2} +\frac{\beta_{1}}{1-\beta_{1}}\left(\max_{i}G_{i}\right)\left(\max_{i}G_{i}\right)\cdot d\left(\frac{\alpha_{t-1}}{\left(1-\beta_{1}^{t-1}\right)}-\frac{\alpha_{t}}{\left(1-\beta_{1}^{t}\right)}\right)\\ 
	&-\frac{\alpha_{t}}{\left(1-\beta_{1}^{t}\right)}\mathbb{E}_{t}\left\Vert \nabla f\left(\theta_t\right)\right\Vert _{2}^{2}+\frac{\beta_{1}\alpha_{t-1}}{1-\beta_{1}} \left(\max_{i}G_{i}\right)\left(\max_{i}G_{i}\right)d + \frac{\beta_{1}\alpha_{t-1}}{1-\beta_{1}} \left(\max_{i}G_{i}\right)\left(2\max_{i}G_{i}\right)d \\ 
	&+\mathbb{E}_{t}\left\langle \nabla f\left(\theta_t\right),-\frac{\alpha_{t}}{1-\beta_{1}^{t}}\zeta_t\right\rangle  
\end{aligned}
\end{equation}

Since the value of $\theta_{t}$ is independent of $g_{t}$, they are statistically independent of $\zeta_{t}$:
\begin{equation}
\begin{aligned} 
	& \mathbb{E}_{t}\left[\left\langle \nabla f\left(\theta_{t}\right),-\frac{\alpha_{t}}{1-\beta_{1}^{t}}\zeta_{t}\right\rangle \right]\\ 
	= & \mathbb{E}_{t}\left[\left\langle -\frac{\alpha_{t}}{1-\beta_{1}^{t}}\nabla f\left(\theta_{t}\right),\zeta_{t}\right\rangle \right]\\ 
	= & \left\langle -\frac{\alpha_{t}}{1-\beta_{1}^{t}}\mathbb{E}_{t}\left[\nabla f\left(\theta_{t}\right)\right],\cancelto{0}{\mathbb{E}_{t}\left[\zeta_{t}\right]}\right\rangle =0 
\end{aligned}
\end{equation}

Summing up both sides of the inequality for $t=1,2,\ldots, T$:

•  Left side of the inequality (can be reduced to maintain the inequality)
\begin{equation}
\begin{aligned}
	\sum_{t=1}^{T}\textrm{LHS of the inequality}
	= & \sum_{t=1}^{T}\mathbb{E}_{t}\left[f\left(\xi_{t+1}\right)-f\left(\xi_t\right)\right]\\ 
	=&\sum_{t=1}^{T}\mathbb{E}_{t}\left[f\left(\xi_{t+1}\right)\right]-\mathbb{E}_{t}\left[f\left(\xi_t\right)\right]\\ 
	=&\sum_{t=1}^{T}\mathbb{E}_{t}\left[f\left(\xi_{t+1}\right)\right]-\mathbb{E}_{t-1}\left[f\left(\xi_t\right)\right]\\
	=&\mathbb{E}_{T}\left[f\left(\xi_{T+1}\right)\right]-\mathbb{E}_{0}\left[f\left(\xi_1\right)\right] 
\end{aligned}
\end{equation}

Since $f\left(\xi_{T+1}\right)\geq\min_{\theta}f\left(\theta\right)=f\left(\theta^{*}\right)$, $\xi_{1}=\theta_{1}$, and both are deterministic:
\begin{equation}
\begin{aligned}
	\sum_{t=1}^{T}\mathbb{E}_{t}\left[f\left(\xi_{t+1}\right)-f\left(\xi_t\right)\right]
	\geq&\mathbb{E}_{T}\left[f\left(\theta^{*}\right)\right]-\mathbb{E}_{0}\left[f\left(\theta_1\right)\right]\\ 
	= & f\left(\theta^{*}\right)-f\left(\theta_1\right) 
\end{aligned}
\end{equation}

• The right side of the inequality (can be enlarged to keep the inequality valid)

We perform a series of substitutions to simplify the symbols:

When $t>2$,
\begin{enumerate}
	\item $\frac{L}{2}\frac{\beta_{1}^{2}}{\left(1-\beta_{1}\right)^{2}}\alpha_{t-1}^{2}\sum_{i=1}^{d}G_{i}^{2} \triangleq C_{1}\alpha_{t-1}^{2}$
	\item $L\cdot2\frac{\beta_{1}^{2}}{\left(1-\beta_{1}\right)^{2}}\left(\max_{i}G_{i}\right)^{2}\frac{d\alpha_{1}}{\left(1-\beta_{1}\right)}\cdot\left(\frac{\alpha_{t-1}}{\left(1-\beta_{1}^{t-1}\right)}-\frac{\alpha_{t}}{\left(1-\beta_{1}^{t}\right)}\right) \triangleq C_{2}\left(\frac{\alpha_{t-1}}{\left(1-\beta_{1}^{t-1}\right)}-\frac{\alpha_{t}}{\left(1-\beta_{1}^{t}\right)}\right)$
	\item $L\cdot4\frac{\alpha_{t}^{2}}{\left(1-\beta_{1}\right)^{2}}\sum_{i=1}^{d}G_{i}^{2} \leq L\cdot4\frac{\alpha_{t}^{2}}{\left(1-\beta_{1}\right)^{2}}\sum_{i=1}^{d}G_{i}^{2} \triangleq C_{3}\alpha_{t}^{2}$
	\item $\frac{\beta_{1}}{1-\beta_{1}}\left(\max_{i}G_{i}\right)\left(\max_{i}G_{i}\right)\cdot d\left(\frac{\alpha_{t-1}}{\left(1-\beta_{1}^{t-1}\right)}-\frac{\alpha_{t}}{\left(1-\beta_{1}^{t}\right)}\right) \triangleq C_{4}\left(\frac{\alpha_{t-1}}{\left(1-\beta_{1}^{t-1}\right)}-\frac{\alpha_{t}}{\left(1-\beta_{1}^{t}\right)}\right)$
	\item $-\frac{\alpha_{t}}{\left(1-\beta_{1}^{t}\right)}\mathbb{E}_{t}\left[\left\Vert \nabla f\left(\theta_{t}\right)\right\Vert _{2}^{2}\right] \leq -\alpha_{t}\mathbb{E}_{t}\left[\left\Vert \nabla f\left(\theta_{t}\right)\right\Vert _{2}^{2}\right]$
	\item $\frac{\beta_{1}\alpha_{t-1}}{1-\beta_{1}} \left(\max_{i}G_{i}\right)\left(\max_{i}G_{i}\right)d + \frac{\beta_{1}\alpha_{t-1}}{1-\beta_{1}} \left(\max_{i}G_{i}\right)\left(2\max_{i}G_{i}\right)d \triangleq C_{5}\alpha_{t-1}$
\end{enumerate}

When $t=1$,
\begin{enumerate}
	\item $L \cdot \frac{\alpha_{t}^{2}}{\left(1-\beta_{1}\right)^{2}}\sum_{i=1}^{d}G_{i}^{2}\leq L\cdot4\frac{\alpha_{t}^{2}}{\left(1-\beta_{1}\right)^{2}}\sum_{i=1}^{d}G_{i}^{2}=C_{3}\alpha_{t}^{2}$
	\item $-\frac{\alpha_{t}}{\left(1-\beta_{1}\right)}\mathbb{E}_{t}\left[\left\Vert \nabla f\left(\theta_{t}\right)\right\Vert _{2}^{2}\right]\leq-\alpha_{t}\mathbb{E}_{t}\left[\left\Vert \nabla f\left(\theta_{t}\right)\right\Vert _{2}^{2}\right]$
	\item $\frac{2\alpha_{t}}{1-\beta_{1}}\cdot \sum_{i=1}^{d} G_{i}^{2}\triangleq C_{6} \alpha_{t}$
\end{enumerate}

After substitution,
\begin{equation}
\begin{aligned} 
	&\sum_{t=1}^{T}\textrm{RHS of the inequality}\leq\sum_{t=2}^{T}C_{1}\alpha_{t-1}^{2}+\sum_{t=1}^{T}C_{3}\alpha_{t}^{2}-\sum_{t=1}^{T}\alpha_{t}\mathbb{E}_{t}\left[\left\Vert \nabla f\left(\theta_{t}\right)\right\Vert _{2}^{2}\right]\\ 
	&+\sum_{t=2}^{T}\left(C_{2}+C_{4}\right)\left(\frac{\alpha_{t-1}}{\left(1-\beta_{1}^{t-1}\right)}-\frac{\alpha_{t}}{\left(1-\beta_{1}^{t}\right)}\right)+\sum_{t=1}^{T}C_{5}\alpha_{t-1}+\sum_{t=1}^{T}C_{6}\alpha_{t}\\ 
	&=\sum_{t=2}^{T}C_{1}\alpha_{t-1}^{2}+\sum_{t=1}^{T}C_{3}\alpha_{t}^{2}-\sum_{t=1}^{T}\alpha_{t}\mathbb{E}_{t}\left[\left\Vert \nabla f\left(\theta_{t}\right)\right\Vert _{2}^{2}\right]+\sum_{t=1}^{T}C_{5}\alpha_{t-1}+\sum_{t=1}^{T}C_{6}\alpha_{t}\\ 
	&+\sum_{i=1}^{d}\left(C_{2}+C_{4}\right)\sum_{t=2}^{T}\left(\frac{\alpha_{t-1}}{\left(1-\beta_{1}^{t-1}\right)}-\frac{\alpha_{t}}{\left(1-\beta_{1}^{t}\right)}\right)\\ 
	&=\sum_{t=2}^{T}C_{1}\alpha_{t-1}^{2}+\sum_{t=1}^{T}C_{3}\alpha_{t}^{2}-\sum_{t=1}^{T}\alpha_{t}\mathbb{E}_{t}\left[\left\Vert \nabla f\left(\theta_{t}\right)\right\Vert _{2}^{2}\right]+\sum_{t=1}^{T}C_{5}\alpha_{t-1}+\sum_{t=1}^{T}C_{6}\alpha_{t}\\
	&+\sum_{i=1}^{d}\left(C_{2}+C_{4}\right)\left(\frac{\alpha_{1}}{\left(1-\beta_{1}\right)}-\frac{\alpha_{T}}{\left(1-\beta_{1}^{T}\right)}\right)\\ 
	& \leq\left(C_{1}+C_{3}+C_{5}+C_{6}\right)\sum_{t=1}^{T}\alpha_{t}^{2}-\sum_{t=1}^{T}\alpha_{t}\mathbb{E}_{t}\left[\left\Vert \nabla f\left(\theta_{t}\right)\right\Vert _{2}^{2}\right] +\sum_{i=1}^{d}\left(C_{2}+C_{4}\right)\frac{\alpha_{1}}{\left(1-\beta_{1}\right)}\\ 
	& \leq\left(C_{1}+C_{3}+C_{5}+C_{6}\right)\sum_{t=1}^{T}\alpha_{t}^{2}-\sum_{t=1}^{T}\alpha_{t}\mathbb{E}_{t}\left[\left\Vert \nabla f\left(\theta_{t}\right)\right\Vert _{2}^{2}\right]+\left(C_{2}+C_{4}\right)\frac{\alpha_{1}}{\left(1-\beta_{1}\right)} 
\end{aligned}
\end{equation}

Combining the results of scaling on both sides of the inequality:
\begin{equation}
\begin{aligned} 
	&f\left(\theta^{*}\right)-f\left(\theta_{1}\right)\leq\left(C_{1}+C_{3}+C_{5}+C_{6}\right)\sum_{t=1}^{T}\alpha_{t}^{2}-\sum_{t=1}^{T}\alpha_{t}\mathbb{E}_{t}\left[\left\Vert \nabla f\left(\theta_{t}\right)\right\Vert _{2}^{2}\right] +\left(C_{2}+C_{4}\right)\frac{\alpha_{1}}{\left(1-\beta_{1}\right)}\\ 
	\Longrightarrow & \sum_{t=1}^{T}\alpha_{t}\mathbb{E}_{t}\left[\left\Vert \nabla f\left(\theta_{t}\right)\right\Vert _{2}^{2}\right]\leq\left(C_{1}+C_{3}+C_{5}+C_{6}\right)\sum_{t=1}^{T}\alpha_{t}^{2}+f\left(\theta_{1}\right)-f\left(\theta^{*}\right)+\left(C_{2}+C_{4}\right)\frac{\alpha_{1}}{\left(1-\beta_{1}\right)} 
\end{aligned}
\end{equation}

Due to $\mathbb{E}_{t}\left[\left\Vert \nabla f\left(\theta_{t}\right)\right\Vert _{2}^{2}\right]=\mathbb{E}_{t-1}\left[\left\Vert \nabla f\left(\theta_{t}\right)\right\Vert _{2}^{2}\right]$,
\begin{equation}
\begin{aligned}
	\sum_{t=1}^{T}\alpha_{t}\mathbb{E}_{t}\left[\left\Vert \nabla f\left(\theta_t\right)\right\Vert _{2}^{2}\right]= & \sum_{t=1}^{T}\alpha_{t}\mathbb{E}_{t-1}\left[\left\Vert \nabla f\left(\theta_t\right)\right\Vert _{2}^{2}\right]\\ 
	\geq & \sum_{t=1}^{T}\alpha_{t}\min_{t=1,2,\ldots,T}\mathbb{E}_{t-1}\left[\left\Vert \nabla f\left(\theta_t\right)\right\Vert _{2}^{2}\right]\\ 
	= & \min_{t=1,2,\ldots,T}\mathbb{E}_{t-1}\left[\left\Vert \nabla f\left(\theta_t\right)\right\Vert _{2}^{2}\right]\sum_{t=1}^{T}\alpha_{t}\\ 
	= & \cdot \mathbb{E}\left(T\right)\cdot\sum_{t=1}^{T}\alpha_{t} 
\end{aligned}
\end{equation}

Then let $C_{1}+C_{3}+C_{5}+C_{6}\triangleq C_{7}$,  $\underset{\geq0}{\underbrace{f\left(\theta_{1}\right)-f\left(\theta^{*}\right)}}+\left(C_{2}+C_{4}\right)\frac{\alpha_{1}}{\left(1-\beta_{1}\right)}\triangleq C_{8}$, therefore
\begin{equation}
\begin{aligned} 
	&  \mathbb{E}\left(T\right)\cdot\sum_{t=1}^{T}\alpha_{t}\leq C_{7}\sum_{t=1}^{T}\alpha_{t}^{2}+C_{8}\\ 
		\Longrightarrow & \mathbb{E}\left(T\right)\leq\frac{C_{7}\sum_{t=1}^{T}\alpha_{t}^{2}+C_{8}}{\sum_{t=1}^{T}\alpha_{t}} \end{aligned}
\end{equation}

Since $\alpha_{t}=\alpha/\sqrt{t}, \sum_{t=1}^{T} \frac{1}{t} \leq 1+\log T$, we have:
\begin{equation}
	\mathbb{E}(T) \leq \frac{C_{7}\alpha^2 (\log T + 1) + C_{8}}{2\alpha\sqrt{T}}
\end{equation}
    
\end{proof}

\newpage
\section{Detailed Experimental Supplement}
\label{sec:appendixe}
We performed extensive comparisons with other optimizers, including SGD \cite{1951a}, Adam\cite{kingma2014adam}, RAdam\cite{liu2019variance} and AdamW\cite{loshchilov2017decoupled}. The experiments include: (a) image classification on CIFAR dataset\cite{krizhevsky2009learning} with VGG \cite{2014Very}, ResNet \cite{he2016deep} and DenseNet \cite{huang2017densely}, and image recognition with ResNet on ImageNet \cite{Deng2009}.

\subsection{Image classification with CNNs on CIFAR}

\begin{figure}[htp]
    \centering
    \begin{subfigure}[t]{0.33\textwidth}
        \centering
        \includegraphics[width=\linewidth]{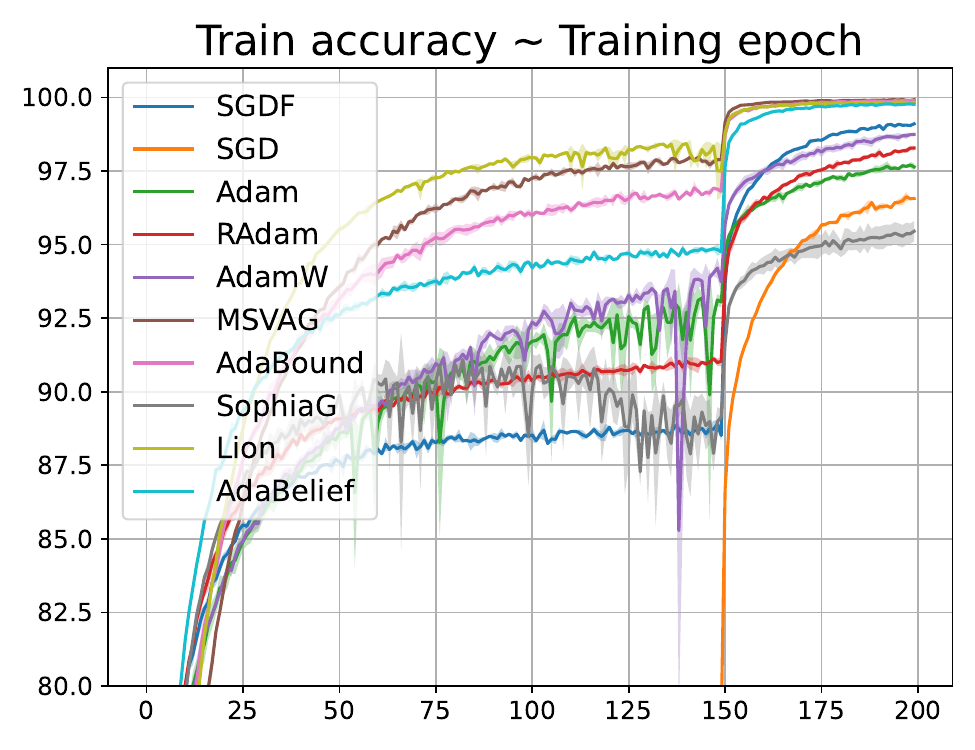}
        \caption{VGG11 on CIFAR-10 (Training)}
        \label{subfig:vgg_train_cifar10}
    \end{subfigure}
    \begin{subfigure}[t]{0.33\textwidth}
        \centering
        \includegraphics[width=\linewidth]{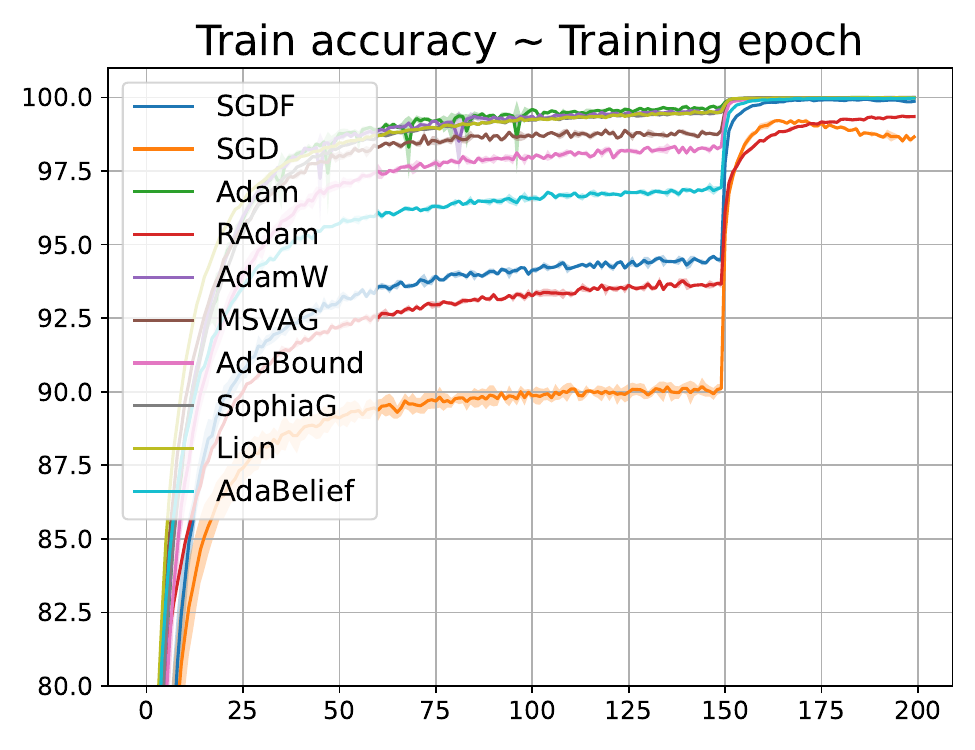}
        \caption{ResNet34 on CIFAR-10 (Training)}
        \label{subfig:resnet_train_cifar10}
    \end{subfigure}
    \begin{subfigure}[t]{0.33\textwidth}
        \centering
        \includegraphics[width=\linewidth]{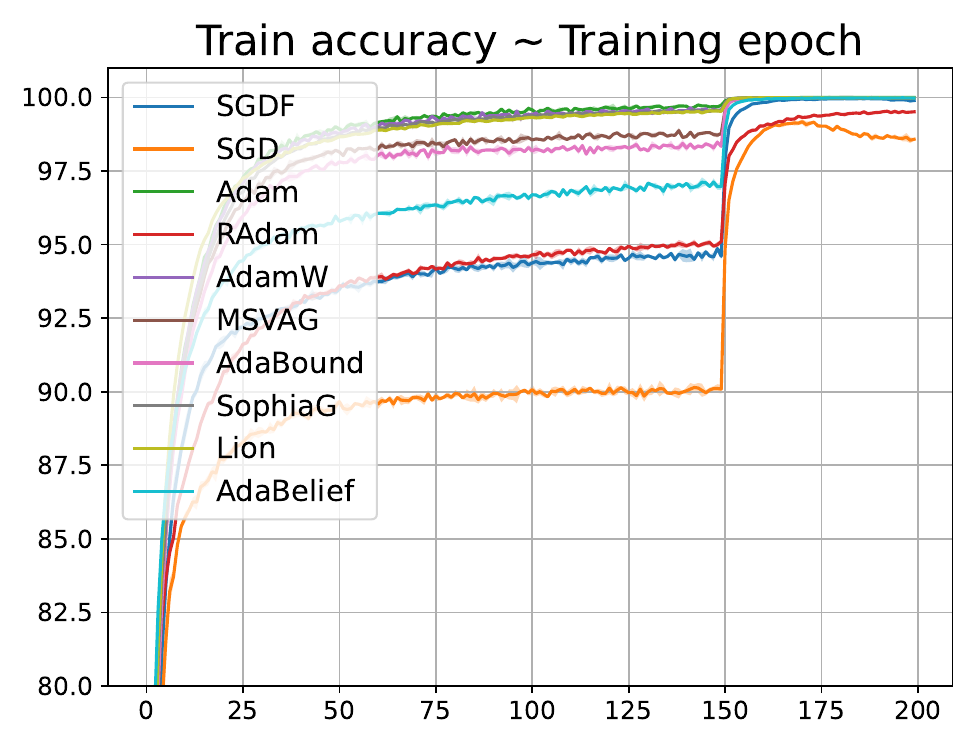}
        \caption{DenseNet121 on CIFAR-10 (Training)}
        \label{subfig:densenet_train_cifar10}
    \end{subfigure}
    \begin{subfigure}[t]{0.33\textwidth}
        \centering
        \includegraphics[width=\linewidth]{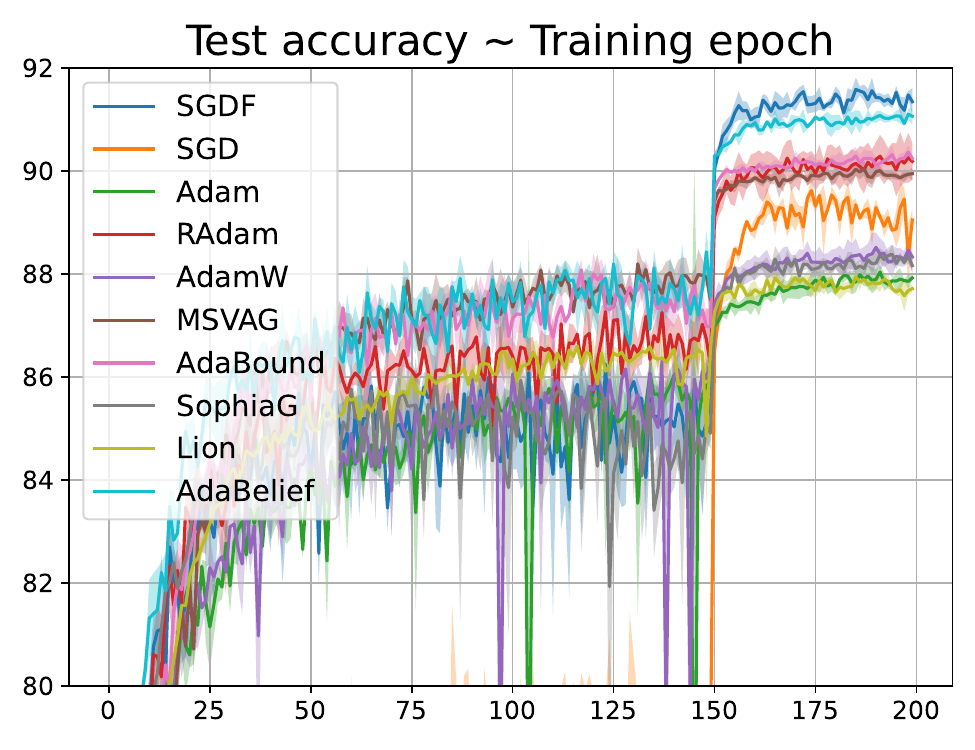}
        \caption{VGG11 on CIFAR-10 (Test)}
        \label{subfig:vgg_test_cifar10}
    \end{subfigure}
    \begin{subfigure}[t]{0.33\textwidth}
        \centering
        \includegraphics[width=\linewidth]{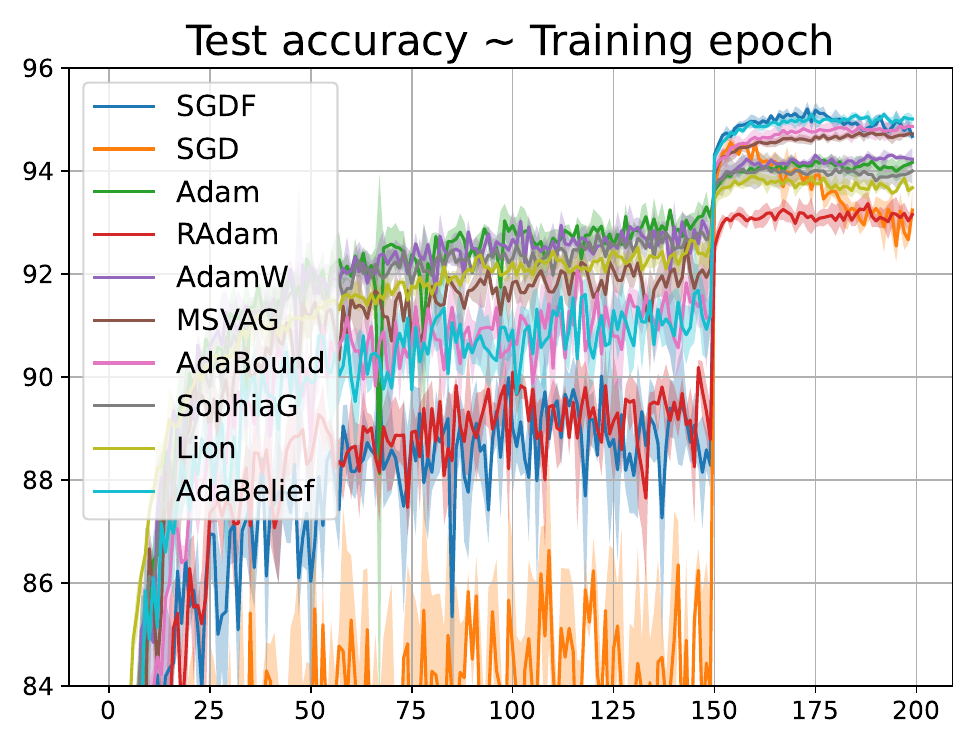}
        \caption{ResNet34 on CIFAR-10 (Test)}
        \label{subfig:resnet_test_cifar10}
    \end{subfigure}
    \begin{subfigure}[t]{0.33\textwidth}
        \centering
        \includegraphics[width=\linewidth]{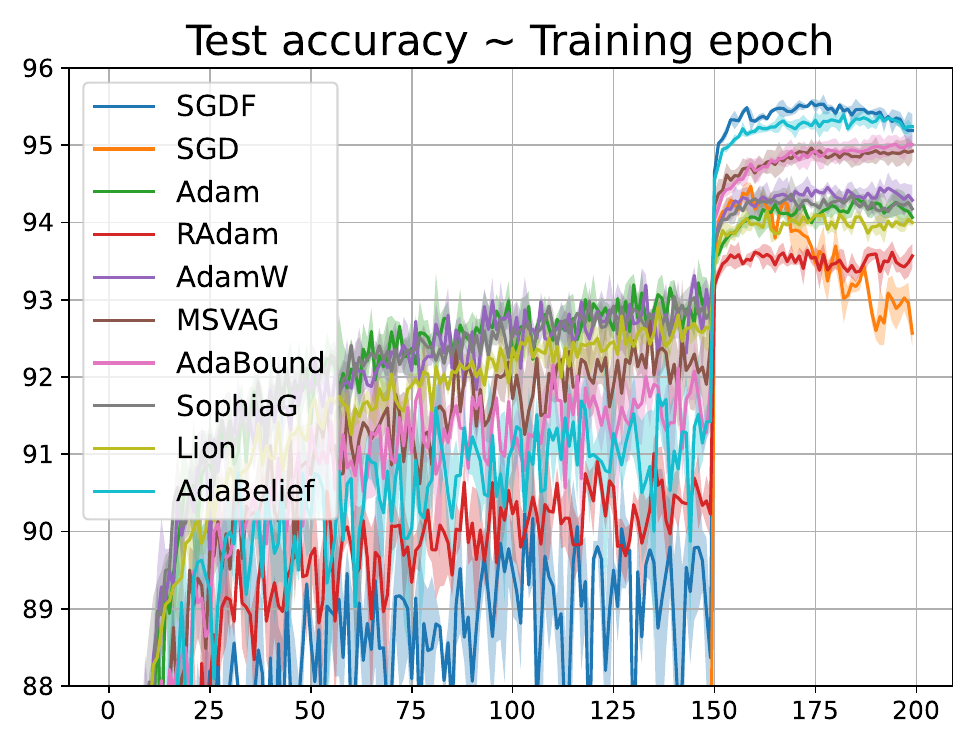}
        \caption{DenseNet121 on CIFAR-10 (Test)}
        \label{subfig:densenet_test_cifar10}
    \end{subfigure}
    \caption{Training (top row) and test (bottom row) accuracy of CNNs on CIFAR-10 dataset. We report confidence interval ([$\mu \pm \sigma$]) of 3 independent runs.}
    \label{fig:cifar10}
\end{figure}

\begin{figure}[htp]
    \centering
    \begin{subfigure}[t]{0.33\textwidth}
        \centering
        \includegraphics[width=\linewidth]{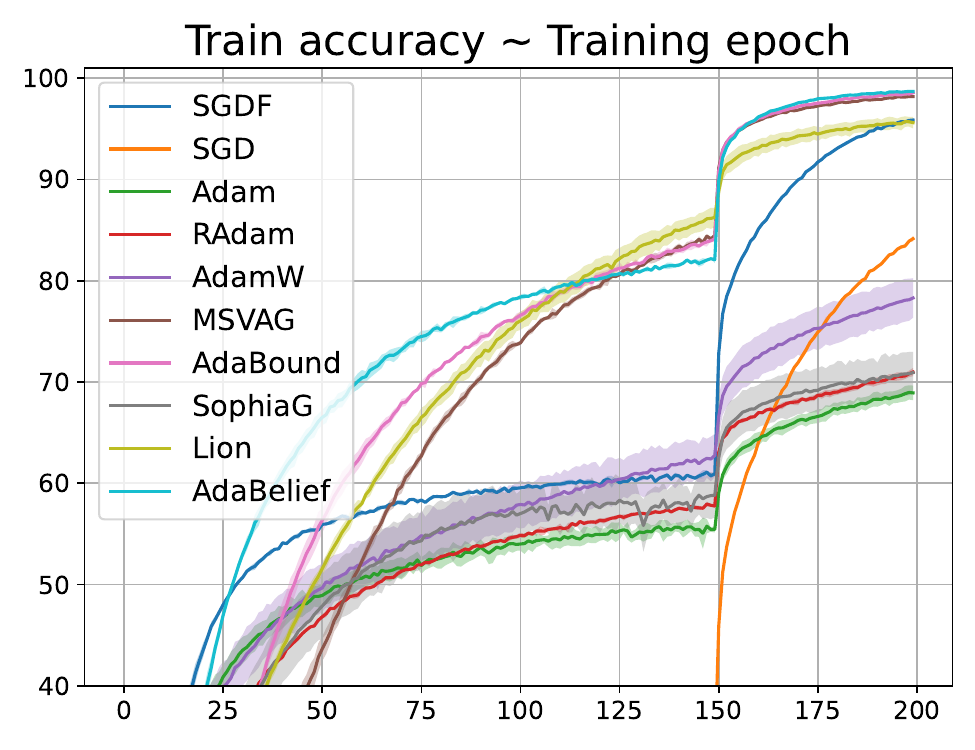}
        \caption{VGG11 on CIFAR-100 (Training)}
        \label{subfig:vgg_train_cifar100}
    \end{subfigure}
    \begin{subfigure}[t]{0.33\textwidth}
        \centering
        \includegraphics[width=\linewidth]{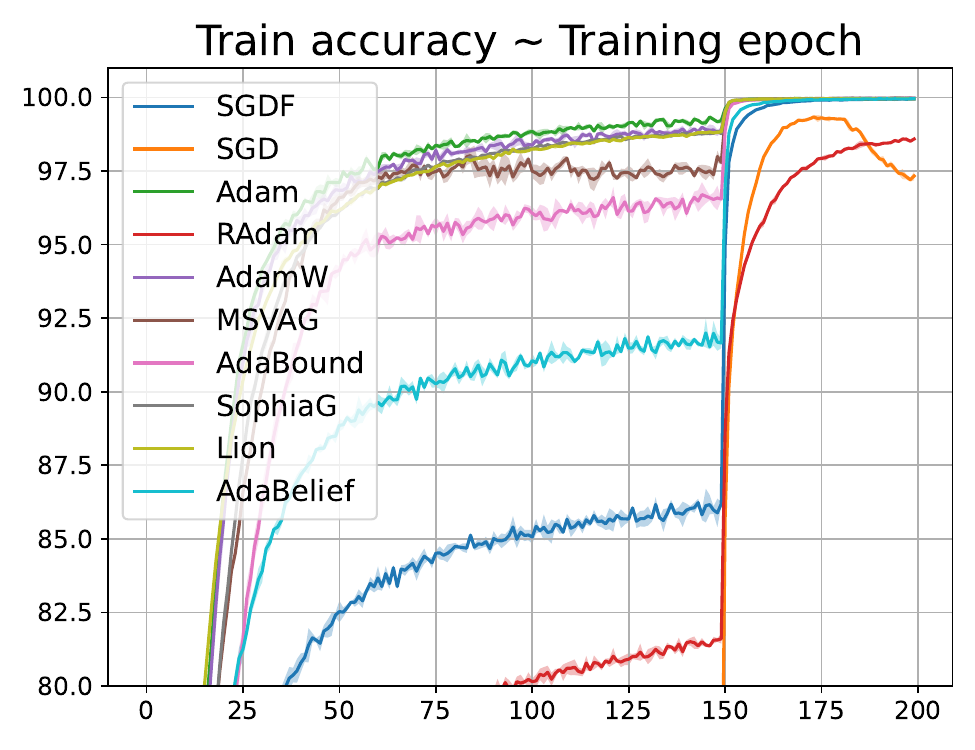}
        \caption{ResNet34 on CIFAR-100 (Training)}
        \label{subfig:resnet_train_cifar100}
    \end{subfigure}
    \begin{subfigure}[t]{0.33\textwidth}
        \centering
        \includegraphics[width=\linewidth]{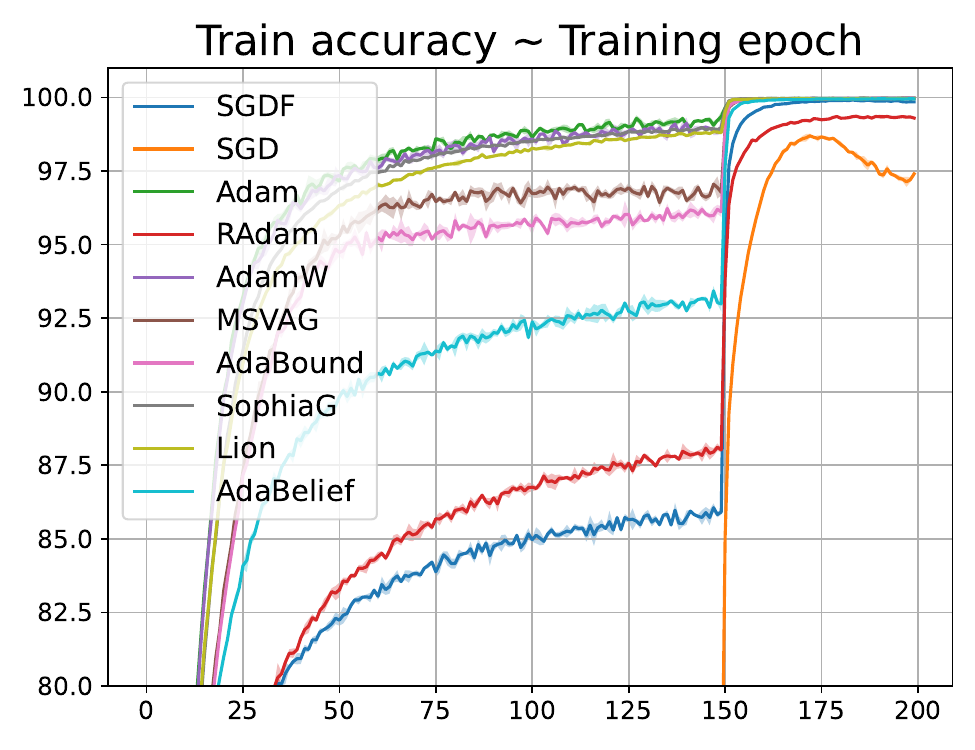}
        \caption{DenseNet121 on CIFAR-100 (Training)}
        \label{subfig:densenet_train_cifar100}
    \end{subfigure}
    \vspace{1em}
    \begin{subfigure}[t]{0.33\textwidth}
        \centering
        \includegraphics[width=\linewidth]{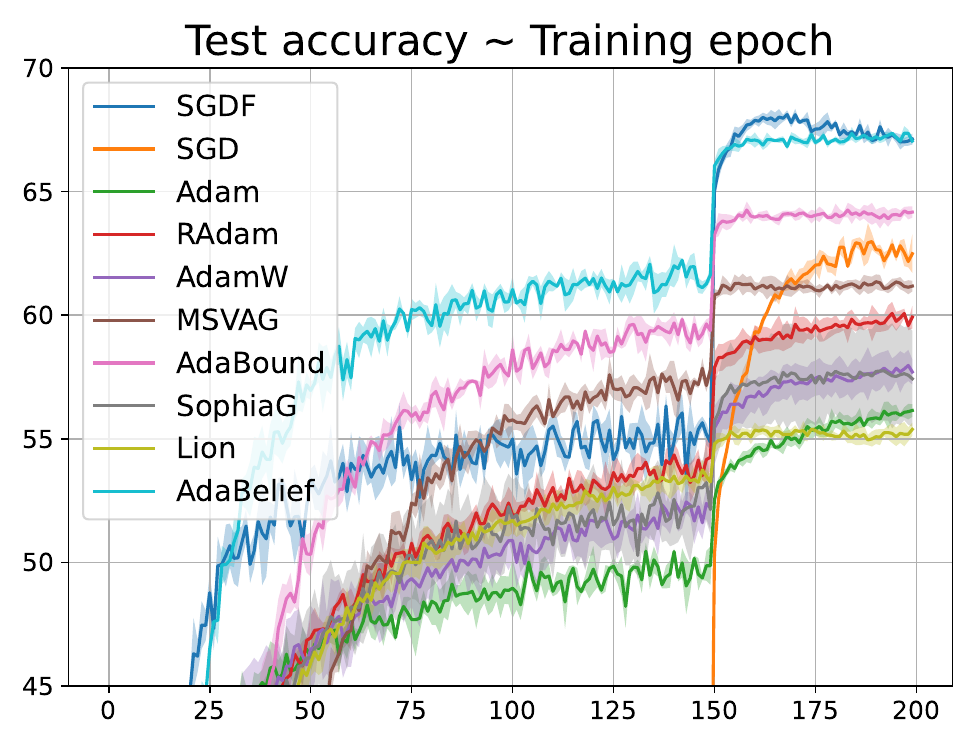}
        \caption{VGG11 on CIFAR-100 (Test)}
        \label{subfig:vgg_test_cifar100}
    \end{subfigure}
    \begin{subfigure}[t]{0.33\textwidth}
        \centering
        \includegraphics[width=\linewidth]{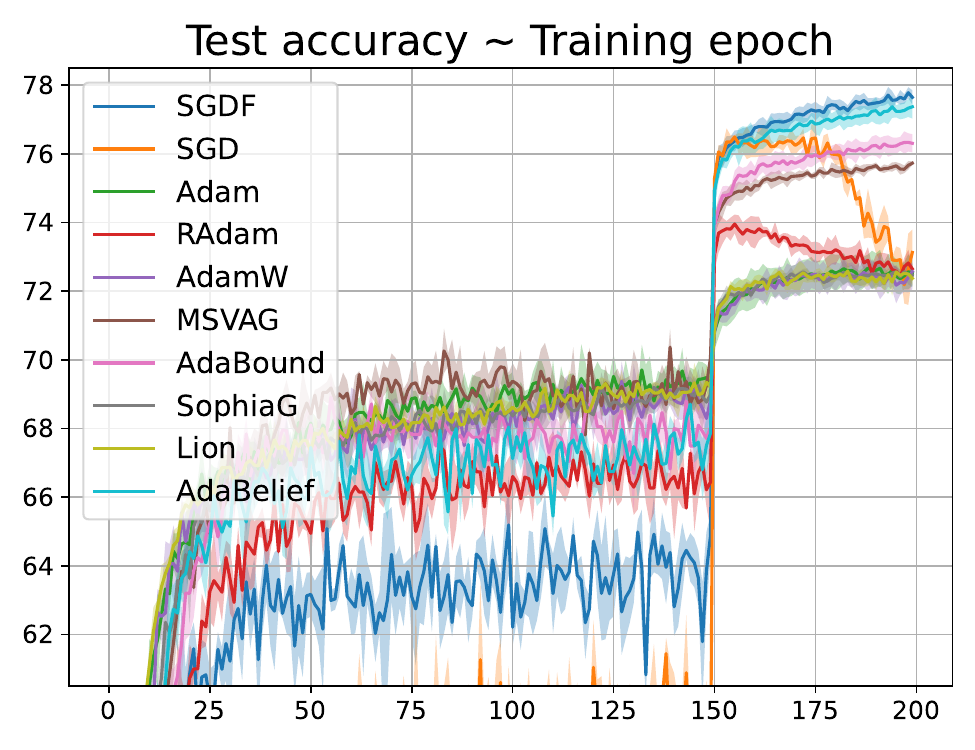}
        \caption{ResNet34 on CIFAR-100 (Test)}
        \label{subfig:resnet_test_cifar100}
    \end{subfigure}
    \begin{subfigure}[t]{0.33\textwidth}
        \centering
        \includegraphics[width=\linewidth]{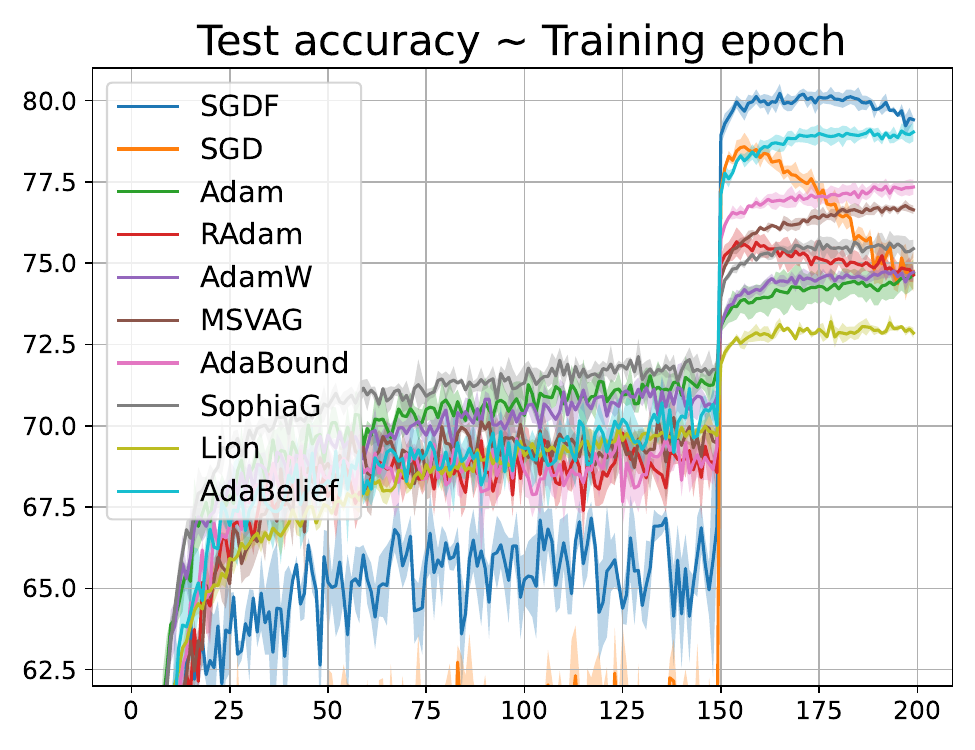}
        \caption{DenseNet121 on CIFAR-100 (Test)}
        \label{subfig:densenet_test_cifar100}
    \end{subfigure}

    \caption{Training (top row) and test (bottom row) accuracy of CNNs on CIFAR-100 dataset. We report confidence interval ([$\mu \pm \sigma$]) of 3 independent runs.}
    \label{fig:cifar100}
\end{figure}

For all experiments, the model is trained for 200 epochs with a batch size of 128, and the learning rate is multiplied by 0.1 at epoch 150. We performed extensive hyperparameter search as described in the main paper. Here, we report both training and test accuracy in Fig.~\ref{fig:cifar10} and Fig.~\ref{fig:cifar100}. Detailed experimental parameters we place in Tab.~\ref{tab:cifarhyperparameters}. We summarize the mean best test accuracies and their standard deviations for each algorithm in Tab.~\ref{tab:cifar_values}. The best results are highlighted in bold font. SGDF not only achieves the highest test accuracy but also a smaller gap between training and test accuracy compared with other optimizers. We ran each experiment three times with different seeds \{0, 1, 2\} to ensure the robustness of the results.

\begin{table}[h]
	\centering
	\caption{Hyperparameters used for CIFAR-10 and CIFAR-100 datasets.}
    \resizebox{1.0\linewidth}{!}{
	\begin{tabular}{@{}lcccccccc@{}}
		\toprule
		Optimizer & Learning Rate & $\beta_1$ & $\beta_2$ & Epochs & Schedule & Weight Decay & Batch Size & $\varepsilon$ \\
		\midrule
		SGDF    & 0.5 & 0.9   & 0.999  & 200 & StepLR      & 0.0005 & 128 & 1e-8 \\
		SGD     &  0.1  & 0.9 & -   & 200 & StepLR  & 0.0005 & 128 & -    \\
		Adam    &  0.001 & 0.9 & 0.999 & 200 & StepLR  & 0.0005 & 128 & 1e-8 \\
		RAdam   &  0.001 & 0.9 & 0.999 & 200 & StepLR  & 0.0005 & 128 & 1e-8 \\
		AdamW   &  0.001 & 0.9 & 0.999 & 200 & StepLR  & 0.01   & 128 & 1e-8 \\
		MSVAG   &  0.1 & 0.9 & 0.999 & 200 & StepLR  & 0.0005 & 128 & 1e-8 \\
		AdaBound&  0.001 & 0.9 & 0.999 & 200 & StepLR   & 0.0005 & 128 & - \\
		Sophia  & 0.0001 & 0.965 & 0.99 & 200 & StepLR  & 0.1 & 128 & - \\
		Lion    &  0.00002 & 0.9 & 0.99   & 200 & StepLR  & 0.1 & 128 & -    \\
		\bottomrule
	\end{tabular}
    }
	\label{tab:cifarhyperparameters}
\end{table}

\begin{table}[ht]
\centering
\caption{Test Accuracies for CIFAR-10 and CIFAR-100 across different models and algorithms.}
\resizebox{1.0\linewidth}{!}{
\begin{tabular}{l|ccc|ccc}
\toprule
\multirow{2.5}{*}{Algorithm} & \multicolumn{3}{c}{CIFAR-10} & \multicolumn{3}{c}{CIFAR-100} \\
\cmidrule{2-7}
& VGG11    & ResNet34   & DenseNet121  & VGG11  & ResNet34  & DenseNet121     \\
\midrule
SGDF & $\textbf{91.76}_{\pm 0.11}$ & $\textbf{95.29}_{\pm 0.09}$ & $\textbf{95.63}_{\pm 0.04}$ & $\textbf{68.29}_{\pm 0.15}$ & $\textbf{77.80}_{\pm 0.18}$ & $\textbf{80.33}_{\pm 0.24}$ \\
SGD  & $89.83_{\pm 0.05}$ & $94.62_{\pm 0.07}$ & $94.52_{\pm 0.03}$ & $63.48_{\pm 0.39}$ & $76.88_{\pm 0.12}$ & $78.77_{\pm 0.27}$ \\
Adam & $88.12_{\pm 0.10}$ & $94.30_{\pm 0.06}$ & $94.37_{\pm 0.17}$ & $56.27_{\pm 0.32}$ & $72.81_{\pm 0.45}$ & $74.67_{\pm 0.45}$ \\
AdamW & $88.59_{\pm 0.20}$ & $94.42_{\pm 0.00}$ & $94.61_{\pm 0.06}$ & $58.09_{\pm 0.69}$ & $72.74_{\pm 0.45}$ & $74.96_{\pm 0.10}$ \\
RAdam & $90.47_{\pm 0.34}$ & $93.41_{\pm 0.21}$ & $93.75_{\pm 0.04}$ & $60.20_{\pm 0.37}$ & $74.08_{\pm 0.35}$ & $75.82_{\pm 0.28}$ \\
MSVAG & $90.08_{\pm 0.13}$ & $94.79_{\pm 0.08}$ & $95.01_{\pm 0.12}$ & $61.55_{\pm 0.23}$ & $75.75_{\pm 0.06}$ & $76.84_{\pm 0.13}$ \\
Lion & $88.04_{\pm 0.06}$ & $93.97_{\pm 0.10}$ & $94.26_{\pm 0.02}$ & $55.59_{\pm 0.15}$ & $72.79_{\pm 0.14}$ & $73.41_{\pm 0.10}$ \\
SophiaG & $88.53_{\pm 0.04}$ & $94.15_{\pm 0.26}$ & $94.53_{\pm 0.13}$ & $58.01_{\pm 1.85}$ & $72.83_{\pm 0.18}$ & $75.81_{\pm 0.23}$ \\
AdaBound & $90.41_{\pm 0.12}$ & $94.93_{\pm 0.12}$ & $95.06_{\pm 0.13}$ & $64.51_{\pm 0.15}$ & $76.37_{\pm 0.29}$ & $77.43_{\pm 0.18}$ \\
AdaBelief & $91.24_{\pm 0.04}$ & $95.18_{\pm 0.01}$ & $95.44_{\pm 0.04}$ & $67.59_{\pm 0.03}$ & $77.47_{\pm 0.34}$ & $79.20_{\pm 0.16}$ \\
\bottomrule
\end{tabular}
}
\label{tab:cifar_values}
\end{table}

\vspace{5em}

\subsection{Image Classification on ImageNet}

\begin{figure}[htp]
    \centering
    \begin{subfigure}[t]{0.45\textwidth}
        \centering
        \includegraphics[width=\linewidth]{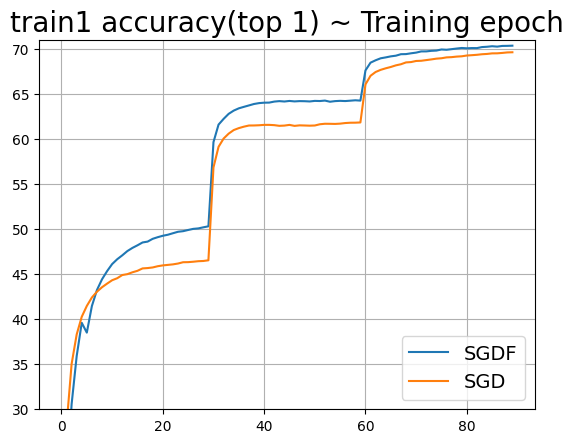}
        \caption{Training accuracy (top-1)}
        \label{subfig:imagenet_train}
    \end{subfigure}
    \begin{subfigure}[t]{0.45\textwidth}
        \centering
        \includegraphics[width=\linewidth]{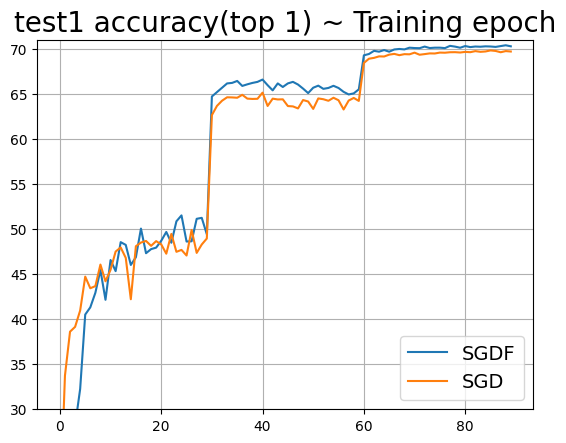}
        \caption{Test accuracy (top-1)}
        \label{subfig:imagenet_test}
    \end{subfigure}
    
    \caption{Training and test accuracy (top-1) of ResNet18 on ImageNet.}
    \label{fig:imagenet_curve}
\end{figure}

We experimented with a VGG / ResNet / DenseNet on ImageNet classification task. For SGDF and SGD, we set the initial learning rate of 0.5 same as CIFAR experiments. The weight decay is set as \(10^{-4}\)  for both cases to match the settings in \cite{liu2019variance}. Here $\beta_1$ serves to capture the gradient mean. The more closer $\beta_1$ is to 1, the longer the moving window is and the wider the historical mean is captured. Since ImageNet dataset has more iterations than CIFAR dataset, we directly set $\beta_1$ = 0.5 to prevent $K_t$ from being overly influenced by the historical mean gradient. For sure, setting $beta_1$ to 0.9, consistent with CIFAR experiments can also be superior to SGD, and adjusting $\beta_1$ to 0.5 or 0.9 according to the size of the dataset and batch size can bring better results. Detailed experimental parameters we place in Tab.~\ref{tab:imagenethyperparameters}. As shown in Fig.~\ref{fig:imagenet_curve}, SGDF outperformed SGD. 

\begin{table}[htp]
	\centering
	\caption{Hyperparameters used for ImageNet.}
	\begin{tabular}{@{}lcccccccc@{}}
		\toprule
		Optimizer & Learning Rate & $\beta_1$ & $\beta_2$ & Epochs & Schedule & Weight Decay & Batch Size & $\varepsilon$ \\
		\midrule
		 SGDF    & 0.5 & 0.5   & 0.999  & 100/90 & Cosine      & 0.0001 & 256 & 1e-8 \\
		 SGD     & 0.1  & 0.9 & -     & 100/90 & Cosine      & 0.0001 & 256 & - \\
		\bottomrule
	\end{tabular}
	\label{tab:imagenethyperparameters}
\end{table}
\vspace{-3mm}

\vspace{-3mm}

\subsection{Objective Detection on PASCAL VOC}

\begin{figure*}[ht]
    \centering
    \begin{subfigure}[t]{0.19\textwidth}
        \centering
        \includegraphics[width=\linewidth]{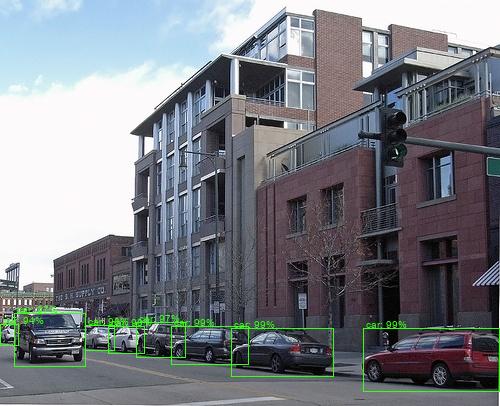}
        \caption{SGDF}
        \label{subfig:sgdf_test1}
    \end{subfigure}
    \begin{subfigure}[t]{0.19\textwidth}
        \centering
        \includegraphics[width=\linewidth]{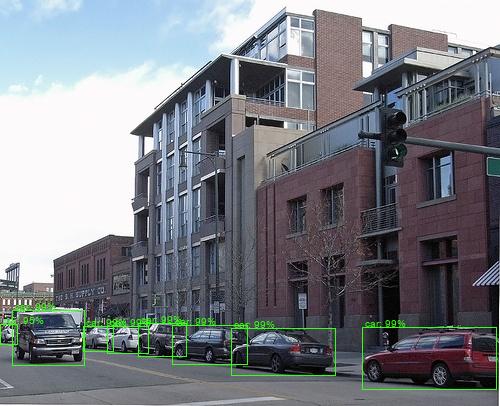}
        \caption{SGDM}
        \label{subfig:sgdm_test1}
    \end{subfigure}
    \begin{subfigure}[t]{0.19\textwidth}
        \centering
        \includegraphics[width=\linewidth]{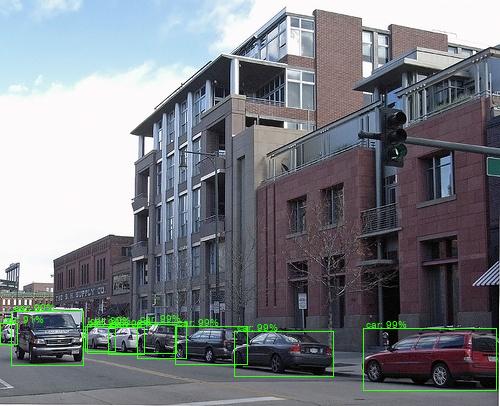}
        \caption{Adam}
        \label{subfig:adam_test1}
    \end{subfigure}
    \begin{subfigure}[t]{0.19\textwidth}
        \centering
        \includegraphics[width=\linewidth]{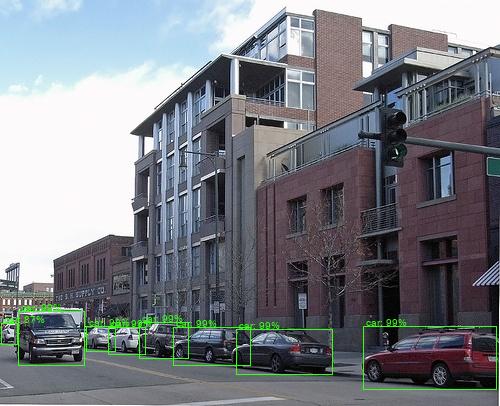}
        \caption{AdamW}
        \label{subfig:adamw_test1}
    \end{subfigure}
    \begin{subfigure}[t]{0.19\textwidth}
        \centering
        \includegraphics[width=\linewidth]{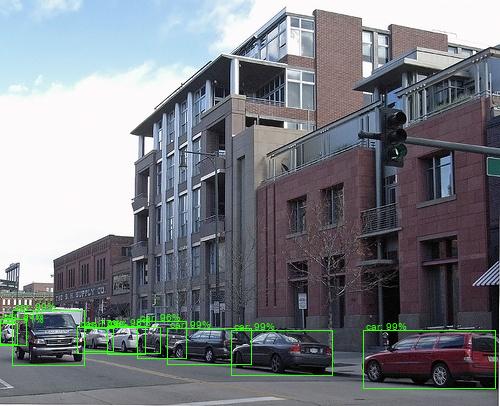}
        \caption{RAdam}
        \label{subfig:radam_test1}
    \end{subfigure}
    \begin{subfigure}[t]{0.19\textwidth}
        \centering
        \includegraphics[width=\linewidth]{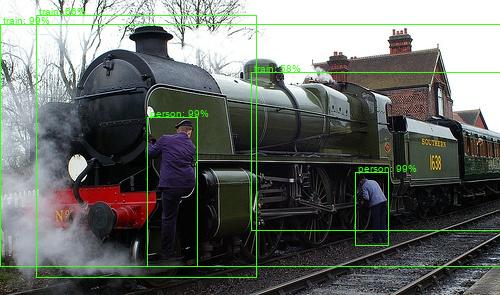}
        \caption{SGDF}
        \label{subfig:sgdf_test2}
    \end{subfigure}
    \begin{subfigure}[t]{0.19\textwidth}
        \centering
        \includegraphics[width=\linewidth]{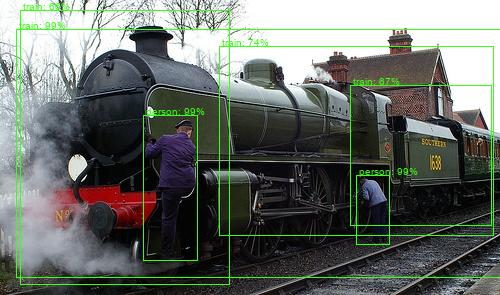}
        \caption{SGDM}
        \label{subfig:sgdm_test2}
    \end{subfigure}
    \begin{subfigure}[t]{0.19\textwidth}
        \centering
        \includegraphics[width=\linewidth]{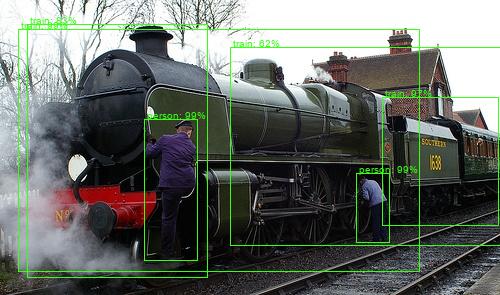}
        \caption{Adam}
        \label{subfig:adam_test2}
    \end{subfigure}
    \begin{subfigure}[t]{0.19\textwidth}
        \centering
        \includegraphics[width=\linewidth]{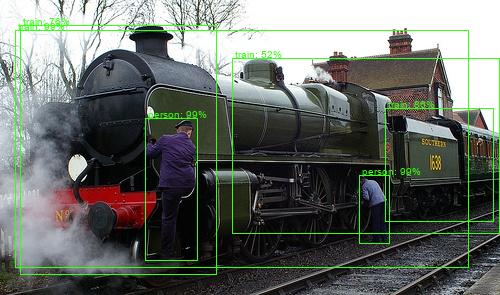}
        \caption{AdamW}
        \label{subfig:adamw_test2}
    \end{subfigure}
    \begin{subfigure}[t]{0.19\textwidth}
        \centering
        \includegraphics[width=\linewidth]{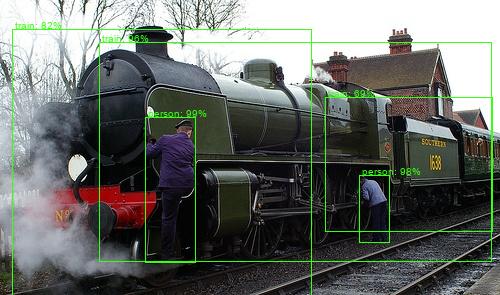}
        \caption{RAdam}
        \label{subfig:radam_test2}
    \end{subfigure}

    \caption{Detection examples using Faster-RCNN + FPN trained on PASCAL VOC.}
    \label{fig:detection_examples}
\end{figure*}

We conducted object detection experiments on the PASCAL VOC dataset~\citep{pascal-voc-2010}. The model used in these experiments was pre-trained on the COCO dataset~\citep{coco-dataset}, obtained from the official website. We trained this model on the VOC2007 and VOC2012 trainval dataset (17K) and evaluated it on the VOC2007 test dataset (5K). The utilized model was Faster-RCNN~\cite{faster_rcnn} with FPN, and the backbone was ResNet50~\citep{he2016deep}. We train 4 epochs and adjust the learning rate decay by a factor of 0.1 at the last epoch. We show the results on PASCAL VOC\cite{pascal-voc-2010}. Object detection with a Faster-RCNN model\cite{faster_rcnn}. Detailed experimental parameters we place in Fig.~\ref{tab:hyperparameters_object_detection}. The results are reported in Tab.~\ref{tab:object_detection}, and detection examples are shown in Fig.~\ref{fig:detection_examples}. These results also illustrate that our method is still efficient in object detection tasks.

\begin{table*}[htp]
	\centering
	\caption{Hyperparameters for object detection on PASCAL VOC using Faster-RCNN+FPN with different optimizers.}
    \resizebox{0.9\linewidth}{!}{
	\begin{tabular}{@{}lcccccccc@{}}
		\toprule
		Optimizer & Learning Rate & $\beta_1$& $\beta_2$ & Epochs & Schedule & Weight Decay &Batch Size & $\varepsilon$ \\
		\midrule
		SGDF       & 0.01 & 0.9 & 0.999 & 4 & StepLR & 0.0001 &2& 1e-8 \\
		SGD        & 0.01 & 0.9 & -     & 4 & StepLR & 0.0001 &2& - \\
		Adam       & 0.0001 & 0.9 & 0.999 & 4 & StepLR & 0.0001 &2& 1e-8 \\
		AdamW      & 0.0001 & 0.9 & 0.999 & 4 & StepLR & 0.0001 &2& 1e-8 \\
		RAdam      & 0.0001 & 0.9 & 0.999 & 4 & StepLR & 0.0001 &2& 1e-8 \\
		\bottomrule
	\end{tabular}
    }
	\label{tab:hyperparameters_object_detection}
\end{table*}

\vspace{-5mm}

\subsection{Image Generation.}
The stability of optimizers is crucial, especially when training Generative Adversarial Networks (GANs). If the generator and discriminator have mismatched complexities, it can lead to imbalance during GAN training, causing the GAN to fail to converge. This is known as model collapse. For instance, Vanilla SGD frequently causes model collapse, making adaptive optimizers like Adam and RMSProp the preferred choice. Therefore, GAN training provides a good benchmark for assessing optimizer stability. For reproducibility details, please refer to the parameter table in Tab.~\ref{tab:image_generation_hyperparameters}.

We evaluated the Wasserstein-GAN with gradient penalty (WGAN-GP)~\cite{salimans2016improved}. Using well-known optimizers~\cite{bernstein2020distance,zaheer2018adaptive}, the model was trained for 100 epochs. We then calculated the Frechet Inception Distance (FID)~\cite{2017GANs} which is a metric that measures the similarity between the real image and the generated image distribution and is used to assess the quality of the generated model (lower FID indicates superior performance). Five random runs were conducted, and the outcomes are presented in Tab.\ref{tab:fid-wgan-gp}. Results for SGD and MSVAG were extracted from Zhuang~\etal~ 
\cite{zhuang2020adabelief}.

%
\begin{table*}[htbp]
\centering
\caption{FID score of WGAN-GP.}
\resizebox{1.0\textwidth}{!}{%
\begin{tabular}{c|ccccccccc}
\toprule
Method & SGDF & Adam & RMSProp & RAdam & Fromage & Yogi & AdaBound & SGD & MSVAG \\
\midrule
FID & $89.3 \pm 4.9$ & $\textbf{78.6} \pm 4.8$ & $109.2 \pm 14.5$ & $93.4 \pm 8.3$ & $101.5 \pm 28.9$ & $138.7 \pm 21.2$ & $119.8 \pm 24.6$ & $250.3 \pm 30.1$ & $239.7 \pm 5.2$ \\
\bottomrule
\end{tabular}%
}
\label{tab:fid-wgan-gp}
\end{table*}

Experimental results demonstrate that SGDF significantly enhances WGAN-GP model training, achieving a FID score higher than vanilla SGD and outperforming most adaptive optimization methods. The integration of a Wiener filter in SGDF facilitates smooth gradient updates, mitigating training oscillations and effectively addressing the issue of pattern collapse.

\begin{table}[h]
	\centering
	\caption{Hyperparameters for Image Generation Tasks.}
    \resizebox{0.8\linewidth}{!}{
	\begin{tabular}{@{}lcccccc@{}}
		\toprule
		Optimizer & Learning Rate & $\beta_1$ & $\beta_2$ & Epochs & Batch Size & $\varepsilon$ \\
		\midrule
		SGDF     & 0.01 & 0.5 & 0.999 & 100 & 64 & 1e-8 \\
		Adam      & 0.0002 & 0.5 & 0.999 & 100 & 64 & 1e-8 \\
		AdamW     & 0.0002 & 0.5 & 0.999 & 100 & 64 & 1e-8 \\
		Fromage   & 0.01   & 0.5 & 0.999 & 100 & 64 & 1e-8 \\
		RMSProp   & 0.0002 & 0.5 & 0.999 & 100 & 64 & 1e-8 \\
		AdaBound  & 0.0002 & 0.5 & 0.999 & 100 & 64 & 1e-8 \\
		Yogi      & 0.01   & 0.5 & 0.999 & 100 & 64 & 1e-8 \\
		RAdam     & 0.0002 & 0.5 & 0.999 & 100 & 64 & 1e-8 \\
		\bottomrule
	\end{tabular}
    }
	\label{tab:image_generation_hyperparameters}
\end{table}

\subsection{Fine-tuning in ViT}
To evaluate SGDF's performance, we used Vision Transformers (ViT)~\cite{dosovitskiy2020image} on six benchmark datasets: CIFAR-10, CIFAR-100, Oxford-IIIT-Pets~\cite{parkhi2012cats}, Oxford Flowers-102~\cite{nilsback2008automated}, Food101~\cite{bossard2014food}, and ImageNet-1K. Two ViT variants, ViT-B/32 and ViT-L/32, pretrained on ImageNet-21K, were selected. For fine-tuning, we replaced the original MLP classification head with a new fully connected layer, tailored to the dataset categories. All Transformer backbone weights were retained, preserving the rich representations learned from ImageNet-21K. We increased the image resolution (\eg, from $224 \times 224$ to $384 \times 384$) to improve accuracy, while adjusting positional encoding through 2D interpolation to match the new resolution. For optimization, SGDF was compared to SGD with momentum as a baseline (We research learning set \{ 0.001, 0.003, 0.01, 0.03\} same as~\cite{dosovitskiy2020image}. For ours method, we're not tuning and just mirror the hyperparameter in the CIFAR experiments.), using cosine learning rate decay and no weight decay. A batch size of 512 and global gradient clipping (norm of 1) were used to prevent gradient explosion. All experiments were trained uniformly for 10 epochs and the random seed is set to 2025. We set the random seed to 2025. Results are summarized in Table~\ref{tab:vitresult}. We summarized the hyperparameter in Tab.~\ref{tab:vithyperparameters}. 
\begin{table}[htp]
	\centering
	\caption{Hyperparameters used for fine-tuning ViT.}
    \resizebox{1.0\linewidth}{!}{
	\begin{tabular}{@{}lccccccccc@{}}
		\toprule
		Optimizer & Learning Rate & $\beta_1$ & $\beta_2$ & Epochs & Schedule & Weight Decay & Batch Size & $\varepsilon$ & Resolution \\
		\midrule
		 SGDF    & 0.5 & 0.9   & 0.999  & 10 & Cosine      & 0 & 512 & 1e-8 & 384\\
		 SGD     & 0.03  & 0.9 & -     & 10 & Cosine      & 0 & 512 & - & 384\\
		\bottomrule
	\end{tabular}
    }
	\label{tab:vithyperparameters}
\end{table}

\subsection{Top Eigenvalues of Hessian and Hessian Trace}
\begin{figure*}[htbp] 
    \centering
    \begin{subfigure}[t]{0.23\textwidth}
        \centering
        \begin{overpic}[width=\textwidth]{experiments/hessian_spectra/histagram_hessian/sgdf.pdf}
            \put(60,67){\scriptsize Trace: 192.47 }
            \put(64,60){\scriptsize $\lambda_{\text{max}}$: 13.32}
        \end{overpic}
        \caption{SGDF}
        \label{subfig:hessian_sgdf}
    \end{subfigure}%
    \begin{subfigure}[t]{0.23\textwidth}
        \centering
        \begin{overpic}[width=\textwidth]{experiments/hessian_spectra/histagram_hessian/sgd.pdf}
            \put(62,67){\scriptsize Trace: 419.30}
            \put(67,60){\scriptsize $\lambda_{\text{max}}$: 22.51}
        \end{overpic}
        \caption{SGD}
        \label{subfig:hessian_sgd}
    \end{subfigure}%
    \begin{subfigure}[t]{0.23\textwidth}
        \centering
        \begin{overpic}[width=\textwidth]{experiments/hessian_spectra/histagram_hessian/ema.pdf}
            \put(63,67){\scriptsize Trace: 284.38}
            \put(68,60){\scriptsize $\lambda_{\text{max}}$: 24.11}
        \end{overpic}
        \caption{SGD-EMA}
        \label{subfig:hessian_ema}
    \end{subfigure}%
    \begin{subfigure}[t]{0.23\textwidth}
        \centering
        \begin{overpic}[width=\textwidth]{experiments/hessian_spectra/histagram_hessian/sgdm.pdf}
            \put(63,67){\scriptsize Trace: 491.63}
            \put(68,60){\scriptsize $\lambda_{\text{max}}$: 32.61}
        \end{overpic}
        \caption{SGD-CM}
        \label{subfig:hessian_sgdm}
    \end{subfigure}
    
    \begin{subfigure}[t]{0.23\textwidth}
        \centering
        \begin{overpic}[width=\textwidth]{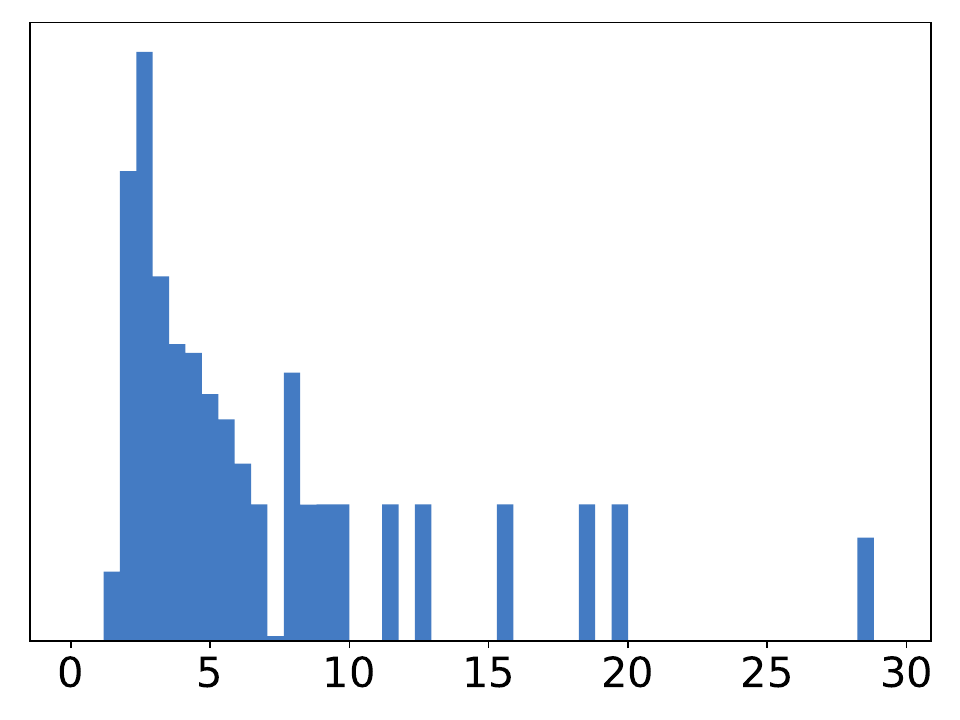}
            \put(62,67){\scriptsize Trace: 427.72 }
            \put(67,60){\scriptsize $\lambda_{\text{max}}$: 28.38}
        \end{overpic}
        \caption{AdaBelief}
        \label{subfig:hessian_adabelief}
    \end{subfigure}%
    \begin{subfigure}[t]{0.23\textwidth}
        \centering
        \begin{overpic}[width=\textwidth]{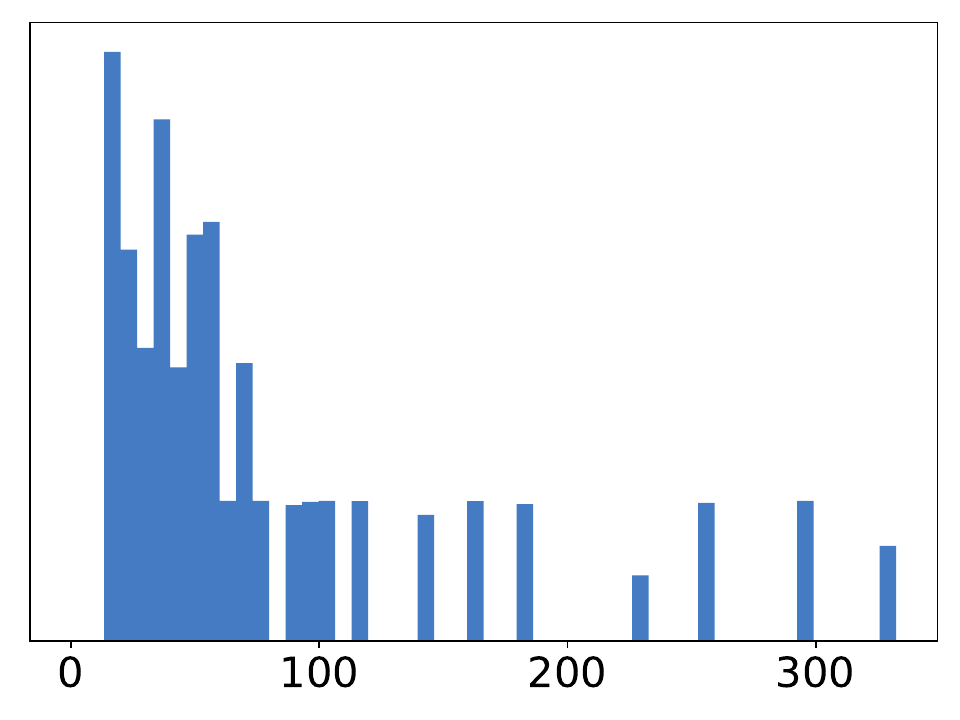}
            \put(59,67){\scriptsize Trace: 3970.98}
            \put(64,60){\scriptsize $\lambda_{\text{max}}$: 330.04}
        \end{overpic}
        \caption{Adam}
        \label{subfig:hessian_adam}
    \end{subfigure}%
    \begin{subfigure}[t]{0.23\textwidth}
        \centering
        \begin{overpic}[width=\textwidth]{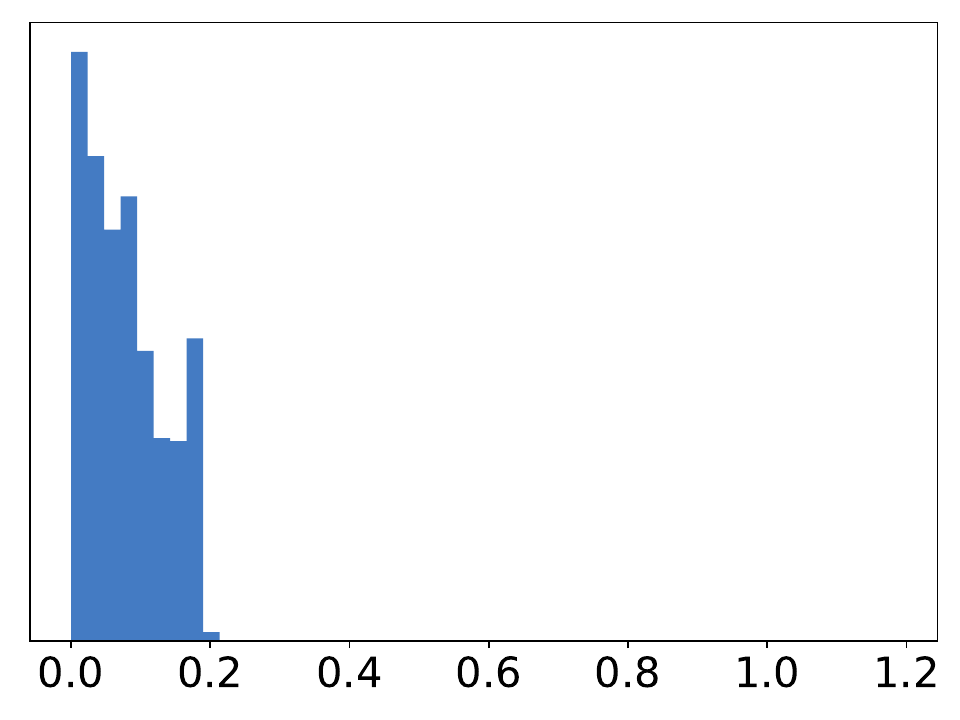}
            \put(69,67){\scriptsize Trace: 0.39}
            \put(71,60){\scriptsize $\lambda_{\text{max}}$: 0.11}
        \end{overpic}
        \caption{AdamW}
        \label{subfig:hessian_adamw}
    \end{subfigure}%
    \begin{subfigure}[t]{0.23\textwidth}
        \centering
        \begin{overpic}[width=\textwidth]{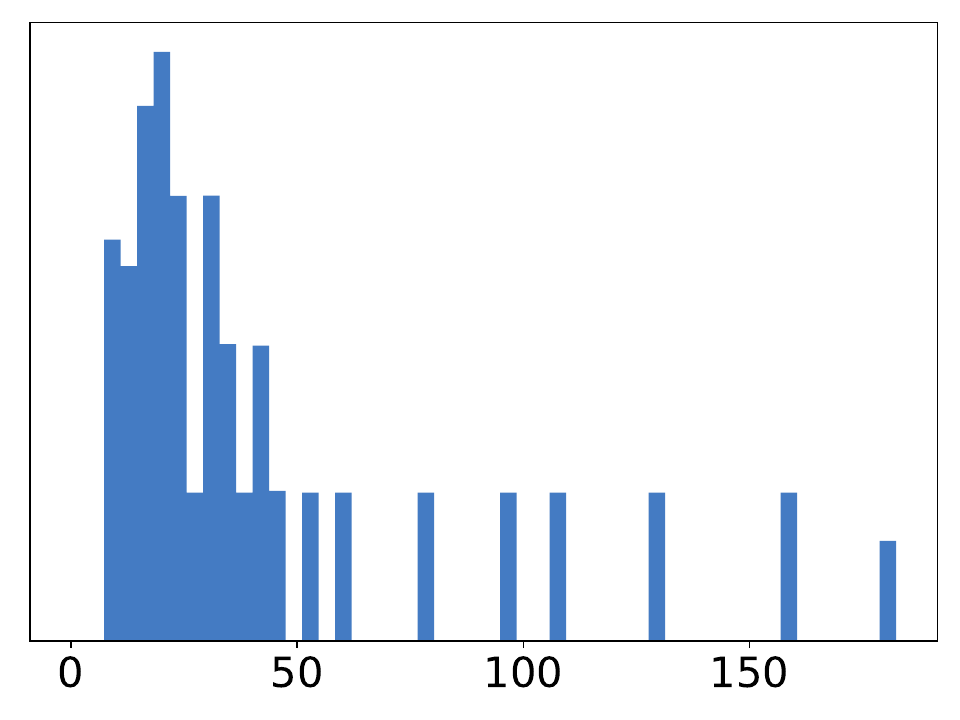}
            \put(59,67){\scriptsize Trace: 1927.24}
            \put(63,60){\scriptsize $\lambda_{\text{max}}$: 180.98}
        \end{overpic}
        \caption{RAdam}
        \label{subfig:hessian_radam}
    \end{subfigure}
    
    \caption{Histogram of Top 50 Hessian Eigenvalues.}
    \label{fig:hessian_spectrum}
\end{figure*}

We computed the Hessian spectrum of ResNet-18 trained on the CIFAR-100 dataset for 200 epochs using more optimization methods: SGDF, SGD, SGD-EMA, SGD-CM, Adabelief, Adam, AdamW, and RAdam. We employed power iteration~\citep{2018Hessian} to compute the top eigenvalues of Hessian and Hutchinson’s method~\citep{2020PyHessian} to compute the Hessian trace. Histograms illustrating the distribution of the top 50 Hessian eigenvalues for each optimization method are presented in Fig.~\ref{fig:hessian_spectrum}. SGDF brings lower eigenvalue and trace of the hessian matrix, which explains the fact that SGDF demonstrates better performance than SGD as the categorization category increases. Note that~\ref{subfig:hessian_adamw} shows that AdamW achieves very low hessian matrix eigenvalues and traces, but the final test set accuracy is about 4\% lower than the other methods, and that AdamW's unique decouple weight decay changes the nature of the converged solution (We apply decoupled weight decay to other algorithms and similar results occur). 

\subsection{Visualization of Landscapes}
\begin{figure}[htbp]
    \centering
    \begin{subfigure}[t]{0.24\linewidth}
        \centering
        \includegraphics[width=\linewidth]{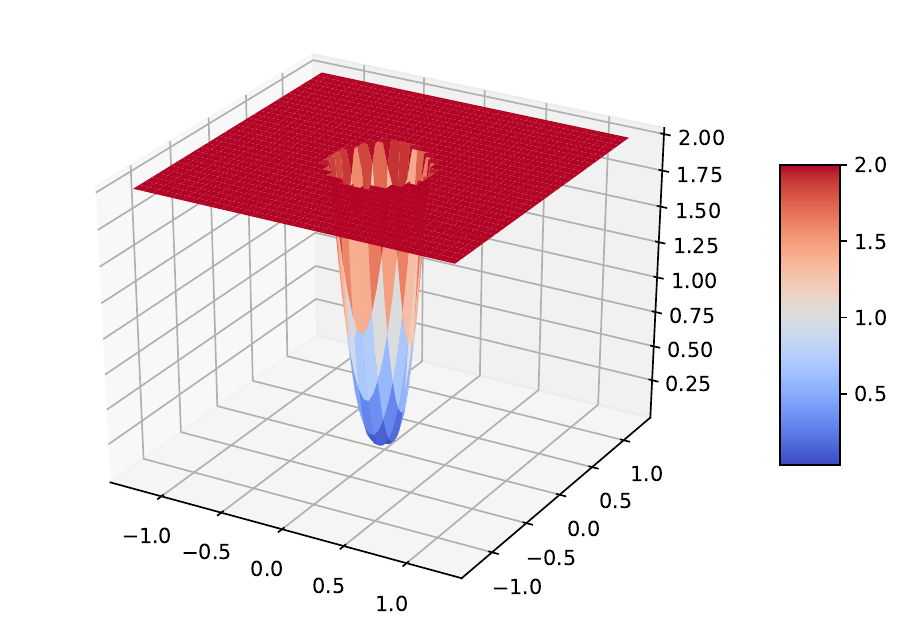}
        \caption{Adam}
        \label{subfig:adam}
    \end{subfigure}
    \hfill
    \begin{subfigure}[t]{0.24\linewidth}
        \centering
        \includegraphics[width=\linewidth]{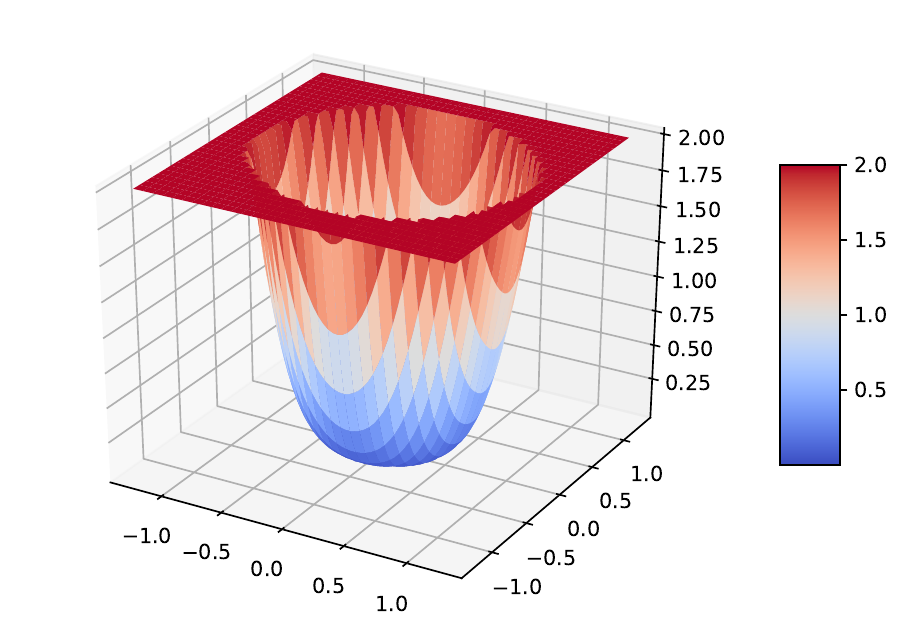}
        \caption{SGD}
        \label{subfig:sgd}
    \end{subfigure}
    \hfill
    \begin{subfigure}[t]{0.24\linewidth}
        \centering
        \includegraphics[width=\linewidth]{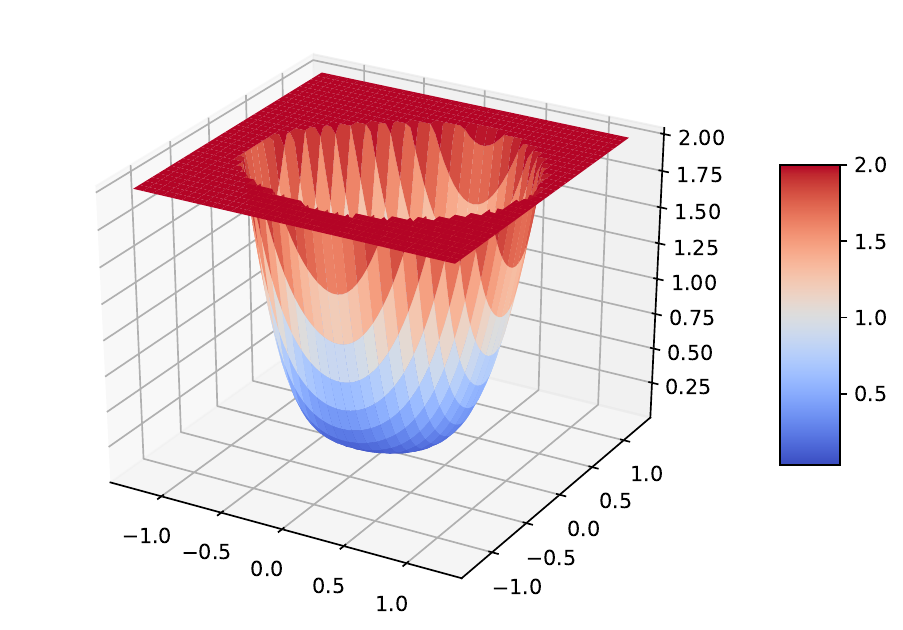}
        \caption{SGDM}
        \label{subfig:sgdm}
    \end{subfigure}
    \hfill
    \begin{subfigure}[t]{0.24\linewidth}
        \centering
        \includegraphics[width=\linewidth]{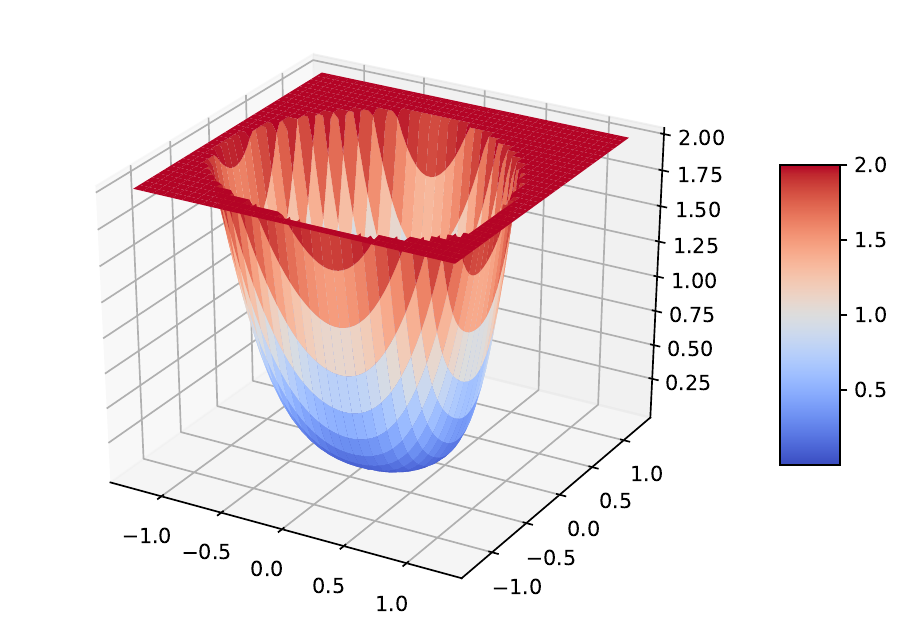}
        \caption{SGDF}
        \label{subfig:sgdf}
    \end{subfigure}

    \caption{Visualization of loss landscape. Adam converges to sharp minima.}
    \label{fig:loss_landscape}
\end{figure}

We visualized the loss landscapes of models trained with SGD, SGDM, SGDF, and Adam using the ResNet-18 model on CIFAR-100, following the method in~\citep{li2018visualizing}. All models are trained with the same hyperparameters for 200 epochs, as detailed in Sec.~\ref{sec:section4.1}. As shown in Fig.~\ref{fig:loss_landscape}, SGDF finds flatter minima. Notably, the visualization reveals that Adam is more prone to converge to sharper minima.

\subsection{Extended Experiment.}

\begin{figure}[ht]
    \centering
    \begin{subfigure}[t]{0.32\linewidth}
        \centering
        \includegraphics[width=\linewidth]{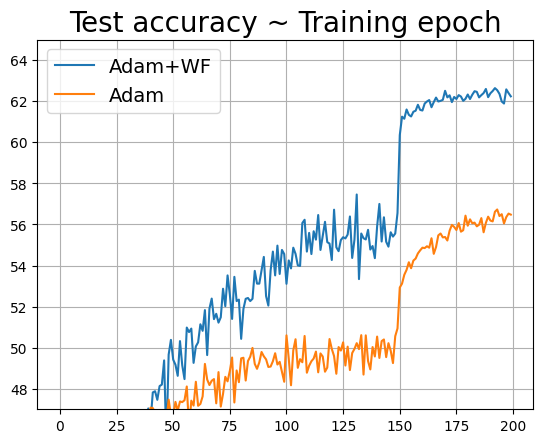}
        \caption{VGG11 on CIFAR-100}
        \label{subfig:vgg_cifar100}
    \end{subfigure}
    \hfill
    \begin{subfigure}[t]{0.32\linewidth}
        \centering
        \includegraphics[width=\linewidth]{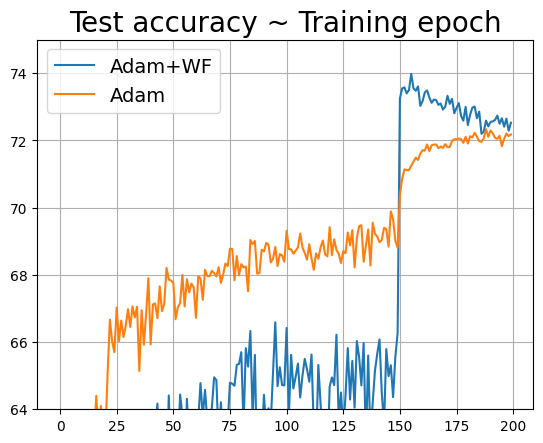}
        \caption{ResNet34 on CIFAR-100}
        \label{subfig:resnet_cifar100}
    \end{subfigure}
    \hfill
    \begin{subfigure}[t]{0.32\linewidth}
        \centering
        \includegraphics[width=\linewidth]{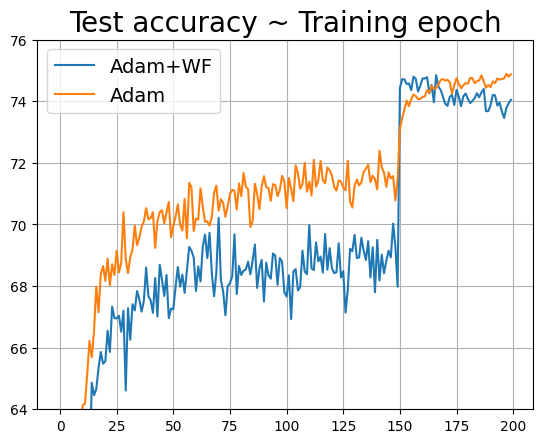}
        \caption{DenseNet121 on CIFAR-100}
        \label{subfig:densenet_cifar100}
    \end{subfigure}

    \caption{Test accuracy of CNNs on CIFAR-100 dataset. We train vanilla Adam and Adam combined with Wiener Filter.}
    \label{fig:compare}
\end{figure}

The study involves evaluating the vanilla Adam optimization algorithm and its enhancement with a Wiener filter on the CIFAR-100 dataset. Fig.~\ref{fig:compare} contains detailed test accuracy curves for both methods across different models. The results indicate that the adaptive learning rate algorithms exhibit improved performance when supplemented with the proposed first-moment filter estimation. This suggests that integrating a Wiener filter with the Adam optimizer may improve performance.

\subsection{Optimizer Test.}
\begin{figure}[t]
    \centering
    \begin{subfigure}[t]{0.33\linewidth}
        \centering
        \includegraphics[width=\linewidth]{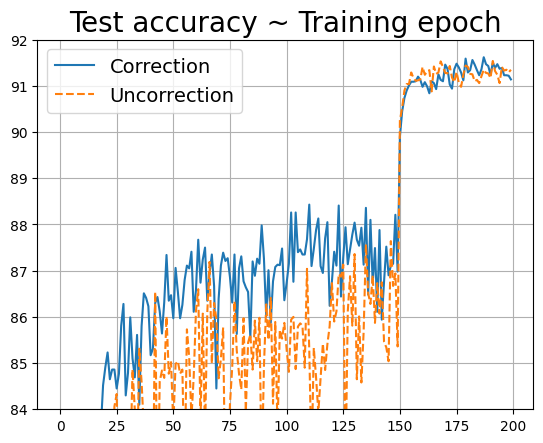}
        \caption{VGG11 on CIFAR-10}
        \label{subfig:cifar10_vgg}
    \end{subfigure}
    \hfill
    \begin{subfigure}[t]{0.33\linewidth}
        \centering
        \includegraphics[width=\linewidth]{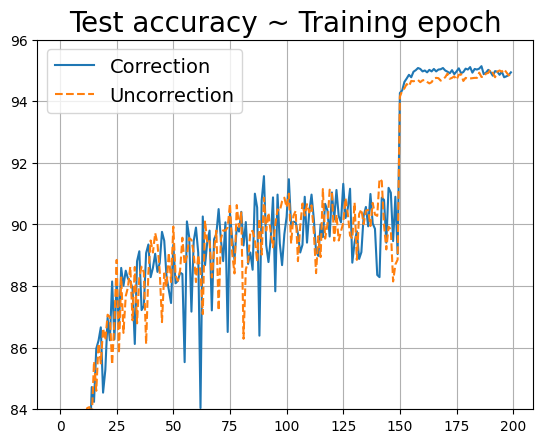}
        \caption{ResNet34 on CIFAR-10}
        \label{subfig:cifar10_resnet}
    \end{subfigure}
    \hfill
    \begin{subfigure}[t]{0.33\linewidth}
        \centering
        \includegraphics[width=\linewidth]{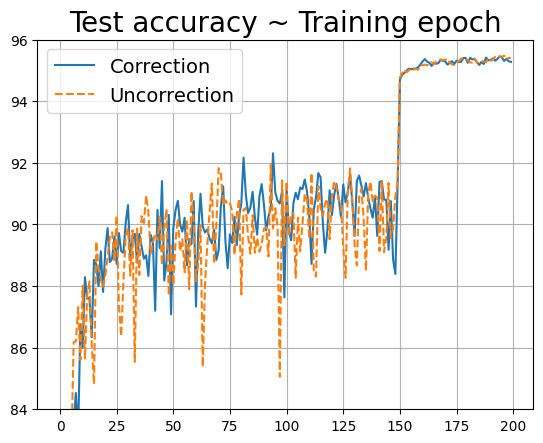}
        \caption{DenseNet121 on CIFAR-10}
        \label{subfig:cifar10_densenet}
    \end{subfigure}
    
    \vspace{2mm}
    
    \begin{subfigure}[t]{0.33\linewidth}
        \centering
        \includegraphics[width=\linewidth]{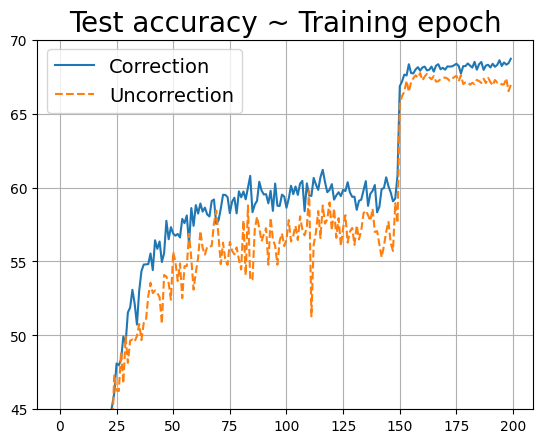}
        \caption{VGG11 on CIFAR-100}
        \label{subfig:cifar100_vgg}
    \end{subfigure}
    \hfill
    \begin{subfigure}[t]{0.33\linewidth}
        \centering
        \includegraphics[width=\linewidth]{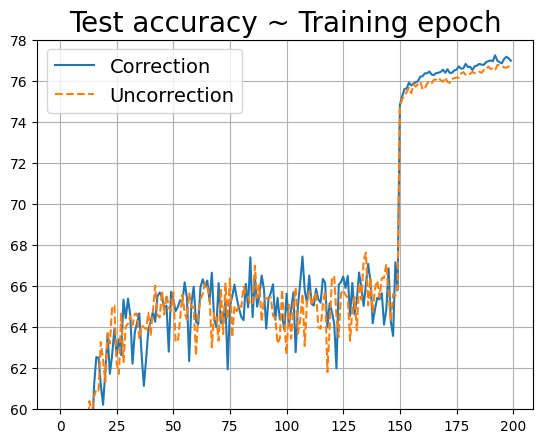}
        \caption{ResNet34 on CIFAR-100}
        \label{subfig:cifar100_resnet}
    \end{subfigure}
    \hfill
    \begin{subfigure}[t]{0.33\linewidth}
        \centering
        \includegraphics[width=\linewidth]{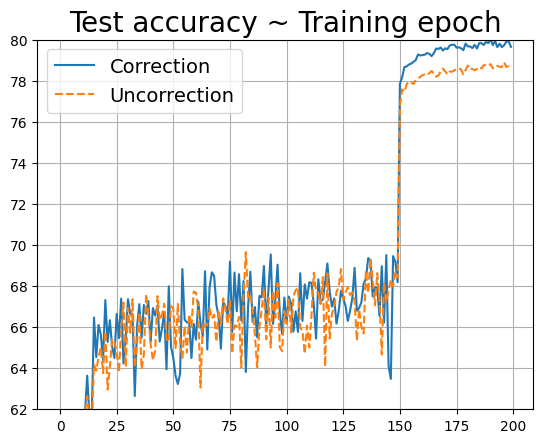}
        \caption{DenseNet121 on CIFAR-100}
        \label{subfig:cifar100_densenet}
    \end{subfigure}

    \caption{SGDF with or without the correction factor. The curve shows the accuracy of the test.}
    \label{fig:correction}
\end{figure}

We derived a correction factor $(1- \beta_{1})(1 - \beta_{1}^{2t}) / (1 + \beta_{1})$ from the geometric progression to correct the variance of by the correction factor. So we test the SGDF with or without correction in VGG, ResNet, DenseNet on CIFAR. We report both test accuracy in Fig.~\ref{fig:correction}. It can be seen that the SGDF with correction exceeds the uncorrected one.

\subsection{Discussion.}
\label{sec:appendix_discuss}
In our framework, the EMA-Momentum is treated as a low-pass filter, in the nature of noise reduction. Cutkosky~\etal~\cite{cutkosky2020momentum} also proves the property that EMA-Momentum cancels out noise, further supporting our analysis. We further discuss classical momentum here. 

Theoretically, we show that momentum converges faster than SGD in the setting of $\mu$-strong acceleration, but deep learning optimization does not always conform to this. Leclerc~\etal~\cite{leclerc2020two} tested the classical momentum at different learning rates, taking the momentum factor \{0, 0.5, 0.9\}. It is empirically found that it is at small learning rates that the classical momentum speeds up the convergence of training losses. That is, SGD-CM can be either better or worse than SGD. In addition, Kunstner~\etal~\cite{kunstner2023noise} found that the classical momentum can only show an advantage over SGD when the batch size increases and approaches the full gradient, at which point the noise introduced by random sampling is almost non-existent. In our proof, we mentioned that SGD-CM introduces both bias and variance, but with a full gradient, SGD-CM does not introduce noise and only causes the gradient to produce bias.

We have not analyzed the nature of bias and variance for convergent solutions, but a certain amount of bias may lead to better results when the noise is reduced, and intuitively this may help the algorithm to discuss saddle points or local minima and converge to flatter regions, in a similar nature to the implicitly flat regularity introduced by noise~\cite{yang2023stochastic}. Because the algorithm converges, the gradient at the position of convergence must be stable, and the classical momentum accumulation gradient, with its large values, must go to a smooth plateau in order to avoid oscillations. Also, it is implied that the gradient bias may not produce irretrievable results, since the bias decreases as the gradient converges, and the direction of the gradient may be more important. Sign SGD~\cite{bernstein2018signsgd} takes sign for the gradient, which also converges, and only needs to be applied to the cosine learning rate decay.

Our overall opinion is that CM does not accelerate SGD, but brings better generalization. Early deep learning optimizations focused on reducing the noise introduced by SGD, resulting in several variance reduction algorithms, where reducing variance increases the speed of convergence \cite{bottou2018optimization}. The noise introduced by CM hinders convergence, but bias brings better generalization. Thus, the above empirical observation that the momentum method can only be accelerated at small learning rates is due to the reduced step size of SGD, which naturally slows down the convergence rate. Whereas the bias from CM offsets the effect of the reduced step size, and the step size reduces the variance of the gradient sequence. This also implies why deep learning uses warm-up to make the gradient more stable in the pre-training period\cite{liu2019variance}.

\begin{figure}[htbp]
    \centering
    \begin{subfigure}{0.48\textwidth}
        \centering
        \includegraphics[width=\textwidth]{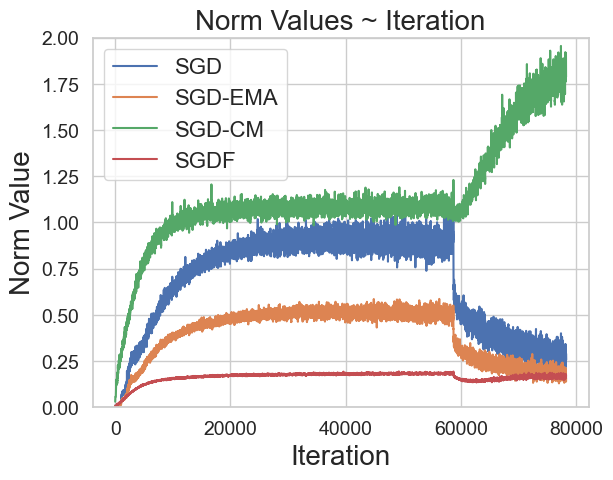}
        \caption{VGG Norm Values}
        \label{fig:vgg-norm}
    \end{subfigure}  
    \begin{subfigure}{0.48\textwidth}
        \centering
        \includegraphics[width=\textwidth]{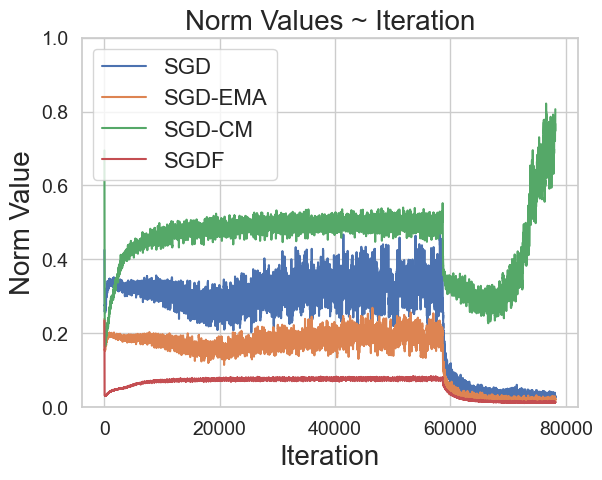}
        \caption{ResNet Norm Values}
        \label{fig:resnet-norm}
    \end{subfigure}
    \caption{Norm Values Comparison for different algorithms.}
    \label{fig:norm-comparison}
\end{figure}

\begin{figure}[htbp]
\centering
\begin{subfigure}{0.36\textwidth}
    \centering
    \includegraphics[width=\textwidth]{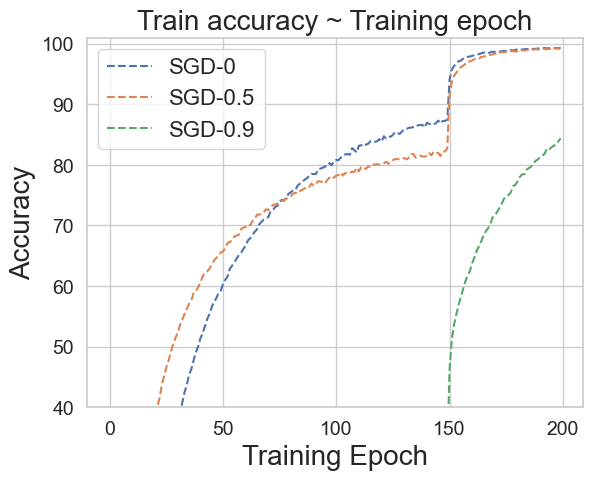}
    \caption{VGG train accuracy.}
    \label{fig:cm-vgg-train}
\end{subfigure}
\begin{subfigure}{0.36\textwidth}
    \centering
    \includegraphics[width=\textwidth]{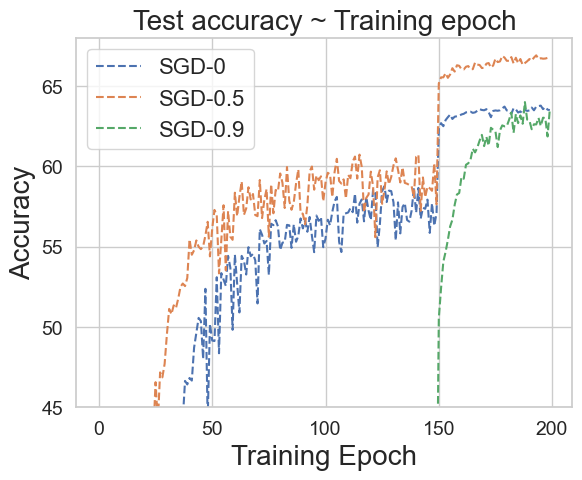}
    \caption{VGG test accuracy.}
    \label{fig:cm-vgg-test}
\end{subfigure}
\vspace{2mm}
\begin{subfigure}{0.36\textwidth}
    \centering
    \includegraphics[width=\textwidth]{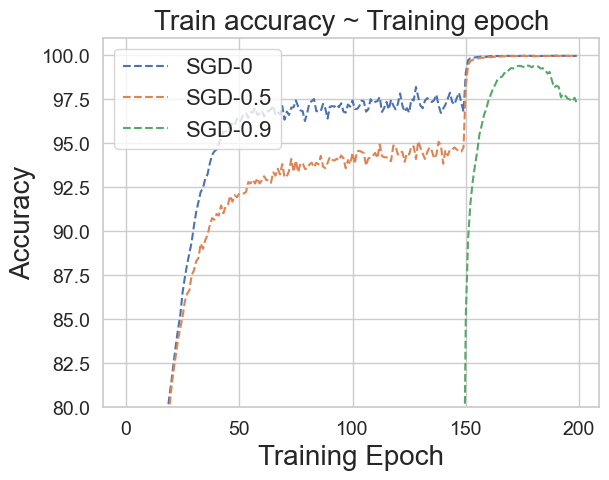}
    \caption{ResNet train accuracy.}
    \label{fig:cm-resnet-train}
\end{subfigure}
\begin{subfigure}{0.36\textwidth}
    \centering
    \includegraphics[width=\textwidth]{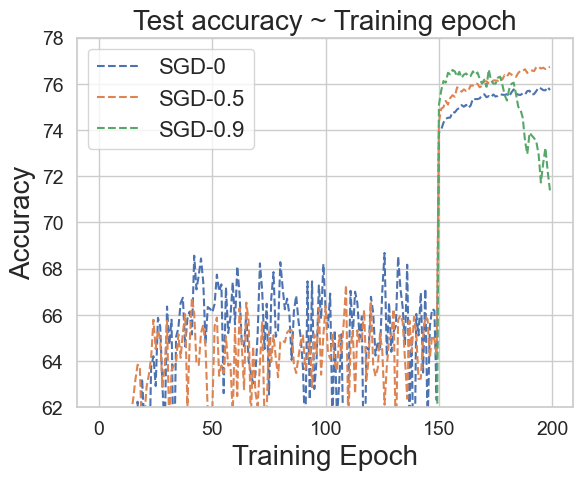}
    \caption{ResNet test accuracy.}
    \label{fig:cm-resnet-test}
\end{subfigure}
\caption{ResNet and VGG with different classical momentum factors.}
\label{fig:cm-comparison}
\end{figure}

Finally, we illustrate the performance of the classical momentum (CM) factors $\{0, 0.5, 0.9\}$ under a consistent learning rate of 0.1 and compare the 2-norm of gradients for SGD, SGD-EMA, SGD-CM, and SGDF during the training of ResNet-18 and VGG-11 networks, as shown in Fig.~\ref{fig:norm-comparison} and Fig.~\ref{fig:cm-comparison}. According to Fig.~\ref{fig:norm-comparison}, the 2-norm of SGDF remains notably stable compared to SGD and SGD-EMA, while the cumulative momentum in SGD-CM surpasses that of the other methods, resulting in an increased 2-norm as the learning rate decreases. From Fig.~\ref{fig:cm-comparison}, it is evident that reducing the momentum factor does not enable SGD-CM to outperform SGD on the training set in terms of acceleration; however, it consistently yields improved results on the test set. 

\end{document}